\documentclass[preprint,12pt]{elsarticle}

\usepackage{amssymb}

\usepackage{url}
\usepackage{amsmath}
\usepackage{amsthm}
\usepackage{subcaption}
\usepackage{pgfplots}
\usepackage{pgfplotstable}
\usepackage{amsfonts}
\usepackage{multirow}
\usepackage{algorithm}
\usepackage{algorithmicx}
\usepackage[noend]{algpseudocode}
\usepackage{placeins}
\usepackage{dsfont}
\usepackage{hyperref}
\usepackage{adjustbox}

\newtheorem{theorem}{Theorem}
\newtheorem{definition}{Definition}

\newtheorem{remark}{Remark}
\newtheorem{corollary}{Corollary}

\DeclareMathOperator*{\sign}{sign}
\newcommand{\E}{\mathbb{E}}
\DeclareMathOperator*{\regularization}{\ensuremath{{\theta}}}
\newcommand{\set}[1]{\mathcal{#1}}
\newcommand{\pnorm}[1]{\lVert{#1}\rVert}
\DeclareMathOperator*{\loss}{{\ell}}
\newcommand{\RN}{\mathbb{R}}
\newcommand{\x}{\ensuremath{\vec{x}}}
\newcommand{\z}{\ensuremath{\vec{z}}}
\newcommand{\y}{\ensuremath{y}}
\newcommand{\xorig}{\ensuremath{\vec{x}_{\text{orig}}}}
\newcommand{\yorig}{\ensuremath{y_{\text{orig}}}}
\newcommand{\setX}{\ensuremath{\set{X}}}
\newcommand{\setY}{\ensuremath{\set{Y}}}
\newcommand{\xcf}{\ensuremath{\vec{x}_{\text{cf}}}}
\newcommand{\ycf}{\ensuremath{y_{\text{cf}}}}

\newcommand{\deltacf}{\ensuremath{\vec{\delta}_\text{cf}}}
\newcommand{\w}{\ensuremath{\vec{w}}}

\newcommand{\density}{\ensuremath{\phi}}
\newcommand{\myCF}[2]{\ensuremath{\text{CF}(#1,#2)}}
\newcommand{\dimsym}{d}
\newcommand{\setD}{\set{D}}

\newcommand{\classifier}{\ensuremath{h}}
\newcommand{\refeq}[1]{Eq.~\eqref{#1}}
\newcommand{\refdef}[1]{Definition~\ref{#1}}
\newcommand{\datapoisoning}{\ensuremath{T}}

\usetikzlibrary{shapes.callouts}
\usetikzlibrary{intersections}
\usetikzlibrary{pgfplots.statistics}
\usetikzlibrary{shapes.geometric, arrows.meta}
\usepgfplotslibrary{colorbrewer}

\tikzstyle{arrow} = [thick,->,>=stealth] 
\tikzstyle{eqarrow} = [thick,<->,>=stealth] 
\pgfplotsset{compat=1.13}

\tikzstyle{default} = [rectangle, rounded corners, minimum width=3cm, minimum height=1cm,text centered, draw=black, fill=blue!30]

\begin{document}

\begin{frontmatter}



\title{The Effect of Data Poisoning on Counterfactual Explanations}

\author[label1,label2]{Andr\'e Artelt}
\ead{aartelt@techfak.uni-bielefeld.de}
\author[label3]{Shubham Sharma}
\author[label4]{Freddy Lecu\'e}
\author[label1]{Barbara Hammer}

 \affiliation[label1]{organization={Bielefeld University},
            city={Bielefeld},
             country={Germany}}
 \affiliation[label2]{organization={University of Cyprus},
             city={Nicosia},
             country={Cyprus}}
\affiliation[label3]{organization={J.P. Morgan AI Research},
            city={New York City},
            country={USA}}
\affiliation[label4]{organization={Inria, Sophia Antipolis},
            country={France}}

\begin{abstract}
Counterfactual explanations are a widely used approach for examining the predictions of black-box systems. They can offer the opportunity for computational recourse by suggesting actionable changes on how to alter the input to obtain a different (i.e., more favorable) system output. However, recent studies have pointed out their susceptibility to various forms of manipulation.

This work studies the vulnerability of counterfactual explanations to data poisoning. We formally introduce and investigate data poisoning in the context of counterfactual explanations for increasing the cost of recourse on three different levels: locally for a single instance, a sub-group of instances, or globally for all instances.
In this context, we formally introduce and characterize data poisonings, from which we derive and investigate a general data poisoning mechanism.
We demonstrate the impact of such data poisoning in the critical real-world application of explaining event detections in water distribution networks. Additionally, we conduct an extensive empirical evaluation, demonstrating that state-of-the-art counterfactual generation methods and toolboxes are vulnerable to such data poisoning. Furthermore, we find that existing defense methods fail to detect those poisonous samples.
\end{abstract}

%


\end{frontmatter}


\section{Introduction}
Real-world Artificial Intelligence (AI-) and Machine Learning (ML-) based systems~\cite{zhao2023survey,ho2022cascaded} show an impressive performance but are still not perfect -- e.g., failures, issues of fairness, and vulnerability to manipulations can cause harm.
ML-based systems can be manipulated or attacked at different stages to harm the general predictive accuracy, introduce failures, increase the unfairness, or place backdoors in the system. In this context, adversarial attacks~\cite{qian2022survey}, backdoor attacks~\cite{Li_Jiang_Li_Xia_2024}, and data poisoning~\cite{Fan_Yan_Li_Qu_Xiao_2022} constitute the most popular methods for manipulating ML-based systems.

Adversarial attacks~\cite{qian2022survey,zhao2022adversarial} implement an attack occurring at run-time that aims at imperceptible input manipulations, inducing system failures such as the computation of wrong outputs.

Backdoor attacks~\cite{Li_Jiang_Li_Xia_2024}, however, affect ML-based systems in the training stage and refer to the implementation of different model behavior that becomes only active when a certain pattern (the backdoor) is present in the input. Backdoor attacks can be realized by data poisoning~\cite{Chen_Liu_Li_Lu_Song_2017} or by using a special loss function that encodes the backdoor behavior~\cite{Li_Jiang_Li_Xia_2024}. Note that the latter implementation approach of using a special loss function poses strong assumptions on the attacker's capabilities.

Like backdoor attacks, data poisoning attacks~\cite{Fan_Yan_Li_Qu_Xiao_2022} affect ML-based systems in the training stage by manipulating training samples or adding new instances such that, for instance, the predictive performance (e.g, accuracy) of the final trained model decreases~\cite{lin2021ml,tolpegin2020data}, or fairness issues arise~\cite{mehrabi2021exacerbating}. Note that, unlike adversarial attacks happening at inference time, data poisoning and backdoor attacks integrate manipulations into the final model and consequently impact its internal reasoning.
More specifically, data poisoning can be performed offline~\cite{lin2021ml} or online~\cite{tolpegin2020data}, and only small modifications are made to the training data, such as changing labels, removing samples, or adding new instances, which are likely to remain unnoticed. This poses a real threat in practice because nowadays many large models are trained on huge (internet-based) data sets~\cite{zhao2023survey,ho2022cascaded}, where it is impossible to check data in detail and therefore poisonous data might affect a large number of models trained directly on the data or indirectly using some pre-trained embeddings or models~\cite{shan2023prompt,bojchevski2019adversarial,yang2023data}.

Given the threat of failures (intentionally caused or not), the transparency of such AI- and ML-based systems becomes a crucial aspect. Transparency is important not only to prevent failures 
 but also to create trust in such systems and understand where and how it is safe to deploy them. This was also recognized by the policymakers and therefore found its way into legal regulations such as the EU's GDPR~\cite{GDPR} or the EU AI act~\cite{euAiAct21}.
Explanations are a popular way of achieving transparency and shaping the field of eXplainable AI (XAI)~\cite{dwivedi2023explainable,LONGO2024102301,arrieta2020explainable}.
Nowadays, many different explanation methods exist~\cite{dwivedi2023explainable,arrieta2020explainable,rawal2021recent}. Counterfactual explanations~\cite{CounterfactualWachter} constitute a popular type of explanation method, which is inspired by human explanations~\cite{CounterfactualsHumanReasoning} and can be used to provide recourse to individuals.
More specifically, a counterfactual explanation provides (computational) recourse by stating actionable recommendations on how to alter the system's output in some desired way -- e.g., how to turn a rejected loan application into an accepted one.

Critically, recent work has shown that many XAI methods are vulnerable to adversarial manipulations~\cite{baniecki2023adversarial,baniecki2022fooling,slack2021counterfactual}, undermining users' trust in XAI methods for revealing the internal logic of a model.
In the context of counterfactual explanations, it was observed that they are neither robust to model changes~\cite{mishra2021survey}, nor to input perturbations~\cite{artelt2021evaluating,virgolin2023robustness,jiang2024robust}, and also not to adversarial training for implementing backdoors~\cite{slack2021counterfactual}.
Data poisoning attacks on counterfactual explanations could make their recommended actions more costly 
-- either for all individuals or for a subset of individuals. 
Since counterfactual explanations state actionable recommendations that are to be executed in the real world, manipulated explanations would directly affect the individuals by enforcing more costly actions or hiding some information from them.
Although counterfactual explanations are a popular and widely used explanation method, the effect of data poisoning on them has not been studied yet.

\textbf{Our contribution:}
The novelty of this work lies in the study of the vulnerability of counterfactual explanations to data poisoning. For this, we formalize and identify a set of data poisoning mechanisms for counterfactual explanations that injects a small set of realistic but poisonous data instances into the training data set such that the counterfactual explanations of the newly trained classifier are more costly to execute for the user. We consider the effect of data poisoning on three different levels: locally for an individual, a sub-group of individuals, and globally for all individuals. Most importantly, our proposed method is model-agnostic and only needs access to an interface for getting predictions and a mechanism for generating any closest counterfactuals, but no access or knowledge about model internals is required. We empirically find that existing state-of-the-art methods for computing counterfactuals are vulnerable to data poisoning and that classic defense mechanisms fail to detect those poisonous training samples.

The remainder of this work is structured as follows: First, we discuss the related work (Section~\ref{sec:relatedwork}), and the necessary foundations of counterfactual explanations and (computational) recourse in Section~\ref{sec:foundations}. Next (in Section~\ref{sec:method}), we introduce our formalization of a data poisoning attack and introduce our proposed data poisoning attack. In Section~\ref{sec:casestudy}, we demonstrate and illustrate the impact of such data poisoning attacks in the critical real-world application of explaining event detection in water distribution networks. In addition to that, we perform an extensive quantitative empirical evaluation of our proposed data poisoning attack on different benchmark scenarios in Section~\ref{sec:experiments}. Finally, this work closes with a summary and conclusion, including the discussion of limitations and possible directions for future research, in Section~\ref{sec:conclusion}.
Note that all proofs and more detailed evaluations of the experiments can be found in the appendix.
\section{Related Work}\label{sec:relatedwork}
\subsection{General Data Poisoning}
Existing data poisoning strategies from the literature impact ML models at the training stage, to affect predictive performance~\cite{lin2021ml,tolpegin2020data} or fairness~\cite{mehrabi2021exacerbating}. However, generic methods applicable to any black-box model are rare~\cite{lin2021ml}. Most (advanced) data poisoning methods are either tailored towards specific models or classes of models (e.g., neural networks), domains (e.g. computer vision), or rely on assumptions such as feature extractors~\cite{lin2021ml}. The Label-Flipping attack~\cite{lin2021ml} (randomly) flips labels in the training data set, and constitutes the most general but also a rather simple, yet highly effective, data poisoning attack.
As a reaction to such data poisoning attacks, potential countermeasures and defense strategies have also been proposed~\cite{Fan_Yan_Li_Qu_Xiao_2022,DBLP:conf/nips/SteinhardtKL17,Paudice_Munoz-Gonzalez_Lupu_2019,Paudice_Munoz-Gonzalez_Gyorgy_Lupu_2018}.
Data sanitization methods~\cite{Cretu_Stavrou_Locasto_Stolfo_Keromytis_2008} constitute among the most popular defense methods. Those methods usually rely on outlier and anomaly detection methods~\cite{Paudice_Munoz-Gonzalez_Gyorgy_Lupu_2018} for detecting the poisonous samples and either correct them~\cite{Paudice_Munoz-Gonzalez_Lupu_2019} or remove them from the training data~\cite{Paudice_Munoz-Gonzalez_Gyorgy_Lupu_2018,Cretu_Stavrou_Locasto_Stolfo_Keromytis_2008}. 
Inspired by formal guarantees on the adversarial robustness, there also exist formal guarantees (under some assumptions) on the effectiveness of certain data sanitization defenses~\cite{DBLP:conf/nips/SteinhardtKL17}.
However, as noted in~\cite{Koh_Steinhardt_Liang_2021}, data sanitization methods remain imperfect in the light of more advanced data poisoning attacks.

\subsection{Vulnerability and Manipulations of XAI}
Most existing work~\cite{baniecki2023adversarial,brown2023making,noppel2024brief} on exposing the vulnerability of explanations is centered in the vision domain and focuses either on adversarial examples or model manipulation.
In particular, it was shown that many XAI methods (incl. counterfactual explanations) are not robust with respect to adversarial manipulations in the input~\cite{baniecki2023adversarial,artelt2021evaluating,Zhang_Chao_Dasegowda_Wang_Kalra_Yan_2022} -- i.e., small and potentially imperceptible changes in the input can lead to completely different explanations. This also relates to fairness issues regarding individual fairness, where one would like to ensure that similar individuals get similar explanations~\cite{Zhao_Wang_Derr_2023,hancox2020robustness}. Similarly, it was also observed that explanations sometimes differ significantly between protected groups, violating group fairness~\cite{Zhao_Wang_Derr_2023,artelt2023ijcai}.

Only very little work considers (domain-independent) data poisoning of XAI methods~\cite{baniecki2023adversarial}. For instance, there exist data poisoning against partial dependence plots~\cite{baniecki2022fooling}, SHAP~\cite{baniecki2022manipulating}, and concept-based explainability tools~\cite{brown2023making}.
The authors of~\cite{baniecki2022fooling} propose a genetic algorithm for perturbing the training data such that SHAP importance scores change. They assume that it is possible to modify (possibly) all samples in the training set, which might constitute a strong and unrealistic assumption in reality. Furthermore, changing many (or all) samples in the training data set might harm the model's predictive performance -- this, however, is not evaluated in~\cite{baniecki2022fooling}.
A similar approach, with the same limitations, is proposed in~\cite{baniecki2022manipulating} where partial dependence plots are targeted.
Surprisingly, none of the existing works on data poisoning of explanations discuss any potential defense mechanisms.

In the context of counterfactual explanations, data poisoning attacks have not been considered so far. The closest study was done by~\cite{slack2021counterfactual}, proposing an adversarial training objective for planting a backdoor in a neural network such that the cost of recourse decreases for a sub-group of individuals, violating group fairness requirements. Note that this approach is model-specific and different from data poisoning since it proposes the use of a malicious cost function and therefore assumes full control over the entire training procedure.

\paragraph{Literature Gap and Our Contribution}
From the literature review, it becomes apparent that 1) data poisoning on counterfactual explanations has not been studied so far, and 2) existing work on data poisoning of XAI methods assumes that all training instances can be manipulated, which constitutes an unrealistic assumption in many attack scenarios. Furthermore, because data poisoning can lead to fairness issues regarding the model's prediction, it might also create fairness issues regarding the explanations -- similar to the group fairness violations created by the backdoor attacks as proposed in~\cite{slack2021counterfactual}. Finally, defense mechanisms against data poisoning attacks on XAI remain an open question.

In this work, we investigate data poisoning of counterfactual explanations by injecting additional training instances into the training data set. We argue that this constitutes a more realistic attack scenario in practice than the manipulation of existing training data instances as done in~\cite{baniecki2022manipulating}. Furthermore, we do not alter the loss function used in the training procedure as done in the backdoor attack proposed by~\cite{slack2021counterfactual}, posing strong assumptions about the attacker's capabilities. We not only investigate the effect of the data poisoning on the average explanation, but also investigate potential fairness issues in the counterfactual explanations arising from the poisoning.
Finally, we empirically evaluate classic defense methods for detecting the poisonous training samples.

\section{Foundations of Counterfactuals \& Computational Recourse}\label{sec:foundations}
A counterfactual explanation (often just called counterfactual) states actionable modifications to the features of a given instance such that the system's output changes. Usually, an explanation is requested in the case of an unexpected or unfavorable outcome~\cite{riveiro2022challenges} -- in the latter case, a counterfactual is also referred to as \textit{recourse}~\cite{karimi2021survey}, i.e.\ recommendations on how to turn the unfavorable into a favorable outcome.
Because counterfactuals can mimic ways in which humans explain~\cite{CounterfactualsHumanReasoning}, they constitute one of the most popular explanation methods in literature and in practice~\cite{molnar2019,CounterfactualReviewChallenges}.

The two important properties of counterfactual explanations~\cite{CounterfactualWachter} are the:
\begin{enumerate}
    \item \emph{contrasting property:} requiring a change in the output of the system.
    \item \emph{cost of the counterfactual:} the cost and effort it takes to execute the counterfactual in the real world should be as low as possible to maximize its usefulness -- e.g.\ counterfactuals with very few modifications or as small as possible modifications.
\end{enumerate}
Both properties can be combined into an optimization problem (see \refdef{def:counterfactual}).
\begin{definition}[(Closest) Counterfactual Explanation]\label{def:counterfactual}
Assume a classifier (binary or multi-class) $\classifier:\RN^\dimsym \to \setY$ is given. Computing a counterfactual $\deltacf \in \RN^\dimsym$ for a given instance $\xorig \in \RN^\dimsym$ is phrased as the following optimization problem:
\begin{equation}\label{eq:counterfactualoptproblem}
\underset{\deltacf \,\in\, \RN^\dimsym}{\arg\min}\; \loss\big(\classifier(\xorig 
+ \deltacf), \ycf\big) + C \cdot \regularization(\deltacf)
\end{equation}
where $\loss(\cdot)$ implements the contrasting property by means of a loss function that
penalizes deviation of the prediction $\classifier(\xcf:=\xorig + \deltacf)$ from the requested outcome $\ycf$; $\regularization(\cdot)$ states the cost of the explanation (e.g.\ cost of recourse) which should be minimized; $C>0$ denotes the regularization strength balancing the two properties.

The short-hand notation $\deltacf = \myCF{\x}{\classifier}$ denotes the counterfactual (i.e. solution to~\refeq{eq:counterfactualoptproblem}) $\deltacf$ of an instance $\x$ under a classifier $\classifier(\cdot)$ iff the target outcome $\ycf$ is uniquely determined.
\end{definition}
Note that the cost of the counterfactual, here modeled by $\regularization(\cdot)$, is highly domain and use-case specific and therefore must be chosen carefully in practice, potentially requiring domain knowledge.
Usually, the number of changes (e.g., changed features) and/or the magnitude of changes are considered as the cost of a counterfactual. Consequently, in many implementations and toolboxes~\cite{guidotti2022counterfactual}, the $p$-norm is used as the default cost function for continuous features because it is most generic and can be easily adjusted by a custom weighting scheme.
\begin{equation}\label{eq:cost_of_recourse}
    \regularization(\deltacf) = \pnorm{\deltacf}_p
\end{equation}
\begin{remark}\label{remark:costrecourse}
    In the case of recourse -- i.e.\ a counterfactual $\deltacf$ (\refdef{def:counterfactual}) for turning an unfavorable into a favorable outcome\footnote{Here, ``favorable'' and ``unfavorable'' refer to specific predicted labels of the classifier.} --, we refer to the cost $\regularization(\deltacf)$, as the \textit{cost of recourse}.
\end{remark}
Because counterfactual explanations are usually requested in the case of an unfavorable outcome~\cite{riveiro2022challenges}, we often refer to the cost of recourse as the quantity of interest in the remainder of this work.

Besides those two essential properties (contrasting and cost), there exist additional relevant aspects such as plausibility~\cite{CounterfactualGuidedByPrototypes,poyiadzi2020face}, diversity~\cite{mothilal2020explaining}, robustness~\cite{wang2023flexible,zhang2023density,leofante2023promoting}, and fairness~\cite{artelt2023ijcai,von2022fairness,sharma2021fair,sharma2020certifai}. which have been addressed in literature~\cite{guidotti2022counterfactual}. However, the basic formalization~\refeq{eq:counterfactualoptproblem} is still very popular and widely used in practice~\cite{CounterfactualReviewChallenges,guidotti2022counterfactual}.
In this context, it is also important to note that the cost of recourse (i.e., cost of the counterfactual) remains the central quantity of interest to the user because it denotes how easy it is for them to achieve their desired goal -- e.g., turning an unfavorable prediction into a favorable one. While other aspects, such as the aforementioned robustness and plausibility, might also be relevant, the cost of recourse, which might be influenced by those additional aspects, remains the most critical quantity for the user and is therefore always evaluated when evaluating the quality of counterfactuals.

Finally, there exist numerous methods and implementations/toolboxes for computing counterfactual explanations in practice~\cite{guidotti2022counterfactual} -- i.e. methods for computing solutions to~\refeq{eq:counterfactualoptproblem}. Note that most of those methods include some additional aspects such as plausibility and diversity:
%
\begin{itemize}
    \item \textit{Counterfactuals Guided by Prototypes}~\cite{CounterfactualGuidedByPrototypes} is a method focusing on plausibility. Here, a set of plausible instances (prototypes) is used to pull the final counterfactual instance (i.e., $\xcf = \xorig + \deltacf$) closer to these plausible instances, making the final counterfactual more realistic.

    \item \textit{DiCE}~\cite{mothilal2020explaining} is a model-agnostic method and Python toolbox
for computing a set of diverse closest counterfactual explanations instead of a single one only. 

    \item \textit{Nearest Unlike Neighbor method}~\cite{Dasarathy_1995} constitutes a baseline method for computing plausible counterfactual explanations that can be implemented by picking the closest sample, with the requested output $\ycf$, from a given set (e.g., training set) as the counterfactual instance.
\end{itemize}

\section{Data Poisoning of Counterfactual Explanations}\label{sec:method}
Given the fact that counterfactual explanations constitute a local explanation, potential data poisonings can have effects on different levels or areas in data space:
\begin{itemize}
    \item \emph{Global effect:} Explanations of all individuals are affected.
    \item \emph{Sub-groups effect:} Explanations of only one or multiple sub-groups are affected
    \item \emph{Local effect:} Explanations of only a single individual/instance are affected.
\end{itemize}
At the same time, data poisoning can aim for different effects on counterfactual explanations, such as hiding attributes or increasing the cost of recourse (i.e., cost of the counterfactual, see Remark~\ref{remark:costrecourse}). Since providing (computational) recourse is a core application of counterfactuals, increasing the cost of recourse has the most severe consequence in the real world because it would harm individuals directly by making the commended actions more costly. 
Therefore, in this work, we focus on data poisoning for increasing the cost of recourse, but also evaluate possible side effects on validity, sparsity, and plausibility. Furthermore, note that the plausibility of counterfactuals can be easily checked, and therefore, attacks targeting the plausibility would be detected more easily than attacks targeting the cost of recourse, for which it is more difficult to say whether this corresponds to the ground truth or was manipulated.

\subsection{Data Poisoning for Increasing the Cost of Recourse}\label{sec:datapoisoning:formal}
In this work, we study the effect of data poisoning on the cost of recourse (Remark~\ref{remark:costrecourse}). That is, we focus on data poisoning with the primary goal of increasing the cost of recourse, in a pre-defined region in data space, as stated in~\refdef{def:datapoisoning:costrecourse}.
\begin{definition}[Data Poisoning for Increasing the Cost of Recourse]\label{def:datapoisoning:costrecourse}
Given an original training data set $\setD_{\text{orig}}\subset\{\setX\times\setY\}^n$ and a probability density $\phi(\cdot)$ assigning a high likelihood to targeted instances, we transform (i.e., poison) $\setD_{\text{orig}}$ into a new data  set $\setD_{\text{poisoned}}\subset\{\setX\times\setY\}^m$ by means of a data poisoning mechanism $\datapoisoning: \{\setX\times\setY\}^n \to \{\setX\times\setY\}^m$, such that the cost of recourse $\regularization(\cdot)$ increases for instances under $\density(\cdot)$:
\begin{equation}\label{eq:problem}
\begin{split}
    &\E_{\x\sim \density}\left[\regularization \circ \myCF{\x}{\classifier_{\setD_{\text{poisoned}}}}\right] \gg \E_{\x\sim \density}\left[\regularization \circ \myCF{\x}{\classifier_{\setD_{\text{orig}}}}\right] \\&
    \text{ where } \setD_{\text{poisoned}} = \datapoisoning(\setD_{\text{orig}})
\end{split}
\end{equation}
where $\circ$ denotes the function composition, $\classifier_{\setD}$ denotes a classifier that was derived from the data set $\setD$, and $\myCF{\cdot}{\cdot}$ refers to some given method for generating counterfactuals by solving~\refeq{eq:counterfactualoptproblem}.
\end{definition}
The density $\density(\cdot)$ allows us to vary the level of the poisoning 
-- e.g., for a global effect,  we could use a class-wise density for targeting all instances from a specific (unfavorable) class, or in the case of a local effect, we could use a delta-density to target a single instance or a small group of instances.

In this work, we focus on data poisoning attacks $\datapoisoning(\cdot)$ that add new (poisonous) instances to the training data set to increase the cost of recourse (\refdef{def:datapoisoning:costrecourse}).
In this context, we formally define (see \refdef{def:recourse_poisoning_data}) a \textit{poisonous data set}~\cite{steinhardt2017certified} for increasing the cost of recourse  -- i.e., a data set that increases the cost of recourse (\refdef{def:datapoisoning:costrecourse}) if added to the original training data $\setD_{\text{orig}}$.
\begin{definition}[Poisonous Data Set]\label{def:recourse_poisoning_data}
     For a training data set $\setD_{\text{orig}}\subset\{\setX\times\setY\}^n$ and a probability density $\phi(\cdot)$ assigning a high likelihood to targeted instances, we say that a data set $\setD_{\text{poison}}\subset \{\setX\times\setY\}^m$ is recourse poisoning 
    iff a classifier $\classifier: \setX\to\setY$ trained on $\setD_{\text{poison}}\cup \setD_{\text{orig}}$ shows an increase in the cost of recourse (\refdef{def:datapoisoning:costrecourse}):
    \begin{equation}
        \E_{\x\sim \density}\left[\regularization \circ \myCF{\x}{\classifier_{\setD_{\text{poison}} \cup \setD_{\text{orig}}}}\right] \gg \E_{\x\sim \density}\left[\regularization \circ \myCF{\x}{\classifier_{\setD_{\text{orig}}}}\right]
    \end{equation}
\end{definition}
Consequently, we write the data poisoning mechanisms $\datapoisoning(\cdot)$ (\refdef{def:datapoisoning:costrecourse}) as follows:
\begin{equation}
    \datapoisoning(\setD_{\text{orig}}) =  \setD_{\text{orig}} \cup \setD_{\text{poison}}
\end{equation}
where $\setD_{\text{poison}}$ refers to a poisonous data set from~\refdef{def:recourse_poisoning_data}.

From a practical point of view, besides increasing the cost of recourse (as stated in~\refdef{def:datapoisoning:costrecourse}), it is desirable for the poisonous data set $\setD_{\text{poison}}$ (\refdef{def:recourse_poisoning_data}) to have the following properties:
\begin{enumerate}
    \item The number of poisonous instances $\setD_{\text{poison}}$ (i.e., necessary poisoning budget) should be kept to a minimum:
    \begin{equation}\label{eq:datapoisoning:smallset}
        \arg\min\; |\setD_{\text{poison}}|
    \end{equation}
    \item The poisonous instances $\setD_{\text{poison}}$ are realistic -- i.e., they are on the data manifold $p_{\text{data}}(\cdot)$ and have a high likelihood:
    \begin{equation}\label{eq:datapoisoning:plausibility}
            \underset{}{\arg\max}\; p_{\text{data}}(\x_i, y_i)\quad \forall\, (\x_i,y_i)\in\setD_{\text{poison}}
    \end{equation}
    \item In the case of aiming for a local or sub-group effect, poisonous instances only target those specified groups/areas, but do not affect any other instances -- i.e.\ the cost of recourse of untargeted instances should not change (significantly):
    \begin{equation}\label{eq:datapoisoning:focused}
        \Big|\E_{\x\sim \density'}\left[\regularization \circ \myCF{\x}{\classifier_{\setD_{\text{poison}} \cup \setD_{\text{orig}}}}\right] - \E_{\x\sim \density'}\left[\regularization \circ \myCF{\x}{\classifier_{\setD_{\text{orig}}}}\right]\Big| \leq \epsilon
    \end{equation}
    where $\epsilon>0$ denotes a small threshold, and $\density'$ denotes the density of all untargeted instances -- in practice, this might be just a region in data space or a set of samples that should not be affected by the poisoning.
    \item The predictive performance of the classifier is maintained\footnote{However, because the decision boundary is altered, some drop in the predictive performance might be inevitable.}:
    \begin{equation}\label{eq:datapoisoning:accuracy}
        \underset{}{\arg\min}\; \E[\loss(\classifier_{\setD_{\text{poison}} \cup \setD_{\text{orig}}}(\x_i)), y_i)]
    \end{equation}
    where $\loss(\cdot)$ denotes some suitable loss function such as the zero-one loss.
\end{enumerate}
Later (see Section~\ref{sec:datapoisoning:algo}), we will merge all those properties into a single optimization problem for computing poisonous instances. But first, we study a few (general) aspects and properties of 
poisonous data sets (\refdef{def:recourse_poisoning_data}) in simple settings. The gained knowledge will serve as a foundation for motivating the final data poisoning algorithm (Algorithm~\ref{algo:datapoisoning:fair_cf}).

\subsubsection{Formal Investigation}\label{sec:datapoisoning:theory}
In the following, we study the data poisoning of counterfactuals in a few simple cases where formal statements on the influence of training samples on the decision boundary are feasible.
While the underlying assumptions of the presented theorems are very strong and somewhat unrealistic in practice, they provide us with inspiration on how a general data poisoning mechanism could be designed (see Section~\ref{sec:datapoisoning:algo}). In particular, they provide us with evidence that recourse poisoning sets (\refdef{def:recourse_poisoning_data}) can be constructed from closest counterfactual explanations or adversarial respectively. In the empirical evaluation (Section~\ref{sec:experiments}), we evaluate the performance and correctness of our proposed data poisoning mechanism (Algorithm~\ref{algo:datapoisoning:fair_cf}). In particular, we also run an ablation study (see Section~\ref{sec:exp:ablation}) in which we empirically evaluate how well the presented theoretical findings generalize to more general scenarios.

\paragraph{Locally Increasing the Cost of Recourse}
As discussed in Section~\ref{sec:foundations}, the simplest way of achieving recourse is through the closest counterfactual as stated in~\refdef{def:counterfactual} -- i.e., the smallest change that reaches/crosses the decision boundary. In this case, for data poisoning on a local level, it can be shown that, under strong assumptions, samples on the decision boundary are data poisoning instances (\refdef{def:recourse_poisoning_data}).
\begin{theorem}[Local Recourse Poisoning Data Sets for 1-Nearest Neighbor Classifiers]\label{Theorem:algo:correct:knn}
    Let $\classifier_{\setD}(\cdot)$ be a k-nearest neighbor classifier (with $k=1$) for some data set $\setD$. For any $(\xorig,\yorig)\in\setD$, let $\x'$ denote the closest instance (assuming uniqueness) on the decision boundary under a p-norm $\regularization(\cdot)$.
    
    Then, $\setD_{\text{poison}}=\{(\x',\yorig)\}$ is a poisonous data set (\refdef{def:recourse_poisoning_data}) at $\xorig$ -- i.e. the cost resources increases for $\xorig$, i.e.:
    \begin{equation}
       \regularization \circ \myCF{x}{\classifier_{\setD\cup\{(\x',\yorig)\}}} > \regularization \circ\myCF{x}{\classifier_{\setD}}
    \end{equation}
\end{theorem}
The proof of Theorem~\ref{Theorem:algo:correct:knn} is given in the appendix.
Although a 1-NN classifier is somewhat simplistic, it is quite flexible and might be a good local approximation if sufficiently many training samples are available. Therefore, Theorem~\ref{Theorem:algo:correct:knn} provides valuable insights on the nature of recourse poisoning data sets (\refdef{def:recourse_poisoning_data}) for locally increasing the cost of recourse.

\paragraph{Globally Increasing the Cost of Recourse}
Similar to Theorem~\ref{Theorem:algo:correct:knn}, it is possible, under strong assumptions, to state a poisonous data set (\refdef{def:recourse_poisoning_data}) in the case of a linear Support Vector Machine classifier (SVM) for globally increasing the cost of recourse.
\begin{theorem}[Global Recourse Poisoning Data Set for linear SVM]\label{Theorem:algo:correct:svm}
    Let $\classifier_{\setD}:\RN^\dimsym\to\{-1, 1\}, \;\x\mapsto \sign(\w^\top\x + b)$ be a linear SVM classifier, and assume that the training data set $\setD$ is linearly separable.
    
    Then, a poisonous data set $\setD_{\text{poison}}$ (\refdef{def:recourse_poisoning_data}) for all negatively classified samples (i.e.\ $\forall \,\x\in\RN^\dimsym:\,\classifier(\x)=-1$) is given as follows:
    \begin{equation}\label{eq:datapoisoningset:svm}
        \setD_{\text{poison}} = \{(\x, -1)\ \mid \x \in\RN^\dimsym \text{ with } -\w^\top\x - b \geq 1 - \xi, \;\xi\in(0,1) \}
    \end{equation}
\end{theorem}
The proof of Theorem~\ref{Theorem:algo:correct:svm} is given in the appendix.
Note that Theorem~\ref{Theorem:algo:correct:svm} 
states that a recourse poisoning data set can be constructed by considering all samples inside the maximum margin of $\classifier(\cdot)$.
In practice, however, depending on the training data set $\setD$, it is likely possible that already a small subset of $\setD_{\text{poison}}$~\refeq{eq:datapoisoningset:svm} constitutes a poisonous data set (\refdef{def:recourse_poisoning_data}) as well.

\subsection{A Data Poisoning Algorithm for Increasing the Cost of Recourse}\label{sec:datapoisoning:algo}
Based on the findings from Section~\ref{sec:datapoisoning:formal}, we formalize a method (see Algorithm~\ref{algo:datapoisoning:fair_cf}) for generating poisonous data sets (\refdef{def:recourse_poisoning_data}) -- i.e., poisonous instances that are added to the training set, to increase the cost of recourse.
Consequently, our proposed Algorithm~\ref{algo:datapoisoning:fair_cf} constitutes and implementation of a data poisoning $\datapoisoning(\cdot)$ from~\refdef{def:datapoisoning:costrecourse}. Notably, the proposed method supports data poisonings on different levels (i.e., local, sub-groups, and global levels).
Furthermore, this approach constitutes a more realistic assumption compared to backdoor attacks in~\cite{slack2021counterfactual}, where an attacker is assumed to be able to manipulate the loss function used in training.
More specifically, we make the following assumptions about the attack scenario and the attacker's capabilities and knowledge:
\begin{itemize}
    \item Arbitrary data points can be added to the training data.
    \item The ML model to be poisoned is a black-box to the attacker. The attacker has access to an interface of the ML model for computing predictions only, but no access to any model internals or any other knowledge about the ML model.
    \item Access to an arbitrary method for computing closest counterfactual explanations (i.e., adversarials) of black-box models.
\end{itemize}
Most importantly, we want to highlight that the attacker does not know 
the specific counterfactual algorithm being used in the evaluation.

For practical purposes, we assume that we have (or created) a set of target samples $\setD_{\text{target}}=\{(\x_j,\y)\}$ all with the same prediction $\y\in\{0,1\}$ and $\x_j\sim\density$, from the region in data space that is targeted by the poisoning -- e.g., this could be a subset of the training data set.
We propose to fix the size of the poisonous data set (\refdef{def:recourse_poisoning_data}) (i.e., the number of poisonous instances $\{\z_i\}$ is fixed) and 
merge all desired properties (see Section~\ref{sec:datapoisoning:formal}) into the following multi-objective optimization problem:
\begin{subequations}\label{eq:datapoisoning:opt:approx}
\begin{align}
    &\underset{\{\z_i\}}{\arg\min}\,\Big(\underset{\x_j\in\setD_{\text{target}}}{\arg\min}\; \pnorm{\z_i - \x_j}_p,
    \E[\loss(\classifier_{\setD_{\text{orig}}\cup\setD_{\text{poison}}}(\x_l)), y_l)]\Big)\\
    \text{s.t. }
    &\sum_{\x\in\setD_{\text{target}}}\regularization \circ \myCF{\x}{\classifier_{\setD_{\text{orig}}\cup\setD_{\text{poison}}}} > \sum_{\x\in\setD_{\text{target}}}\regularization \circ \myCF{\x}{\classifier_{\setD_{\text{orig}}}} \\
    & \E_{\x\sim \density'}\left[\regularization \circ \myCF{\x}{\classifier_{\setD_{\text{orig}}\cup\setD_{\text{poison}}}}\right] - \E_{\x\sim \density'}\left[\regularization \circ \myCF{\x}{\classifier_{\setD_{\text{orig}}}}\right] \leq \epsilon\label{eq:datapoisoning:opt:approx_untargeted}
\end{align}
\end{subequations}
where the poisonous data set $\setD_{\text{poison}}$ (\refdef{def:recourse_poisoning_data}) is constructed as $\setD_{\text{poison}}=\{(\z_i,\y)\}$.

Note that the objective in~\refeq{eq:datapoisoning:opt:approx} also covers the plausibility requirement by constructing poisonous instances that are very similar to the given samples $\setD_{\text{target}}$ -- in particular, it was observed~\cite{chakraborty2021survey} that small perturbations often remain unnoticed by the human, which gave rise to adversarial attacks~\cite{chakraborty2021survey,rauber2020foolbox}. The rationale behind this is to make the poisonous instances more difficult to detect -- we empirically evaluate this in an ablation study in Section~\ref{sec:exp:ablation}.

\paragraph{Implementation}
We propose to compute an approximate solution to~\refeq{eq:datapoisoning:opt:approx} by constructing instances $\z_i$ that are on the decision boundary or behind it and are close to samples in $\setD_{\text{target}}$. Note that Theorem~\ref{Theorem:algo:correct:knn} and Theorem~\ref{Theorem:algo:correct:svm} suggest that samples on or behind the decision boundary form a poisonous data set (\refdef{def:recourse_poisoning_data}).
We can construct such instances by computing closest counterfactual explanations $\delta_j$ (\refdef{def:counterfactual}) of samples $(\x_j,\y)\in\setD_{\text{target}}$ that are close to the decision boundary:
\begin{equation}\label{eq:datapoisoning:cf}
    \z_i = \x_j + \delta_j \quad \text{with } \delta_j=\myCF{\x_j}{\classifier}
\end{equation}
As already mentioned, note that the counterfactual $\delta_i$ used in~\refeq{eq:datapoisoning:cf} is \emph{not} necessarily computed by the same counterfactual generation method that is targeted by the data poisoning (\refdef{def:recourse_poisoning_data}) -- i.e., we use an arbitrary and generic counterfactual generation algorithm to poison any other counterfactual generation algorithm.
\begin{remark}\label{remark:poisongalgo:gradients}
Alternatively, \refeq{eq:datapoisoning:cf} could be approximated by a single gradient descent step on how to flip the prediction:
\begin{equation}\label{eq:datapoisoning:grad}
    \z_i = \x_j \underbrace{- \alpha \nabla \loss(\classifier(\x_j), \ycf)}_{:= \delta_j} \quad \text{for some scaling factor $\alpha>0$}
\end{equation}
This approach offers the advantage of avoiding the computation of a counterfactual $\delta_j$, but it comes at the cost of sacrificing guarantees of correctness, both in terms of the poisoning property and the plausibility of the final instances $\z_i$. Additionally, it requires access to the model or at least access to gradients, which are either equivalent to or even stronger assumptions than those needed for~\refeq{eq:datapoisoning:cf}, where a counterfactual $\delta_j$ is computed. In this context, it is worth noting that there exist methods~\cite{mothilal2020explaining} for computing counterfactuals that do not require gradients.

In the remainder of this work, we stick with~\refeq{eq:datapoisoning:cf} for approximately solving~\refeq{eq:datapoisoning:opt:approx} and leave the investigation of~\refeq{eq:datapoisoning:grad} for future work.
\end{remark}

Maintaining the predictive performance objective and not changing the cost of recourse for untargeted instances are both considered implicitly in~\refeq{eq:datapoisoning:cf}. Because the poisonous instances $\z_i$ are close to the targeted instances in $\setD_{\text{target}}$, a sufficiently flexible classifier should not change its behavior in other regions in data space.
Furthermore, because we only consider samples $\x_j$ that are close to the decision boundary, the corresponding $\z_i$ (which constitute a counterfactual instance of $\x_j$) are expected to be very similar to $\x_j$ and therefore satisfy the plausibility requirement -- we empirical evaluate this in an ablation study in Section~\ref{sec:exp:ablation}.

To increase the robustness of the poisoning, we propose to (optionally) not only consider a single closest counterfactual $\delta_j$ in~\refeq{eq:datapoisoning:cf} but a set of $k$ diverse closest counterfactual explanations.
We also propose to extend the counterfactual direction $\delta_j$ by multiplying it with a factor $\alpha>1$, to create a larger and significant increase in the cost of recourse.

The pseudo-code for generating a data poisoning is given in Algorithm~\ref{algo:datapoisoning:fair_cf}.
\begin{algorithm}[t!]
\caption{Data Poisoning for Increasing the Cost of Recourse}\label{algo:datapoisoning:fair_cf}
\textbf{Input:} Samples $\setD_{\text{target}}=\{(\x_i,\y)\}$ from the data space region that is targeted; 
Mechanism $\myCF{\cdot}{\classifier}$ for generating closest counterfactuals; Number $n$ of poisonous instances;
Hyperparameters: $k,b$
\\
\textbf{Output:} Poisoned training data set $D_\text{poisoned}$
\begin{algorithmic}[1]
  \State $\{\delta_i = \regularization \circ \myCF{\x_i}{\classifier}\quad\forall\x_i\in\setD_{\text{target}}\}$ \Comment{Estimate distances to decision boundary}
 \State $D_\text{poison}$ = \{\} 
  \For{$n$-times}
    \State $(\x,\y) \sim \text{weighted\_sampling}\left(\setD_{\text{target}}, \{\delta_i\}\right)$  \Comment{Prefer samples close to the decision boundary}
    \State $\Delta_\text{cf} = \myCF{\x}{\classifier;k}$ \Comment{$k$ diverse closest CFs}
    \For{$\vec{\delta}_\text{cf} \in \Delta_\text{cf}$}
        \For{$\alpha \in [1, b]$}
            \State $\z = \x + \alpha * \vec{\delta}_\text{cf}$   \Comment{Add samples along $\vec{\delta}_\text{cf}$}
            \State $D_\text{poison} = D_\text{poison} \cup \Big\{\left(\z,\y\right)\Big\}$
        \EndFor
    \EndFor
 \EndFor
 \State $D_\text{poisoned} = D_\text{train} \cup D_\text{poison}$  \Comment{Add $D_\text{poison}$ to training set}
\end{algorithmic}
\end{algorithm}

\paragraph{Correctness}
From Theorem~\ref{Theorem:algo:correct:knn} it follows (see Corollary~\ref{corollary:correctness:knn}) that Algorithm~\ref{algo:datapoisoning:fair_cf} computes valid poisonous data sets for a k-NN classifier with $k=1$.
\begin{corollary}[Correctness of Algorithm~\ref{algo:datapoisoning:fair_cf}]\label{corollary:correctness:knn}
    Let $\classifier_{\setD}(\cdot)$ be a k-nearest neighbor classifier (with $k=1$) for some data set $\setD$. For any $\setD_{\text{target}}=\{(\xorig,\yorig)\}$ where $(\xorig,\yorig)\in\setD$, Algorithm~\ref{algo:datapoisoning:fair_cf} computes a poisonous data set (\refdef{def:recourse_poisoning_data}) that increases the cost of recourse of $\xorig$.
\end{corollary}
The detailed proof of Corollary~\ref{corollary:correctness:knn} is given in the appendix.

Apart from the simplistic case of a k-NN classifier with $k=1$, we perform an extensive empirical evaluation in Section~\ref{sec:experiments} to provide evidence for the correctness of Algorithm~\ref{algo:datapoisoning:fair_cf} in more general and realistic cases.
More specifically, in a simplified manner, the theorems from Section~\ref{sec:datapoisoning:theory} state that samples on the decision boundary or behind it can shift the decision boundary such that the cost of recourse changes. Note that this is one of the core ideas behind the proposed data poisoning Algorithm~\ref{algo:datapoisoning:fair_cf}. 
While the overall idea of those theorems is not the only ingredient for Algorithm~\ref{algo:datapoisoning:fair_cf}, it is the only, and therefore necessary, ingredient for influencing the cost of recourse. An additional ingredient for Algorithm~\ref{algo:datapoisoning:fair_cf} is the sampling strategy (see line 4 in Algorithm~\ref{algo:datapoisoning:fair_cf}) for ensuring that the poisonous samples are realistic and therefore difficult to detect. In the ablation study in Section~\ref{sec:exp:ablation}, we remove this sampling strategy and therefore basically reduce Algorithm~\ref{algo:datapoisoning:fair_cf} to the core findings of the theorems from Section~\ref{sec:datapoisoning:theory}. Thereby, we empirically evaluate in how far those theorems generalize to more realistic scenarios.

\paragraph{Runtime}
The runtime of Algorithm~\ref{algo:datapoisoning:fair_cf} can be broken down to $\mathcal{O}(n \cdot k \cdot \rho)$ where $n$ and $k$ are the hyper-parameters of the algorithm referring to the number of poisonous instances (i.e., size of the poisonous data set), and $\rho$ denotes the computational complexity (i.e., runtime) for computing the poisonous sample(s) as constructed in~\refeq{eq:datapoisoning:cf} (see line 8 in Algorithm~\ref{algo:datapoisoning:fair_cf}) -- note $\rho$ is likely to differ between different counterfactual generation mechanisms and also on the complexity of the black-box ML-model to be poisoned.
The recourse cost definition $\regularization(\cdot)$ is not expected to influence the overall runtime of Algorithm~\ref{algo:datapoisoning:fair_cf} as long as it only depends on the dimensionality of the inputs $\x_i$ -- note that this is naturally the case for classic recourse cost implementations such as weighted sums.
Consequently, the runtime of Algorithm~\ref{algo:datapoisoning:fair_cf} scales linearly with the number of requested poisonous samples -- assuming that $\rho$ is a polynomial.

\paragraph{Trade-offs \& Limitations}
The major limitation of Algorithm~\ref{algo:datapoisoning:fair_cf} is that it requires access to a counterfactual generation method for generating poisonous instances. A gradient-based approximation of the counterfactual generation method could be an alternative to this (see Remark~\ref{remark:poisongalgo:gradients}). However, besides assuming gradients and access to them (not possible for tree-based models), one would lose the stated correctness guarantees.

\section{Qualitative Evaluation Case-Study: Explaining Event Detection in Water Distribution Networks}\label{sec:casestudy}
In this illustrative case study, we aim to highlight the impact of manipulated counterfactuals in event diagnosis systems, i.e., systems for detecting events such as anomalies and identifying their cause. Note that event diagnosis constitutes an essential task for the successful operation of critical infrastructure systems such as water networks, power grids, and transportation networks. In the context of AI-based event detectors, performing counterfactual reasoning, via counterfactual explanations, over possible causes of an observed event, is an increasingly popular approach in the literature~\cite{Artelt_Vrachimis_Eliades_Kuhl_Hammer_Polycarpou_2025,Gadekallu_Kumar_Reddyaddikunta_Boopathy_Deepa_Chengoden_Victor_Wang_Wang_Zhu_Dev_2025}.
Note that, although we are focusing on the exemplary case of identifying sensor failures in water distribution networks, our findings can likely be generalized to other domains.

\subsection{Introduction and Background}
Water distribution networks (WDNs) are critical infrastructure for supplying drinking water. Water utilities and human operators depend on strategically placed sensors to identify anomalies such as sensor faults, leaks, and contaminations. Data-driven methods are used to automate sensor data analysis and to support human decision-making. Due to the significant impact of addressing anomalies, such as dispatching repair teams or adjusting water treatment, operators must understand and trust these methods' predictions.
Recent work~\cite{vaquet2024localizing,arteltOne} in this area proposes the usage of XAI to analyze and explain detected anomalies, ensuring human operators can respond optimally.
The authors of~\cite{arteltOne} propose a counterfactual explanation method for explaining anomalous observations (i.e., raised alarms of an event detector) and, in particular, identifying the cause, such as a faulty sensor. To achieve this, they suggest reconstructing sensor readings using an ensemble of virtual sensors $f(\cdot)$. This means creating a virtual sensor for each existing one -- i.e., each sensor's reading is predicted based on the readings from all other sensors. An alarm is raised (i.e., $\classifier(\cdot)=1$) whenever the difference between predicted and observed sensor reading is too large:
\begin{equation}\label{eq:wdn:eventdetection}
    \classifier(\x_t) = \begin{cases}
        1 & \quad \text{if } ||f(\x_t) - \x_t||_p \geq \zeta \\
        0 & \quad \text{otherwise}
    \end{cases}
\end{equation}
where $\x_t$ denotes the sensor readings at time $t$ (or a small time window), and the hyperparameter $\zeta > 0$ denotes a threshold at which an alarm is raised. The threshold $\zeta$ can either be set manually by the human operator or can be calibrated based on a validation set.
They propose using counterfactual explanations to explain the cause of an alarm in the observed sensor readings $\x_t$ (i.e., input to the event detector). Additionally, they propose an efficient algorithm to compute counterfactuals for~\refeq{eq:wdn:eventdetection} by exploiting the ensemble structure of the virtual sensors $f(\cdot)$. Those counterfactual sensor readings $\deltacf$ can be interpreted as ways to ``undo'' the triggered alarm by reverting the effect of events, such as sensor faults, manifested in the counterfactual $\deltacf$, i.e:
\begin{equation}
    \classifier(\x_t + \deltacf) = 0
\end{equation}
The counterfactual $\deltacf$ states significant differences for the faulty sensors (or sensors close to the anomaly) and therefore assists human operators in identifying and locating the anomaly, enabling them to take appropriate actions~\cite{arteltOne}.

\subsection{Setup}
In this case study, we consider the event diagnosis task of identifying and localizing sensor faults in a Water Distribution Network (WDN). In this context, we demonstrate the impact of data poisoning on counterfactuals in such a counterfactual-based event diagnosis method~\cite{arteltOne}, where counterfactual reasoning is used to identify and localize the sensor fault.

We consider the popular Hanoi benchmark from LeakDB~\cite{Vrachimis2018b} to simulate several WDN scenarios (each 21 days long) by randomly introducing several pressure sensor faults, modeled as Gaussian noise (mimicking an aging sensor), at various locations and times. We simulate those scenarios with EPyT-Flow~\cite{Artelt2024} and apply the event detection and counterfactual explanations as proposed in~\cite{arteltOne}. As aforementioned, the core of the system~\refeq{eq:wdn:eventdetection} consists of a set of virtual sensors (based on linear regression) for each of the four pressure sensors -- i.e., predicting the readings at this sensor based on the readings of all other sensors. More details can be found in~\cite{arteltOne} and in the GitHub repository\footnote{\url{https://github.com/andreartelt/DataPoisoningCounterfactuals}} of this paper.
\begin{figure}[t!]
    \begin{subfigure}[b]{0.48\textwidth}
         \includegraphics[width=\textwidth]{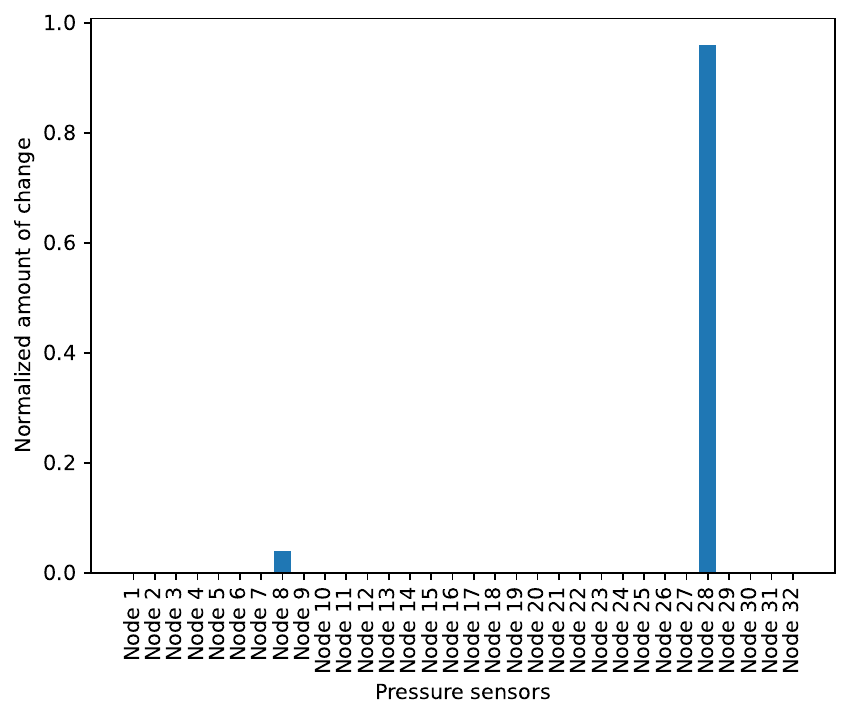}
         \caption{No data poisoning.}
    \end{subfigure}
    \hfill
    \begin{subfigure}[b]{0.48\textwidth}
         \includegraphics[width=\textwidth]{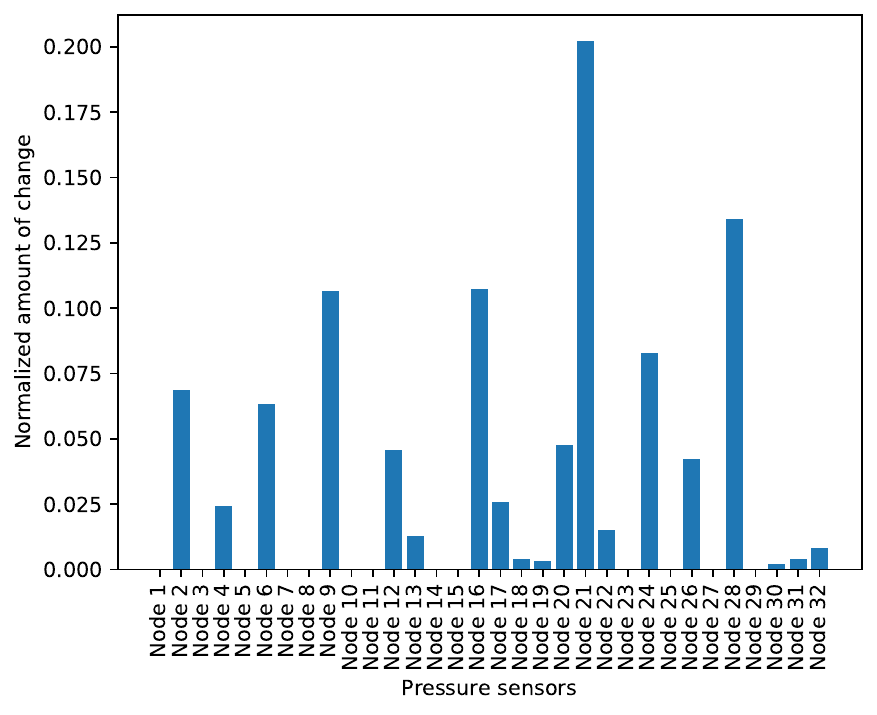}
         \caption{Our proposed data poisoning.}
    \end{subfigure}

    \caption{Sensor fault at node 28 -- original counterfactual vs. poisoned counterfactual.}
    \label{fig:exp:results:wdncasestudy}
\end{figure}
To evaluate the impact of data poisoning and apply our proposed method from Section~\ref{sec:datapoisoning:algo}, we have to specify the cost of a counterfactual in this context. Since counterfactuals are expected to pinpoint the faulty sensor, and each non-zero feature modification indicates a potential sensor fault location requiring manual (and costly) inspection, we propose using the number of modified features (i.e., sparsity) as the measure of recourse cost in this application:
\begin{equation}
    \theta(\deltacf) = \sum_i \mathds{1}_{(\deltacf)_i \neq 0}
\end{equation}
For this case study, we assume a scenario in which the water utility has a database of (simulated) sensor readings that are used to train the event diagnosis systems~\refeq{eq:wdn:eventdetection}. We assume that an attacker has the ability to add samples to this database, and has obtained a copy of the trained event diagnosis system on the original database, but does not know anything about the architecture of the event diagnosis system.
In addition, the attacker has access to a simulator for generating sensor readings data that can be utilized for constructing the poisonous samples. In order to mimic real-world constraints (in particular, the sim-to-real gap), the simulation used for constructing the poisonous samples differs slightly from the one used in the evaluation.
More specifically, we simulate $3$ sets of $3$ scenarios each, whereby each scenario is unique by mimicking real-world uncertainties such as slightly different water demands and slight changes to parameters such as pipe diameter and pipe length:
\begin{enumerate}
    \item The first set does not contain any sensor faults and is used for fitting the virtual sensors and calibrating the event detector~\refeq{eq:wdn:eventdetection} as described in~\cite{arteltOne}.
    \item The second set contains sensor faults (i.e., adding Gaussian noise) at random locations and is used for generating poisonous instances with our proposed Algorithm~\ref{algo:datapoisoning:fair_cf} -- we set $\alpha=1.5$ and use uniform sampling when selecting candidate samples from the target samples $\setD_{target}$. For the target samples $\setD_{target}$, we apply the event detection method~\refeq{eq:wdn:eventdetection}, which was trained on the first set of scenarios, to the newly simulated data and use the true positives as the target samples $\setD_{target}$ in Algorithm~\ref{algo:datapoisoning:fair_cf}.
    For evaluating the impact of the data poisoning, we then add $5\%$ of the generated poisonous data points to the original training set from the first set of scenarios -- this translates to adding approximately $50$ poisonous samples to a training set of size $1000$.
    \item The third set also contains the same type of sensor faults (i.e., adding Gaussian noise) at random locations and is used for evaluation only -- i.e., evaluating the counterfactual explanations of the detected sensor faults with and without data poisoning.
    Note that this set of scenarios differs slightly from the other sets, mimicking the sim-to-real gap, thus ensuring a realistic and fair evaluation.
\end{enumerate}

\subsection{Results \& Conclusion}
\begin{figure}[t!]
    \centering
    \includegraphics[width=0.5\linewidth]{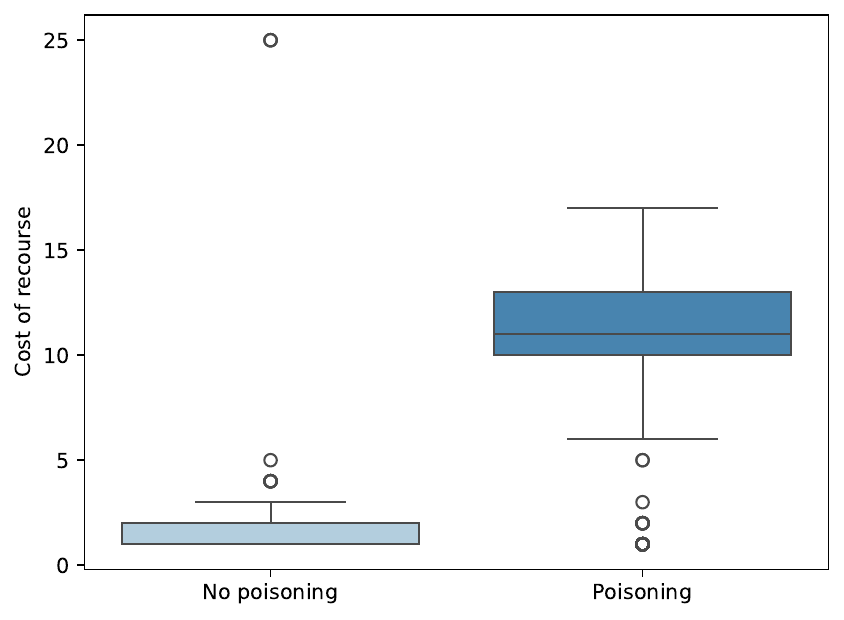}
    \caption{Cost of recourse (i.e. sparsity): Original counterfactuals vs. poisoned counterfactuals -- lower scores are better.}
    \label{fig:exp:results:wdncasestudy-costofrecourse}
\end{figure}

For the evaluation, we only consider true positives -- i.e., a sensor fault was correctly detected and must now be localized. In Figure~\ref{fig:exp:results:wdncasestudy-costofrecourse}, we show the cost of recourse (i.e., the sparsity of the counterfactual) for both the original and poisoned event diagnosis system. Additionally, Figure~\ref{fig:exp:results:wdncasestudy} illustrates a single counterfactual -- comparing the original versus the poisoned event diagnosis system. We observe that our data poisoning approach significantly decreases the sparsity of the counterfactuals, thereby making it more challenging and costly to identify the faulty sensor.

More generally, this case study demonstrates that an attacker can compromise the ability of a counterfactual-based event diagnosis system for identifying the cause of anomalous observations and thereby hindering the detection of malicious activities such as cyber-physical attacks.
Consequently, it highlights the severe impact of data poisoning on counterfactuals in critical real-world applications and stresses the importance of data integrity for building data-driven event diagnosis methods.

\section{Quantitative Evaluation on ML Benchmarks}\label{sec:experiments}
We empirically evaluate the robustness of counterfactual explanations against data poisoning by applying our proposed data poisoning Algorithm~\ref{algo:datapoisoning:fair_cf} on combinations of several different benchmark data sets, classifiers, and state-of-the-art counterfactual explanation generation methods and toolboxes. We consider the following three attack scenarios separately:
\begin{enumerate}
    \item Section~\ref{sec:exp:global}: Increasing the cost of recourse \emph{globally}, i.e. for all individuals.
    \item Section~\ref{sec:exp:subgroup}: Increasing the cost of recourse for a \emph{sub-group} of individuals only.
    \item Section~\ref{sec:exp:local}: Increasing the cost of recourse \emph{locally} for single individuals only.
\end{enumerate}

As an initial baseline, we also empirically evaluate the effect of the label flipping attack, as a classic data poisoning attack, on the cost of recourse (Section~\ref{sec:exp:labelflipping}). This allows us to evaluate the necessity of developing specialized data poisoning methods such as our proposed Algorithm~\ref{algo:datapoisoning:fair_cf}.

We also evaluate the effectiveness of class data sanitization methods for detecting the poisonous instances (see Section~\ref{sec:exp:defense}).

Finally, we also conduct an ablation study (Section~\ref{sec:exp:ablation}) to evaluate the effect of the sampling procedure in Algorithm~\ref{algo:datapoisoning:fair_cf}. In particular, how well the formal statements from Section~\ref{sec:datapoisoning:theory} generalize to more complex and realistic scenarios.

The Python implementation of all conducted experiments (including all data sets) is available on GitHub\footnote{\url{https://github.com/andreartelt/DataPoisoningCounterfactuals}}.

\subsection{Benchmark Data Sets \& Classifiers}
We consider three commonly used data sets from the literature that all contain a sensitive attribute, such that we can also evaluate the difference in the cost of recourse between protected groups in our empirical evaluation:
\begin{itemize}
\item The ``Diabetes'' data set~\cite{efron2004least} (denoted as \textit{Diabetes}) contains data from $442$ diabetes patients, each described by For each patient,
$9$ numeric attributes such as age, body mass index, average blood pressure, and six blood serum measurements
are available -- in addition, the sensitive attribute ``sex'' of each patient is given.
The target for predictions is a binarized quantitative measure of disease progression one year after baseline.
The data set is balanced concerning the class labels ($238$ vs. $204$ samples).

\item The ``Communities \& Crime'' data set~\cite{dheeru2017uci} (denoted as \textit{Crime}) contains $1994$ socio-economic data, including the sensitive attribute ``race'', records from the USA. Following the pre-processing as suggested in~\cite{le2022survey}, we are left with $100$ encoded attributes that are used to predict the crime rate (low vs. high). The data set contains a single sensitive attribute ``race''.
The data set is extremely unbalanced concerning the class labels ($1872$ vs. $122$ samples). We therefore randomly under-sample the majority class in every train-test split.

\item The ``German Credit Data set''~\cite{GermanCreditDataSet} (denoted as \textit{Credit}) is a data set for loan approval and contains $1000$ instances each annotated with $7$ numerical and $13$ categorical attributes, including the sensitive attribute ``sex'', 
with a binary target value (``accept'' or ``reject'').
We use only the seven numerical features.
Because the data set is heavily class-imbalanced ($700$ vs. $300$ samples), we randomly under-sample the majority class in every train-test split.
\end{itemize}
All data sets are standardized to have a mean of zero and unit variance for improved numerical stability; this scaling is done in each train-test split.

\subsection{Machine Learning Classifiers and Counterfactual Generation Methods}
In order to evaluate the broader impact of data poisoning on counterfactuals, we consider a diverse set of ML classifiers $\classifier(\cdot)$ and counterfactual generation methods \& toolboxes.

We consider the following classifiers: $3$-layer neural network with ReLU activation functions (denoted as \textit{DNN}), random forests (denoted as \textit{RNF}), and linear SVMs (denoted as \textit{SVC}).
The hyperparameters of those classifiers have been tuned separately by a grid search and are kept fixed during the data poisoning experiments for better comparability  
-- i.e., retraining on a poisoned training data set is done under the same hyperparameters as on the original/clean data set.

We consider a diverse set of different and popular state-of-the-art methods for computing computational recourse. Note that all of these methods define and compute counterfactual recourse in slightly different ways as discussed in Section~\ref{sec:foundations}:
\begin{itemize}
\item Nearest Unlike Neighbor~\cite{Dasarathy_1995} (denoted as \textit{NUN}), as simple but strong baseline.
\item Counterfactuals guided by Prototypes~\cite{CounterfactualGuidedByPrototypes} (denoted as \textit{Proto}) for computing plausible counterfactuals.
\item DiCE~\cite{mothilal2020explaining} for computing diverse counterfactuals.
\end{itemize}

\subsection{Setup}
In all experiments where we evaluate our proposed data poisoning method Algorithm~\ref{algo:datapoisoning:fair_cf}, we use DiCE~\cite{mothilal2020explaining} as a counterfactual generation mechanism for computing three diverse closest counterfactuals (i.e., $k=3$ in Algorithm~\ref{algo:datapoisoning:fair_cf}) that are as close as possible to the original sample. Furthermore, we set $b=1.5$ (see Algorithm~\ref{algo:datapoisoning:fair_cf}) and use the $\ell_2$-distance to the decision boundary for computing the sample weights in line 4 of Algorithm~\ref{algo:datapoisoning:fair_cf} -- i.e., the probability of the i-th sample scales with $p_i=\frac{1}{\pnorm{\delta_i}_2}$.

All experiments are run in $5$-fold cross-validation.
We use the $\ell_1$ norm as a popular implementation~\cite{guidotti2022counterfactual} of the cost of recourse -- i.e., we set $\regularization(\cdot)=\pnorm{\cdot}_1$ --, and w.l.o.g., we refer to $y=0$ as the unfavorable, and $y=1$ as the favorable outcome.
In all global and sub-group data poisoning scenarios, we inject different amounts (5\% to 70\% of the original training data) of poisonous instances into the training data set -- i.e., original training data + poisoned instances.
Note that we do not make any specific assumptions on the poisoning budget of the attacker, but instead investigate the sensitivity of the effects regarding different data poisoning sizes. Obviously, in practice, smaller poisoning budgets are preferred and probably more realistic -- we therefore pay special attention to the smallest poisoning budget that leads to a statistically significant increase in the cost of recourse.

We not only evaluate the influence of the number of poisonous instances on the cost of recourse, but also their influence on the classifiers' predictive performance. Furthermore, we evaluate the sparsity (i.e., number of changed features) and the plausibility (i.e., the log-likelihood), under a kernel density estimator with a Gaussian kernel, of the generated counterfactual explanations, as well as the time it took to generate them.
In order to analyze the associated uncertainties of all evaluations, we perform a Mann-Whitney U test on all evaluated quantities, assessing the statistical significance of the reported results. Furthermore, all figures also visualize the standard deviation of the reported results.

Note that we did to find any statistically significant effect of data poisoning on the time it takes to generate a counterfactual explanation, and also not in the number of times where the computation of counterfactuals failed.

For better readability, some results are moved to the appendix.

\begin{remark}
Note that, although we use the DiCE method~\cite{mothilal2020explaining} for constructing the poisonous instances in Algorithm~\ref{algo:datapoisoning:fair_cf}, it can still be considered a model-agnostic method because it is also able to attack other counterfactual generation methods -- \textit{i.e., other counterfactual generation methods can be poisoned by utilizing DiCE in constructing the poisonous samples~\refeq{eq:datapoisoning:cf}.}
\end{remark}

\subsubsection{Effect of a label flipping attack on the cost of recourse}\label{sec:exp:labelflipping}
Before empirically evaluating our proposed data poisoning method Algorithm~\ref{algo:datapoisoning:fair_cf} for increasing the cost of recourse, we first evaluate the effect of a label-flipping poisoning attack~\cite{lin2021ml} on the global cost of recourse. By this, we empirically evaluate the difference between "classic" data poisoning for decreasing a model's predictive performance, and our data poisoning method for increasing the cost of recourse. 
Furthermore, note that while label flipping is known to have a strong effect on the attacked model~\cite{lin2021ml}, it is also known to be a somewhat "noisy" data poisoning that can be detected by outlier detection methods~\cite{paudice2019label,jiang2023data}.

For every negative classified sample in the test set, we compute a counterfactual explanation.
We evaluate the global increase in the cost of recourse by computing the difference in the cost of recourse:
\begin{equation}\label{eq:eval:diffcostrecourse}
    \regularization\circ\myCF{\x_i}{\classifier_{\setD_{\text{orig}} \cup \setD_{\text{poison}}}} - \regularization\circ\myCF{\x_i}{\classifier_{\setD_{\text{orig}}}}
    \quad\forall\,\x_i,\y_i\in\setD_{\text{test}},\;\classifier(\x_i) = 0
\end{equation}
A positive score~\refeq{eq:eval:diffcostrecourse}
means an increase in the recourse cost (due to the label flipping), while a negative or near-zero score implies no change or a lower cost of recourse.
To facilitate the interpretation of~\refeq{eq:eval:diffcostrecourse}, we report the percentage difference instead of the absolute difference:
\begin{equation}\label{eq:eval:diffcostrecourse_percentage}
    \frac{\regularization\circ\myCF{\x_i}{\classifier_{\setD_{\text{orig}} \cup \setD_{\text{poison}}}}}{\regularization\circ\myCF{\x_i}{\classifier_{\setD_{\text{orig}}}}} - 1
\end{equation}
We report the median of~\refeq{eq:eval:diffcostrecourse_percentage} together with the statistical significance level.
The results for poisoning $5\%$ of the training data (i.e., flipping their label) are shown in Table~\ref{table:exp:results:labelflipping}.

\begin{table}[t]
\caption{\textit{Effect of a classic label flipping attack data poisoning attack:} Difference (percentage) in the cost of recourse (see~\refeq{eq:eval:diffcostrecourse_percentage}) -- no poisoning vs. \emph{label flipping}. In all cases, $5\%$ of the training data is poisoned by flipping their labels. Positive numbers denote an increase (due to the label flipping) in the cost of recourse, while negative numbers denote the opposite. We report the median (over all folds) rounded to two decimal places, as well as the statistical significance according to the Mann-Whitney U test (ns $\implies$ p-value $> 0.05$; * $\implies$ p-value $\leq 0.05$; ** $\implies$ p-value $\leq 0.01$; *** $\implies$ p-value $\leq 0.001$).
}
\label{table:exp:results:labelflipping}
\centering
\begin{adjustbox}{max width=\textwidth}
\begin{tabular}{cccccc}
 \hline
 Classifier & Data set & NUN & DiCE & Proto \\
 \hline
  \multirow{3}{*}{{SVC}}
  & Credit & $1\%_{\footnotesize ns}$ & $0\%_{\footnotesize ns}$ & $-4\%_{\footnotesize ns}$  \\
  & Diabetes & $0\%_{\footnotesize ns}$ & $0\%_{\footnotesize ns}$ & $-4\%_{\footnotesize ns}$ \\
  & Crime &  $-10\%_{\footnotesize ***}$ & $-9\%_{\footnotesize ***}$ & $-9\%_{\footnotesize **}$\\
 \hline
\multirow{3}{*}{{RNF}}
  & Credit & $0\%_{\footnotesize ns}$ & $0\%_{\footnotesize ns}$ & $-1\%_{\footnotesize ns}$  \\
  & Diabetes & $3\%_{\footnotesize ns}$ & $0\%_{\footnotesize ns}$ & $1\%_{\footnotesize ns}$ \\
  & Crime &  $-9\%_{\footnotesize ***}$ & $-5\%_{\footnotesize ***}$ & $-3\%_{\footnotesize ***}$ \\
 \hline
  \multirow{3}{*}{{DNN}}
  & Credit & $0\%_{\footnotesize ns}$ & $-1\%_{\footnotesize ns}$ & $-4\%_{\footnotesize ns}$  \\
  & Diabetes & $0\%_{\footnotesize ns}$ & $-3\%_{\footnotesize ns}$ & $8\%_{\footnotesize ns}$ \\
  & Crime &  $-8\%_{\footnotesize ***}$ & $-7\%_{\footnotesize ***}$ & $-10\%_{\footnotesize ***}$\\
 \hline
\end{tabular}
\end{adjustbox}
\end{table}

\begin{table}[t]
\caption{Difference (percentage) in the cost of recourse (see~\refeq{eq:eval:diffcostrecourse_percentage}): no poisoning vs. \emph{global poisoning}. In all cases, we add 5\% of the training data as poisonous instances. Positive numbers denote an increase (due to the data poisoning) in the cost of recourse, while negative numbers denote the opposite. We report the median (over all folds) rounded to two decimal places, as well as the statistical significance according to the Mann-Whitney U test (ns $\implies$ p-value $> 0.05$; * $\implies$ p-value $\leq 0.05$; ** $\implies$ p-value $\leq 0.01$; *** $\implies$ p-value $\leq 0.001$).
}
\label{table:exp:results:global}
\centering
\begin{adjustbox}{max width=\textwidth}
\begin{tabular}{ccccc}
 \hline
 Classifier & Data set & NUN & DiCE & Proto\\
 \hline
  \multirow{3}{*}{{SVC}}
  & Credit & $9\%_{\footnotesize *}$ & $8\%_{\footnotesize ***}$ & $7\%_{\footnotesize ns}$ \\
  & Diabetes & $11\%_{\footnotesize ***}$ & $8\%_{\footnotesize ***}$ & $19\%_{\footnotesize **}$ \\
  & Crime &  $8\%_{\footnotesize ***}$ & $5\%_{\footnotesize ***}$ & $5\%_{\footnotesize ***}$ \\
 \hline
\multirow{3}{*}{{RNF}}
  & Credit & $3\%_{\footnotesize ns}$ & $6\%_{\footnotesize ***}$ & $11\%_{\footnotesize *}$  \\
  & Diabetes & $2\%_{\footnotesize ns}$ & $1\%_{\footnotesize ns}$ & $14\%_{\footnotesize ns}$  \\
  & Crime & $2\%_{\footnotesize ns}$ & $2\%_{\footnotesize ns}$ & $3\%_{\footnotesize ns}$  \\
 \hline
  \multirow{3}{*}{{DNN}}
  & Credit & $2\%_{\footnotesize ns}$ & $7\%_{\footnotesize ***}$ & $1\%_{\footnotesize *}$ \\
  & Diabetes & $4\%_{\footnotesize ns}$ & $7\%_{\footnotesize **}$ & $8\%_{\footnotesize ns}$ \\
  & Crime &  $4\%_{\footnotesize ***}$ & $6\%_{\footnotesize ***}$ & $4\%_{\footnotesize **}$ \\
 \hline
\end{tabular}
\end{adjustbox}
\end{table}

\subsubsection{Data poisoning for increasing the cost of recourse on a global level}\label{sec:exp:global}
We apply our proposed data poisoning method, Algorithm~\ref{algo:datapoisoning:fair_cf}, on a global level.
For every negative classified sample in the test set, we compute a counterfactual explanation, and evaluate the global effect on the cost of recourse, by computing the difference in the cost of recourse~\refeq{eq:eval:diffcostrecourse_percentage}.
Again, a positive score~\refeq{eq:eval:diffcostrecourse_percentage}
refers to an increase in the recourse cost (due to the data poisoning), while a negative or near-zero score implies no change or a lower cost of recourse. We report the median of~\refeq{eq:eval:diffcostrecourse_percentage} together with the statistical significance according to the Mann-Whitney U test.
The effect of adding $5\%$ of poisonous instances to the training data set on the cost of recourse is shown in Table~\ref{table:exp:results:global}. The effects on the sparsity and log-likelihood, as well as the impact of different poisoning budgets, are given in~\ref{appendix:global}.

\subsubsection{Data poisoning for increasing the cost of recourse on a sub-group level}\label{sec:exp:subgroup}
We consider sub-groups created based on the sensitive attribute -- note that this is a reasonable but only one out of many possible ways how sub-groups might be created.
We apply the Algorithm~\ref{algo:datapoisoning:fair_cf} to poison instances from one protected group only, assuming that the sensitive attribute of each instance is known. By this, we aim to increase the difference in the cost of recourse between the two protected groups, which can be interpreted as introducing or increasing group-unfairness in recourse~\cite{artelt2023ijcai,von2022fairness,sharma2021fair}.

For every negative classified sample in the test set (no matter to which sub-group it belongs), we compute a counterfactual.
We evaluate the difference in the cost of recourse between the two sub-groups as follows:
\begin{equation}\label{eq:eval:difffairness}
\begin{split}
    &\underbrace{\pnorm{\regularization\circ\myCF{\x_i|s=0}{\classifier_{\setD_{\text{orig}} \cup \setD_{\text{poison}}}} - \regularization\circ\myCF{\x_i|s=1}{\classifier_{\setD_{\text{orig}} \cup \setD_{\text{poison}}}}}}_{\text{Median difference in the cost of recourse \textbf{under} data poisoning}}
    -\\& \underbrace{\pnorm{\regularization\circ\myCF{\x_i|s=0}{\classifier_{\setD_{\text{orig}}}} - \regularization\circ\myCF{\x_i|s=1}{\classifier_{\setD_{\text{orig}}}}}}_{\text{Median difference in the cost of recourse \textbf{without} data poisoning}} \quad \forall\,\x_i\in\setD_{\text{test}}\;\classifier(\x_i) = 0
\end{split}
\end{equation}
where we denote the sensitive attribute as $s$ -- i.e.\ $\x_i|s=0$ means that we only consider $x_i$ if its sensitive attribute is equal to zero.
A positive score of~\refeq{eq:eval:difffairness} refers to an increase in the difference of the cost of recourse between the protected groups, while a negative score refers to the opposite. 
Note that for the purpose of stability, we use the median (over all folds) in~\refeq{eq:eval:difffairness}.
To facilitate the interpretability of the results, we report the percentage difference instead of the absolute difference:
\begin{equation}\label{eq:eval:difffairness_percentage}
    \frac{\pnorm{\regularization\circ\myCF{\x_i|s=0}{\classifier_{\setD_{\text{orig}} \cup \setD_{\text{poison}}}} - \regularization\circ\myCF{\x_i|s=1}{\classifier_{\setD_{\text{orig}} \cup \setD_{\text{poison}}}}}}{\pnorm{\regularization\circ\myCF{\x_i|s=0}{\classifier_{\setD_{\text{orig}}}} - \regularization\circ\myCF{\x_i|s=1}{\classifier_{\setD_{\text{orig}}}}}} - 1
\end{equation}
We report the results regarding~\refeq{eq:eval:difffairness_percentage}, predictive performance, and sparsity, together with the statistical significance level in~\ref{appendix:subgroup}.

\subsubsection{Data poisoning for increasing the cost of recourse on a local level}\label{sec:exp:local}
\begin{table}[t!]
\centering
 \caption{Difference (percentage) in the cost of recourse (see~\refeq{eq:eval:diffcostrecourse_percentage}): no vs. local poisoning. Positive numbers denote an increase (due to the data poisoning) in the cost of recourse. We report the median (over all folds) rounded to two decimal places, as well as the statistical significance according to the Mann-Whitney U test (ns $\implies$ p-value $> 0.05$; * $\implies$ p-value $\leq 0.05$; ** $\implies$ p-value $\leq 0.01$; *** $\implies$ p-value $\leq 0.001$).}
\begin{adjustbox}{max width=\textwidth}
\begin{tabular}{ccccc}
 \hline
 Classifier & Data set & NUN & DiCE  & Proto  \\
 \hline
  DNN & Diabetes & $39\%_{\footnotesize ***}$ & $27\%_{\footnotesize ***}$ & $72\%_{\footnotesize ***}$ \\
  \hline
 \end{tabular}
\end{adjustbox}
 \label{table:exp:results:local}
\end{table}

We compute a local data poisoning (with our proposed Algorithm~\ref{algo:datapoisoning:fair_cf}) for every negative classified sample in the test set.
However, because of computational limitations -- i.e., for every sample in the test set (over all folds), the entire data poisoning must be run and evaluated --, we only evaluate a single scenario considering a DNN classifier applied to the diabetes data set.
The results, together with the statistical significance level, are shown in Table~\ref{table:exp:results:local}, and box-plots of the distributions of the scores can be found in~\ref{appendix:local}.

\subsubsection{Detection of poisonous instances for increasing the cost of recourse}\label{sec:exp:defense}
Given the potentially severe impact of data poisoning attacks on counterfactual explanations, as illustrated in our case study in Section~\ref{sec:casestudy}, we also evaluate the effectiveness of data sanitization procedures~\cite{Paudice_Munoz-Gonzalez_Lupu_2019,Koh_Steinhardt_Liang_2021} for detecting such poisonous instances (\refdef{def:recourse_poisoning_data}). Recall that data sanitization methods~\cite{Paudice_Munoz-Gonzalez_Lupu_2019,Koh_Steinhardt_Liang_2021,Cretu_Stavrou_Locasto_Stolfo_Keromytis_2008,Paudice_Munoz-Gonzalez_Gyorgy_Lupu_2018} aim to detect the poisonous instances by means of outlier detection methods and remove those detected instances from the poisoned training data set $\setD_{\text{orig}} \cup \setD_{\text{poison}}$.

In this context, we evaluate the effectiveness of two classic outlier detection methods, and three classic data sanitization defense methods from the literature~\cite{Koh_Steinhardt_Liang_2021}:
\begin{itemize}
    \item We consider the Isolation Forest~\cite{liu2008isolation} and Local Outlier Factor (LOF) method~\cite{breunig2000lof}, as two classic and popular outlier detection methods. Both methods are calibrated on the (unpoisoned) test set and applied to $\setD_{\text{orig}} \cup \setD_{\text{poison}}$ to detect the poisonous instances $\setD_{\text{poison}}$.
    \item The k-NN defense~\cite{Frederickson_Moore_Dawson_Polikar_2018}, which flags points that are far from their k-th nearest neighbor:
        \begin{equation}\label{eq:defense:knn}
            \pnorm{\x_i - \vec{z}_{\y}}_2 \geq \nu \text{ with $\vec{z}_{\y}$ being the k-th nearest neighbor in $\setD_{\text{orig}} \cup \setD_{\text{poison}}$}
        \end{equation}
        where the hyperparameter $\nu > 0$ denotes the threshold.
    \item The $\ell_2$-defense~\cite{Koh_Steinhardt_Liang_2021}, which flags instances far from their class centroids in the $\ell_2$ distance:
        \begin{equation}\label{eq:defense:l2}
            \pnorm{\x_i - \E[\x\mid \y]}_2 \geq \nu
        \end{equation}
        where the hyperparameter $\nu > 0$ denotes the threshold.
    \item The slab-defense~\cite{DBLP:conf/nips/SteinhardtKL17}, which constitutes an extension of the $\ell_2$-defense~\cite{Koh_Steinhardt_Liang_2021}, flags instances that are too far from the centroids after they are projected onto the line between the two class centroids:
        \begin{equation}\label{eq:defense:slab}
            \left|(\E[\x\mid \y=0] - \E[\x\mid \y=1])^\top(\x_i - \E[\x\mid \y])\right)\geq \nu
        \end{equation}
        where the hyperparameter $\nu > 0$ denotes the threshold.
\end{itemize}
We empirically evaluate the performance of those defense methods in a global attack scenario, where the data poisoning aims to increase the cost of recourse for all individuals. 
For the threshold-based defense methods~\refeq{eq:defense:knn} - \refeq{eq:defense:slab}, we calibrate the threshold on an unpoisoned hold-out set~\cite{Koh_Steinhardt_Liang_2021}, and for the k-NN defense~\refeq{eq:defense:knn}, we set $k=5$ as suggested in~\cite{Koh_Steinhardt_Liang_2021}.

We evaluate the recall for detecting the poisonous instance under different sizes of poisonings, and report the mean and standard deviation.
The results for the Isolation Forest method on a global poisoning are shown in Figure~\ref{fig:exp:results:outlier}, all other results are given in~\ref{appendix:detection}.
\begin{figure}[!t]
    \begin{subfigure}[b]{0.3\textwidth}
         \includegraphics[width=\textwidth]{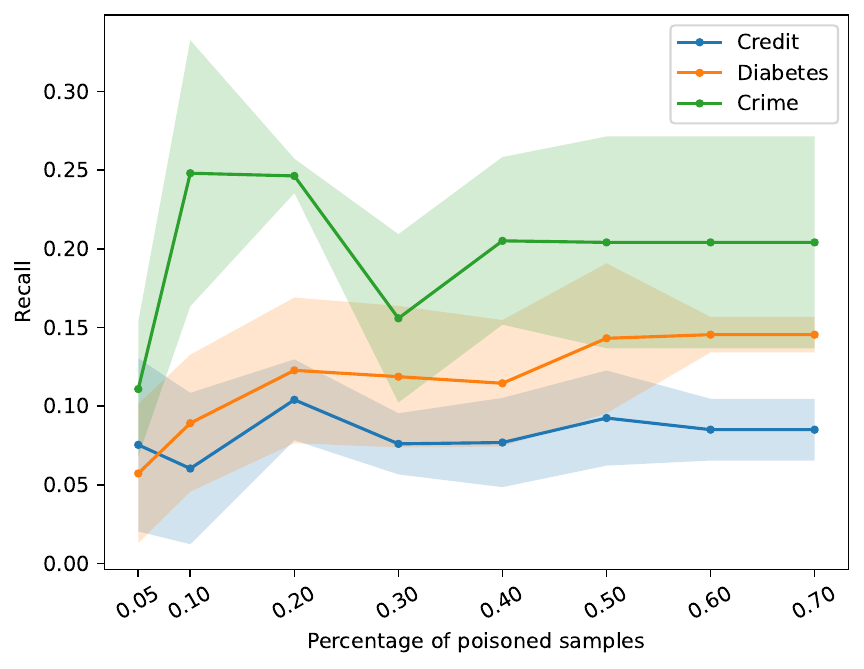}
         \caption{$\classifier(\cdot)$: SVC}
    \end{subfigure}
    \hfill
    \begin{subfigure}[b]{0.3\textwidth}
         \includegraphics[width=\textwidth]{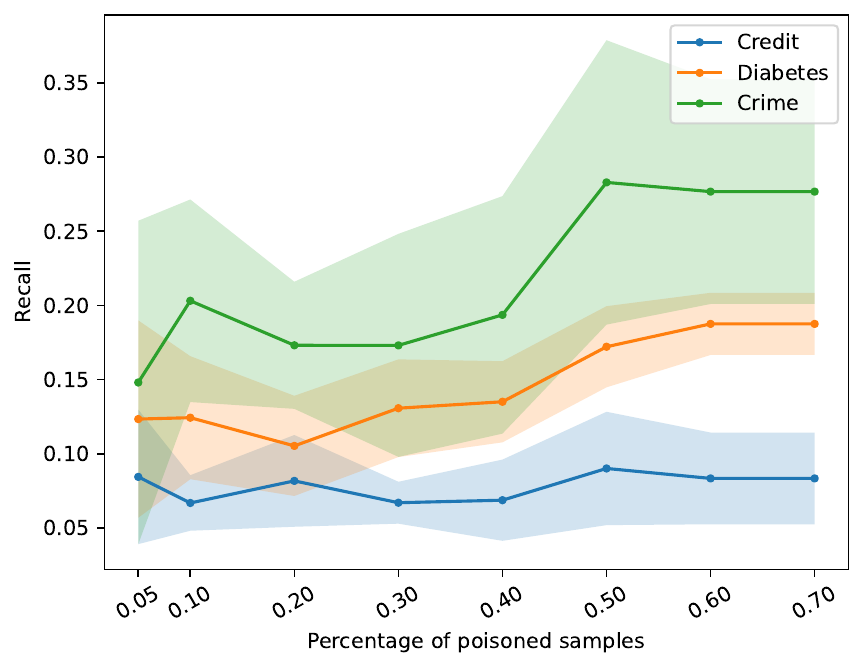}
         \caption{$\classifier(\cdot)$: RandomForest}
    \end{subfigure}
    \hfill
    \begin{subfigure}[b]{0.3\textwidth}
         \includegraphics[width=\textwidth]{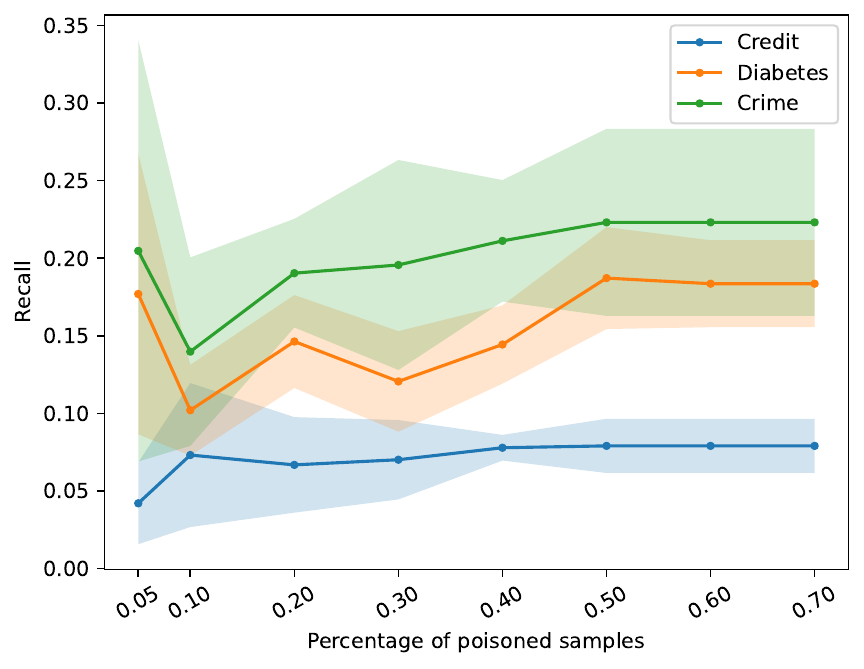}
         \caption{$\classifier(\cdot)$: DNN}
    \end{subfigure}

    \caption{Isolation Forest for detection of the poisonous instances in a \text{global poisoning}. We report the mean and standard deviation (over all folds) of the recall (larger numbers are better).}
    \label{fig:exp:results:outlier}
\end{figure}

\subsubsection{Ablation study}\label{sec:exp:ablation}
In this ablation study, we consider the case of a global poisoning attack on the cost of recourse. We investigate the effect of the sampling strategy in line 4 of Algorithm~\ref{algo:datapoisoning:fair_cf}, which prefers samples close to the decision boundary for generating poisonous samples.
We change the sampling strategy to a uniform sampling and evaluate its effect on the cost of recourse, as well as on the performance of the defense methods for identifying the poisonous samples.

The results for the effect on the cost of recourse are shown in Table~\ref{table:exp:results:ablation:costrecourse}. Additional results for all poisoning budgets can be found in Figure~\ref{appendix:fig:exp:results:ablation:recourse_vs_poisonsamples-1}.
Furthermore, the results of the performance of the outlier detection and data sanitization methods for identifying the poisonous samples can be found in Table~\ref{table:exp:results:ablation:defense} -- here, we compare the performance (i.e., recall) to the case with the original sampling strategy in line 4 of Algorithm~\ref{algo:datapoisoning:fair_cf} that is supposed to ensure plausibility of the poisonous samples.

\begin{table}[t]
\caption{ \emph{Ablation study (uniform sampling)} -- Difference (percentage) in the cost of recourse (see~\refeq{eq:eval:diffcostrecourse_percentage}): no poisoning vs. \emph{global poisoning}. In all cases, we add 5\% of the training data as poisonous instances. Positive numbers denote an increase (due to the data poisoning) in the cost of recourse, while negative numbers denote the opposite. We report the median (over all folds) rounded to two decimal places, as well as the statistical significance according to the Mann-Whitney U test (ns $\implies$ p-value $> 0.05$; * $\implies$ p-value $\leq 0.05$; ** $\implies$ p-value $\leq 0.01$; *** $\implies$ p-value $\leq 0.001$).
}
\label{table:exp:results:ablation:costrecourse}
\centering
\begin{adjustbox}{max width=\textwidth}
\begin{tabular}{ccccc}
 \hline
 Classifier & Data set & NUN & DiCE & Proto\\
 \hline
  \multirow{3}{*}{{SVC}}

  & Credit  & $4\%_{\footnotesize ns}$ & $7\%_{\footnotesize ***}$ & $5\%_{\footnotesize *}$ \\
  & Diabetes& $11\%_{\footnotesize ***}$ & $7\%_{\footnotesize ***}$ & $13\%_{\footnotesize *}$ \\
  & Crime & $7\%_{\footnotesize ***}$ & $7\%_{\footnotesize ***}$ & $7\%_{\footnotesize ***}$ \\
 \hline
\multirow{3}{*}{{RNF}}

  & Credit & $1\%_{\footnotesize ns}$ & $10\%_{\footnotesize ***}$ & $9\%_{\footnotesize ns}$ \\
  & Diabetes & $2\%_{\footnotesize ns}$ & $3\%_{\footnotesize *}$ & $10\%_{\footnotesize ns}$ \\
  & Crime  & $5\%_{\footnotesize *}$ & $1\%_{\footnotesize ns}$ & $13\%_{\footnotesize **}$ \\
 \hline
  \multirow{3}{*}{{DNN}}

  & Credit & $1\%_{\footnotesize ns}$ & $6\%_{\footnotesize ***}$ & $2\%_{\footnotesize ns}$ \\
  & Diabetes & $8\%_{\footnotesize **}$ & $7\%_{\footnotesize ***}$ & $4\%_{\footnotesize ns}$ \\
  & Crime & $6\%_{\footnotesize ***}$ & $6\%_{\footnotesize ***}$ & $10\%_{\footnotesize ***}$ \\
 \hline
\end{tabular}
\end{adjustbox}
\end{table}

\begin{table}[t]
\caption{ \emph{Ablation study (uniform sampling)} -- Difference (percentage) in the recall of detecting the poisonous samples in a global data poisoning scenario. In all cases, we add 5\% of the training data as poisonous instances. Positive numbers denote an increase (due to the uniform sampling) in the cost of recourse, while negative numbers denote the opposite. We report the difference in the median (over all folds) rounded to two decimal places, as well as the statistical significance according to the Mann-Whitney U test (ns $\implies$ p-value $> 0.05$; * $\implies$ p-value $\leq 0.05$; ** $\implies$ p-value $\leq 0.01$; *** $\implies$ p-value $\leq 0.001$).
}
\label{table:exp:results:ablation:defense}
\centering
\begin{adjustbox}{max width=\textwidth}
\begin{tabular}{ccccccc}
 \hline
 Classifier & Data set & Isolation-Forest & LOF & slab-defense & $\ell_2$-defense &  k-NN defense\\
 \hline
  \multirow{3}{*}{{SVC}}

  & Credit  & $82.14\%_{\footnotesize ns}$ & $0\%_{\footnotesize ns}$ & $-31.55\%_{\footnotesize ns}$ & $92.86\%_{\footnotesize ns}$ & $-14.29\%_{\footnotesize ns}$ \\
  & Diabetes & $369.74\%_{\footnotesize **}$ & $0\%_{\footnotesize ns}$ & $123.68\%_{\footnotesize ns}$ & $73.68\%_{\footnotesize ns}$ & $1.05\%_{\footnotesize ns}$ \\
  & Crime & $172.73\%_{\footnotesize **}$ & $-18.18\%_{\footnotesize ns}$ & $0\%_{\footnotesize ns}$ & $13.64\%_{\footnotesize ns}$ & $83.64\%_{\footnotesize **}$ \\
 \hline
\multirow{3}{*}{{RNF}}

  & Credit  & $49.43\%_{\footnotesize ns}$ & $0\%_{\footnotesize ns}$ & $-61.2\%_{\footnotesize ns}$ & $117.46\%_{\footnotesize ns}$ & $28.57\%_{\footnotesize ns}$ \\
  & Diabetes & $215.79\%_{\footnotesize ***}$ & $0\%_{\footnotesize ns}$ & $236.84\%_{\footnotesize *}$ & $57.89\%_{\footnotesize ns}$ & $-12.28\%_{\footnotesize ns}$ \\
  & Crime  & $221.14\%_{\footnotesize **}$ & $0\%_{\footnotesize ns}$ & $0\%_{\footnotesize ns}$ & $13.64\%_{\footnotesize ns}$ & $54.55\%_{\footnotesize **}$ \\
 \hline
  \multirow{3}{*}{{DNN}}

  & Credit  & $200.0\%_{\footnotesize *}$ & $0\%_{\footnotesize ns}$ & $-57.14\%_{\footnotesize ns}$ & $45.42\%_{\footnotesize ns}$ & $157.14\%_{\footnotesize ns}$ \\
  & Diabetes  & $135.79\%_{\footnotesize *}$ & $0\%_{\footnotesize ns}$ & $0\%_{\footnotesize ns}$ & $68.42\%_{\footnotesize *}$ & $0.48\%_{\footnotesize ns}$ \\
  & Crime  & $104.55\%_{\footnotesize ns}$ & $36.36\%_{\footnotesize ns}$ & $0\%_{\footnotesize ns}$ & $0\%_{\footnotesize ns}$ & $-69.7\%_{\footnotesize ns}$ \\
 \hline
\end{tabular}
\end{adjustbox}
\end{table}

\subsection{Results \& Discussion}
\subsubsection{Effect of label flipping on the cost of recourse}
We observe (see Table~\ref{table:exp:results:labelflipping}) that label flipping typically has little to no statistically significant impact on the overall cost of recourse (on a global level). This suggests that traditional data poisoning techniques aimed at reducing predictive performance are inadequate for poisoning counterfactual explanations and increasing the cost of recourse. Consequently, specialized methods and algorithms, such as our proposed data poisoning method, are necessary to affect methods for computing counterfactual explanations.

\subsubsection{General trend}
We observe that in most scenarios, on local as well as on global levels (see Table~\ref{table:exp:results:global} and \ref{appendix:experiments}), even a relatively small amount of poisonous instances such as $5\%$, added to the training data set, leads to a statistically significant increase in the cost of recourse. Increasing the number of poisonous instances leads to an even larger and more significant increase in the cost of recourse. 
Besides an increase in the cost of recourse, we also almost always observe a statistically significant increase in the number of changed features -- i.e., the sparsity of the counterfactuals decreases. Although the increase is statistically significant, it is much smaller than the increase in the cost of recourse.
We do not observe any statistically significant effect on the plausibility (i.e., log-likelihood) of the counterfactual explanation in any of the conducted empirical evaluations (e.g., see Table~\ref{table:exp:results:global_density}). This demonstrates that our proposed data poisoning only affects the cost of recourse and related measures, such as the sparsity, but not the plausibility of the generated counterfactuals. This makes the effects of the data poisoning even more refined, since those poisoned counterfactuals do not look more/less plausible than their non-poisoned counterpart.

However, we also observe differences in the necessary amount of poisonous instances between different counterfactual generation methods and toolboxes -- in particular for the case where we want to increase the cost of recourse on a sub-group level. For the counterfactuals guided by prototypes method~\cite{CounterfactualGuidedByPrototypes} and the nearest unlike neighbor method~\cite{Dasarathy_1995}, we often need more poisonous instances to observe a statistically significant increase in the cost of recourse. Since those methods focus on plausibility,
this might be an indicator that additional plausibility constraints can act as a beneficial regularization for increased stability -- similar to what is reported in~\cite{artelt2021evaluating} for robustness concerning input perturbations. However, as it becomes apparent from the sensitivity analysis regarding the poisoning budget (e.g., see Figure~\ref{appendix:fig:exp:results:recourse_vs_poisonsamples-1}), those two methods for generating counterfactuals are not immune to data poisoning in general, but only more robust to very small amounts of poisonous samples.

Altogether, the empirical evaluation reveals the vulnerability of existing (state-of-the-art) counterfactual generation methods to data poisonings.
Furthermore (as discussed in detail in Section~\ref{sec:results:defense}), the failure of classic outlier detection and defense methods demonstrates that the detection of our generated poisonous instances is non-trivial.

\subsubsection{Data poisoning on a global level}\label{sec:results:global}
In the case of a data poisoning on a global level, we observe an almost monotonic increase in the cost of recourse when increasing the number of poisonous samples. In particular, a small poisoning budget is often already sufficient to create a statistically significant increase in the cost of recourse.
However, we also observe that the increase in the cost of recourse stops at some points and flattens out. This indicates a saturating effect that, at some point, increasing the poisoning budget does not have any further impact on the cost of recourse. More specifically, this suggests that there is an upper bound on the increase in the cost of recourse that can be achieved by our proposed data poisoning.
A similar effect is observed for the decrease in predictive performance of the classifiers and the decrease in sparsity of the generated counterfactuals.

These observations can be explained by the nature of our proposed data poisoning method Algorithm~\ref{algo:datapoisoning:fair_cf}. Since we require that the generated poisonous training samples are similar to the original training samples of the same class (see~\refeq{eq:datapoisoning:opt:approx}), there is an upper bound on how much those poisonous samples can shift the decision boundary of the classifier, leading to the observed results.

\subsubsection{Data poisoning on a sub-group level}
In the case of sub-groups, we observe (see Tables~\ref{table:exp:results:subgroup_mem},\ref{table:exp:results:subgroup_dice},\ref{table:exp:results:subgroup_proto}) similar effects as in the case of a global poisoning.
However, the effects are more unstable in the sense that the increase in the difference of the cost of recourse varies significantly, and sometimes the change in difference is not statistically significant, in particular for the NUN and Proto methods. This is quite likely due to a strong overlap of the distributions of the sub-groups, which makes it difficult to just alter the cost of recourse for one group but not for the other.

\subsubsection{Data poisoning on a local level}
From Figure~\ref{appendix:fig:expresults:plots_local}, we observe that in all cases the local data poisoning attack leads to a statistically significant increase in the cost of recourse for the targeted instances. However, we also observe that the cost of recourse for untargeted instances also increases -- only a small increase compared to the targeted instance, but the difference is already statistically significant.
This demonstrates that our data poisoning method is also able to target specific instances only, without affecting other instances too much. While there is room for improvement, we suspect that the degree of how much untargeted instances are affected depends not only on the flexibility of the attacked classifier but also on the location of the targeted instance in data space. We leave a deeper investigation of such local attacks as future research.

\subsubsection{Effect on the predictive performance}
We observe the expected results that classifiers' predictive performance decreases (statistically significantly) as more poisoned instances are added.
More specifically, for a global data poisoning, the decrease in predictive performance is worse than for sub-group or local data poisonings.

This is to be expected since the manipulation of counterfactual explanations requires manipulating the decision boundary. Despite this, the proposed data poisoning presents a significant threat. Minor drops in predictive performance, resulting from small data poisonings, might go undetected. Yet, they can already lead to a substantial increase in the cost of recourse, as demonstrated in the presented experiments.

\subsubsection{Detection of poisonous instances}\label{sec:results:defense}
Concerning the detectability of the generated poisonous instances, we observe (see Figure~\ref{fig:defenses:results_all}) that all evaluated defense methods for outlier detection struggle to identify the poisonous samples and distinguish them from the original training samples. The performance (i.e., recall) varies slightly between different combinations of classifier, data set, and defense method. However, it is almost always well below $30\%$ and often even below $20\%$, implying that the vast majority of the poisonous samples remain undetected.

These findings demonstrate that our proposed data poisoning method successfully generates poisonous samples that are on the data manifold and difficult to distinguish from normal training samples. This implies that the detection of our generated poisonous instances is non-trivial and requires substantial research efforts.

\subsubsection{Ablation study}\label{sec:results:ablation}
From the results (see Figure~\ref{appendix:fig:exp:results:ablation:recourse_vs_poisonsamples-1}), we observe the same overall effects as we did in the case of a data poisoning of a global level without any modifications to Algorithm~\ref{algo:datapoisoning:fair_cf} (see Section~\ref{sec:results:global}). We therefore conclude that the sampling strategy does not have an impact on the algorithm's ability to increase the cost of recourse. Most importantly, it provides empirical evidence that the findings of the theorems (from Section~\ref{sec:datapoisoning:theory}) indeed generalize to more complex and realistic scenarios.

From Table~\ref{table:exp:results:ablation:defense}, we observe that in most cases, the sampling strategy (line 4 in Algorithm~\ref{algo:datapoisoning:fair_cf}) does not have a statistically significant effect on the performance of the defense methods. Furthermore, we often observe large differences in the median, but these are still not statistically significant due to the high variance (also see Figure~\ref{fig:defenses:results_all}).
However, in the case of the isolation forest (a classic outlier detection method), despite large variances, the differences are almost always statistically significant and indicate a significant increase in detection performance. This indicates that our proposed sampling strategy for ensuring the plausibility of the poisonous samples can make the detection of such samples more difficult, depending on the detection method used.

\section{Conclusion \& Summary}\label{sec:conclusion}
In this work, we examined the resilience of counterfactual explanations against data poisoning. To achieve this, we identified and formalized strategies for data poisoning aimed at increasing the cost of recourse on various levels (local vs. global) by injecting poisonous instances into the training data set. We conducted empirical evaluations to assess the impact of data poisoning across several classifiers, benchmark datasets, and various popular and state-of-the-art counterfactual generation methods. Our findings revealed that even the injection of a small number of poisonous instances into the training data set significantly increases the cost of recourse at all levels -- on a global level as well as on a local level.
Furthermore, we empirically evaluated the effect of a classic data poisoning attack (label flipping), designed to decrease the predictive performance, on the cost of recourse. It turns out that there is no or only a minor effect on the cost of recourse, highlighting the necessity of custom and specialized data poisoning methods such as our proposed Algorithm~\ref{algo:datapoisoning:fair_cf}.

\textbf{Ethical implications and broader impact:}
The presented findings reveal how easily existing classifiers and state-of-the-art counterfactual generation methods and toolboxes can be deceived through manipulation of the training data.
This could significantly undermine users' trust in this XAI method.
More specifically, the severe and wide implications of our study become apparent by demonstrating the vulnerability of explanations to manipulations in sensitive domains such as finance, health, and critical infrastructure, together with the potentially serious consequences of poisoned explanations, such as denying fair recourse, obscuring biases, and compromising system safety. Consequently, our study highlights the urgent need for more robust counterfactual generation methods, toolboxes, and defense strategies against malicious data manipulations. In particular, we suggest not using counterfactual explanations in critical applications if the integrity of the training data cannot be guaranteed.

\textbf{Future research directions:}
As our paper demonstrates the vulnerability of counterfactual explanations to data poisoning, we call for the community to focus on designing inherently more robust algorithms for computing counterfactual explanations to promote the safe deployment of counterfactuals in practice. In this context, we suggest not only focusing on a single evaluation metric, such as the cost of recourse, but also investigating and ensuring robustness to other aspects, such as fairness. Furthermore, we propose to research the nature of such poisonous training samples to gain knowledge for the development of more advanced defense strategies. In particular, there is an urgent need for a better understanding of how and what properties of certain training samples affect methods for computing counterfactuals. In addition, the demonstrated failure of traditional outlier-based defense methods, such as data sanitization methods, highlights the need for novel defense strategies. In this context, concepts from data valuation might constitute promising avenues worth exploring.

We leave those aspects as future research.

\section* {Acknowledgments}
This research was supported by the Ministry of Culture and Science NRW (Germany) as part of the Lamarr Fellow Network. This publication reflects the views of the authors only.\\
\textbf{Disclaimer} 
This paper was prepared for informational purposes by the Artificial Intelligence Research group of JPMorgan Chase \& Co. and its affiliates (``JP Morgan''), and is not a product of the Research Department of JP Morgan. JP Morgan makes no representation and warranty whatsoever and disclaims all liability, for the completeness, accuracy or reliability of the information contained herein. This document is not intended as investment research or investment advice, or a recommendation, offer or solicitation for the purchase or sale of any security, financial instrument, financial product or service, or to be used in any way for evaluating the merits of participating in any transaction, and shall not constitute a solicitation under any jurisdiction or to any person, if such solicitation under such jurisdiction or to such person would be unlawful.

 \appendix

\section{Proofs}\label{appendix:proofs}
\subsection{Proof of Theorem 1}
\begin{proof}
Sketch: For any $\xorig, \classifier(\xorig)=\yorig$, assume uniqueness of the solution $\x'$ -- i.e. the closest sample to $\xorig$ on the decision boundary:
\begin{equation}\label{eq:knn}
\begin{split}
    &\underset{\x'\in\RN^\dimsym}{\arg\min}\,\pnorm{\x' - \xorig}_p \text{ s.t. } \\&\exists i\neq j: (\x_i,\y_i),(\x_j,\y_j)\in\setD, \y_i \neq \y_j, \text{ 
with }\pnorm{\x'-\x_i}_p=\pnorm{\x'-\x_j}_p
\end{split}
\end{equation}
where we (w.l.o.g.) assume the use of the p-norm as the distance function in the 1-NUN neighbor classifier.

Adding $(\x',\yorig)$ to the training data $\setD$ implies that $\x'$ is no longer the solution to~\refeq{eq:knn}.
Therefore, the new closest sample on the decision boundary must have a larger distance to $\xorig$ than $\x'$, otherwise it would have been $\x'$ before!
\end{proof}

\subsection{Proof of Theorem 2}
\begin{proof}
Sketch:
From the triangle-inequality and $\lambda > \pnorm{\x_i-\x_j}_2$ it follows that:
\begin{equation}\label{eq:bound1}
    \pnorm{\x_i-\x_j}_2 + \delta_j' \geq \underbrace{\delta_i + \lambda}_{\delta_i'} \quad \leftrightarrow\quad \delta_j' \geq \delta_i + \lambda - \pnorm{\x_i-\x_j}_2
\end{equation}
Because of $\delta_j>\delta_i$, we know that $\delta_j = \alpha \delta_i$ for some $\alpha > 1$.
This allows us to rewrite~\refeq{eq:bound1}:
\begin{equation}
    \delta_j' \geq \underbrace{\frac{\delta_j}{\alpha}}_{\delta_i} + \lambda - \pnorm{\x_i-\x_j}_2
\end{equation}
The desired results follows from choosing $\lambda\geq 2\alpha\delta_j + \pnorm{\x_i-\x_j}_2$ yields:
\begin{equation}
\begin{split}
    \delta_j' &\geq \frac{\delta_j}{\alpha} + \lambda - \pnorm{\x_i-\x_j}_2\\
            & \geq  \frac{\delta_j}{\alpha} + 2\alpha\delta_j + \pnorm{\x_i-\x_j}_2 - \pnorm{\x_i-\x_j}_2\\
            &= \delta_j
\end{split}
\end{equation}
\end{proof}

\section{Experiments}\label{appendix:experiments}

\subsection{Global Poisoning Attack}\label{appendix:global}

\begin{table}[h!]
\caption{Difference (percentage) in the log-likelihood (i.e., plausibility) under a kernel density estimation of the counterfactuals: no poisoning vs. \emph{global poisoning}. In all cases, we add 70\% of the training data as poisonous instances. Positive numbers denote an increase (due to the data poisoning) in the plausibility of counterfactuals, while negative numbers denote the opposite. We report the median (over all folds) rounded to two decimal places, as well as the statistical significance according to the Mann-Whitney U test (ns $\implies$ p-value $> 0.05$; * $\implies$ p-value $\leq 0.05$; ** $\implies$ p-value $\leq 0.01$; *** $\implies$ p-value $\leq 0.001$).
}
\label{table:exp:results:global_density}
\centering
\begin{adjustbox}{max width=\textwidth}
\begin{tabular}{ccccc}
 \hline
 Classifier & Data set & NUN & DiCE & Proto\\
 \hline
  \multirow{3}{*}{{SVC}}
  & Credit & $1.73 * 10^{-6}\%_{ns}$  & $-2.15 * 10^{-3}\%_{ns}$ & $-6.13 * 10^{-1}\%_{ns}$  \\
  & Diabetes & $-3.57 * 10^{-1}\%_{ns}$ & $-2.65 * 10^{-1}\%_{ns}$ & $3.84 * 10^{-1}\%_{ns}$ \\
  & Crime &  $-6.16 * 10^{-9}\%_{ns}$ & $2.74 * 10^{-6}\%_{ns}$ & $3.74 * 10^{-1}\%_{ns}$ \\
 \hline
\multirow{3}{*}{{RNF}}
  & Credit & $-5.99 * 10^{-7}\%_{ns}$  & $-2.15 * 10^{-3}\%_{ns}$ & $-8.91 * 10^{-1}\%_{ns}$  \\
  & Diabetes & $6.63 * 10^{-1}\%_{ns}$ & $-3.95 * 10^{-1}\%_{ns}$ & $-8.46 * 10^{-1}\%_{ns}$ \\
  & Crime &  $-9.04 * 10^{-10}\%_{ns}$ & $-1.49 * 10^{-10}\%_{ns}$ & $-1.61 * 10^{-1}\%_{ns}$ \\
 \hline
  \multirow{3}{*}{{DNN}}
  & Credit & $-3.86 * 10^{-6}\%_{ns}$  & $-2.07 * 10^{-6}\%_{ns}$ & $8.72 * 10^{-1}\%_{ns}$  \\
  & Diabetes & $4.50 * 10^{-1}\%_{ns}$ & $-6.92 * 10^{-2}\%_{ns}$ & $-5.83 * 10^{-1}\%_{ns}$ \\
  & Crime &  $-1.34 * 10^{-8}\%_{ns}$ & $4.89 * 10^{-10}\%_{ns}$ & $7.91 * 10^{-1}\%_{ns}$ \\
 \hline
\end{tabular}
\end{adjustbox}
\end{table}

\begin{figure}[h!]
    \centering

    \begin{subfigure}{0.3\textwidth}
         \caption{$\classifier(\cdot)$: SVC -- CF: \emph{NUN}}
         \includegraphics[width=\textwidth]{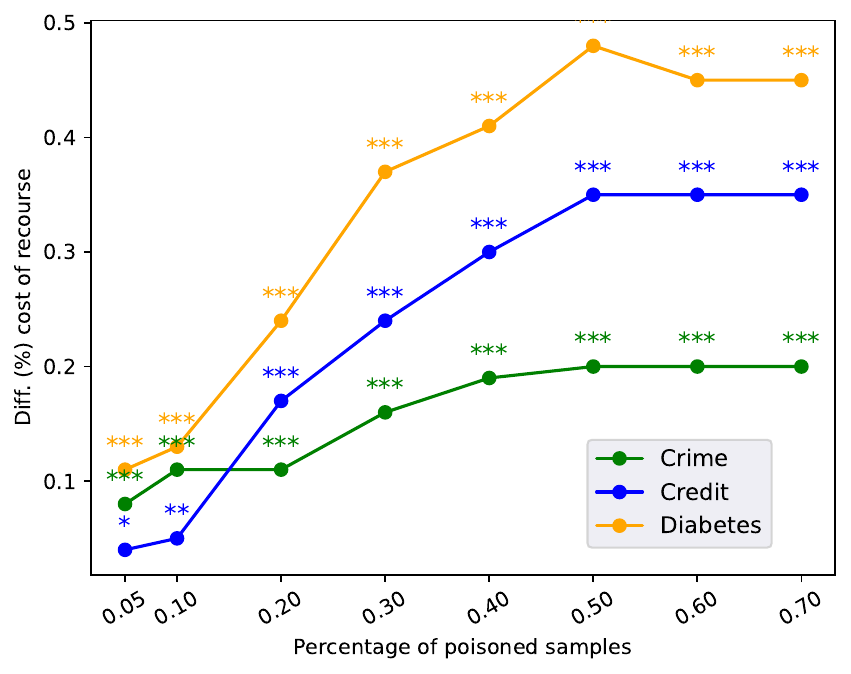}
    \end{subfigure}
    \begin{subfigure}{0.3\textwidth}
         \caption{$\classifier(\cdot)$: SVC -- CF: \emph{DiCE}}
         \includegraphics[width=\textwidth]{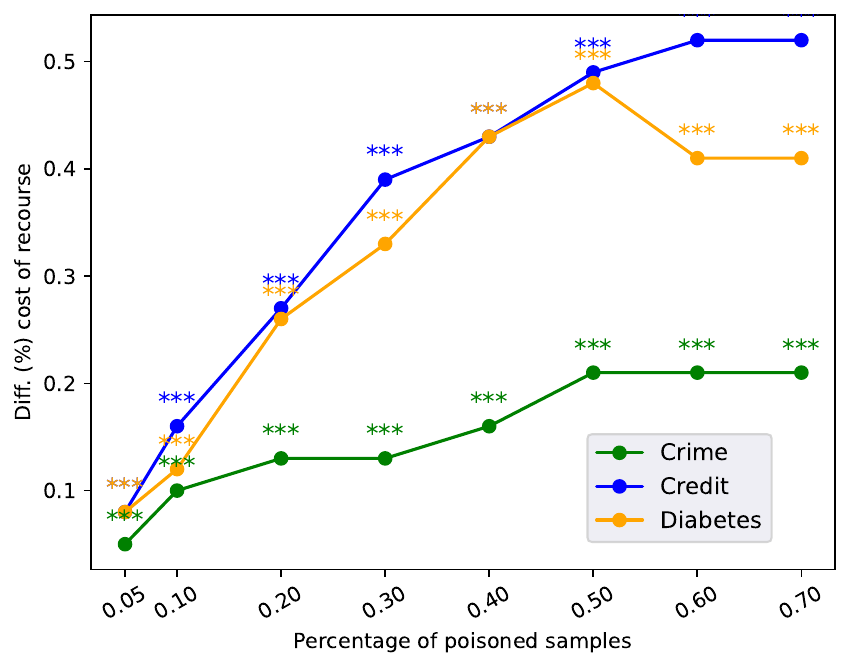}
    \end{subfigure}
    \begin{subfigure}{0.3\textwidth}
         \caption{$\classifier(\cdot)$: SVC -- CF: \emph{Proto}}
         \includegraphics[width=\textwidth]{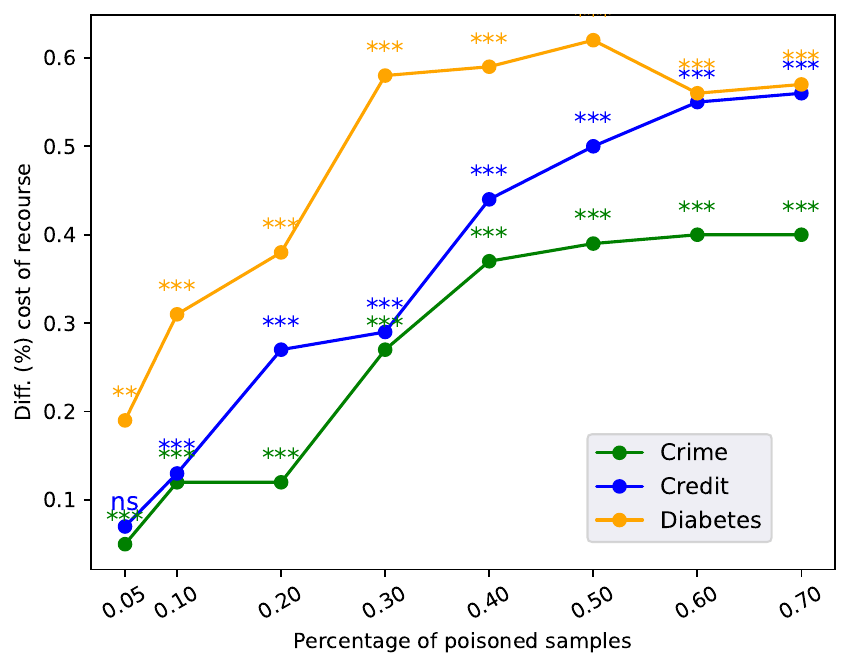}
    \end{subfigure}
    
    \begin{subfigure}{0.3\textwidth}
         \caption{$\classifier(\cdot)$: RNF -- CF: \emph{Proto}}
         \includegraphics[width=\textwidth]{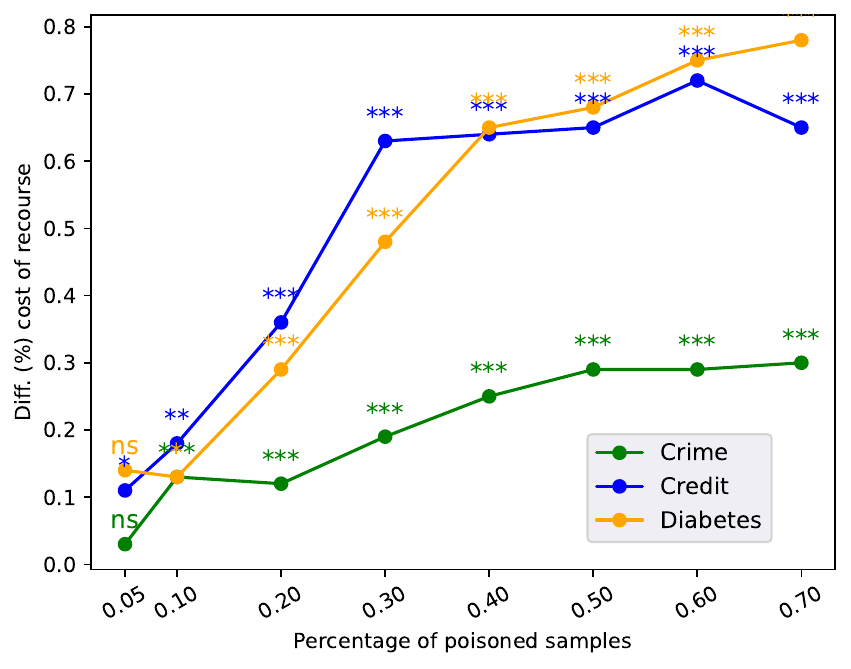}
    \end{subfigure}
        \begin{subfigure}[b]{0.3\textwidth}
         \caption{$\classifier(\cdot)$: RNF -- CF: \emph{NUN}}
         \includegraphics[width=\textwidth]{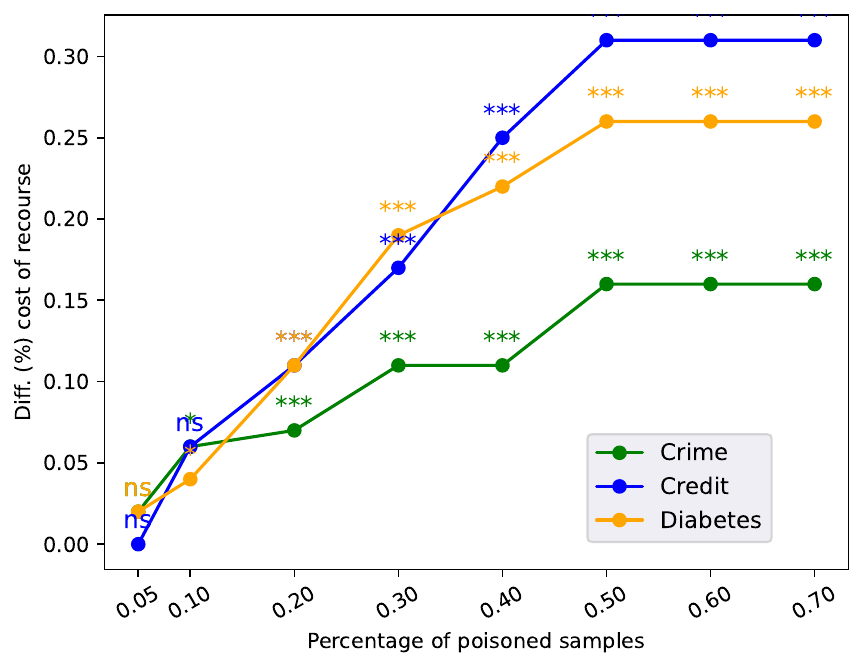}
    \end{subfigure}
    \begin{subfigure}[b]{0.3\textwidth}
         \caption{$\classifier(\cdot)$: RNF -- CF: \emph{DiCE}}
         \includegraphics[width=\textwidth]{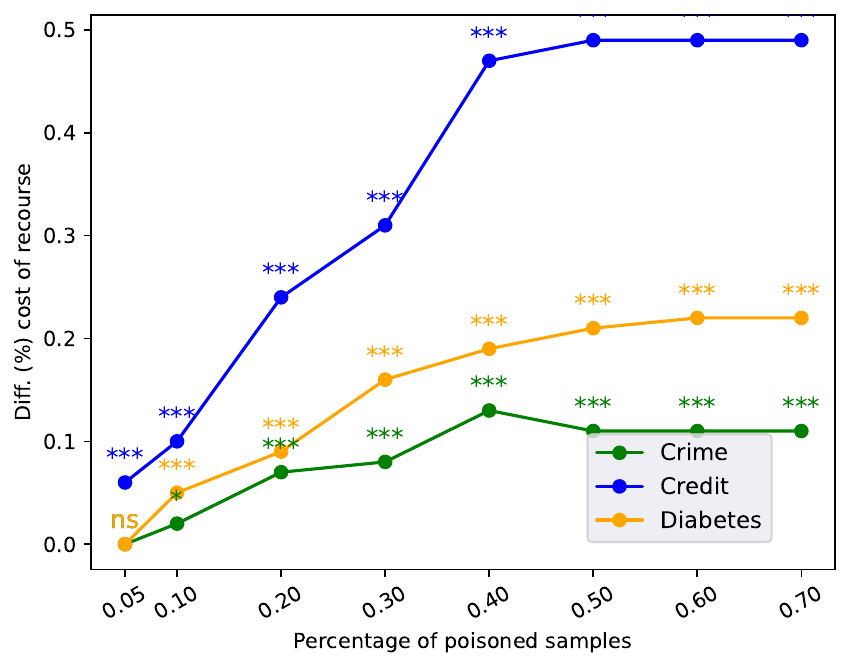}
    \end{subfigure}

    \begin{subfigure}[b]{0.3\textwidth}
         \caption{$\classifier(\cdot)$: DNN -- CF: \emph{NUN}}
         \includegraphics[width=\textwidth]{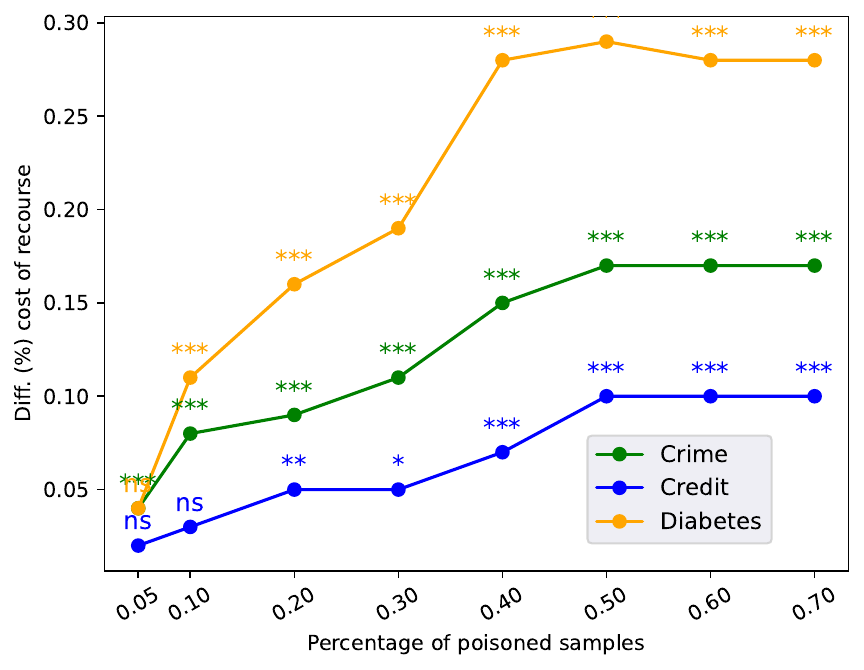}
    \end{subfigure}
    \begin{subfigure}[b]{0.3\textwidth}
         \caption{$\classifier(\cdot)$: DNN -- CF: \emph{DiCE}}
         \includegraphics[width=\textwidth]{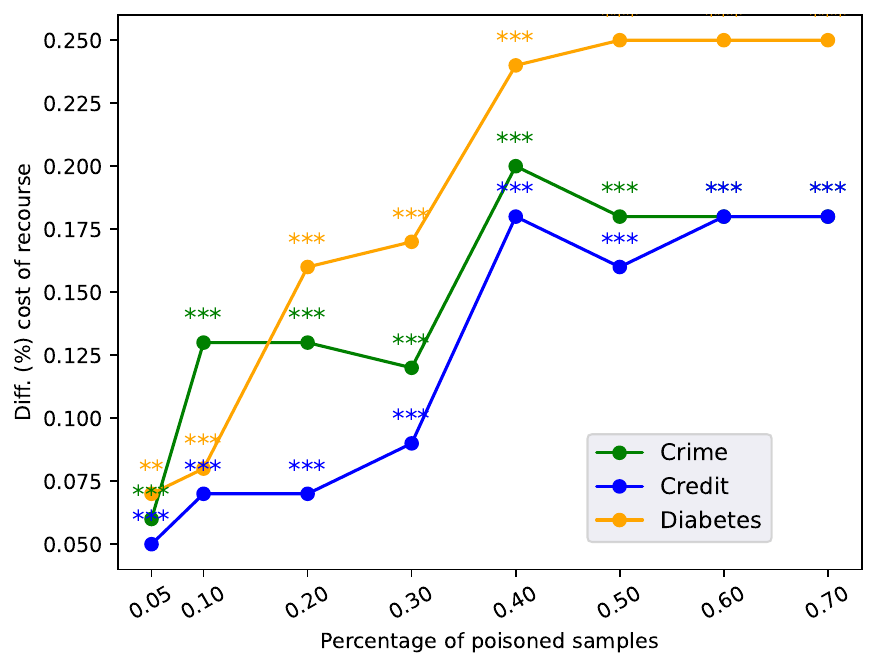}
    \end{subfigure}
       \begin{subfigure}{0.3\textwidth}
         \caption{$\classifier(\cdot)$: DNN -- CF: \emph{Proto}}
         \includegraphics[width=\textwidth]{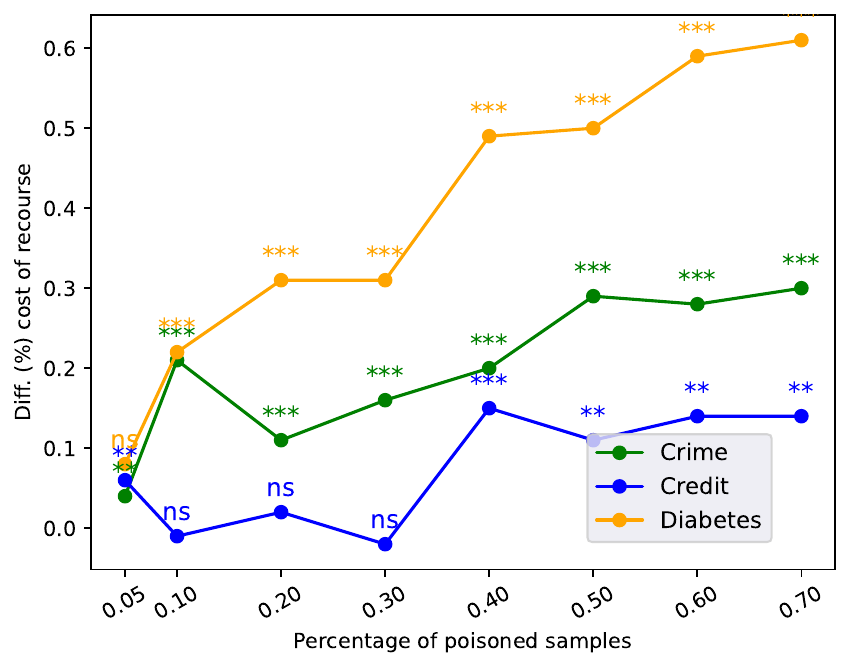}
    \end{subfigure}
    
    \caption{Global data poisoning attack: Difference (percentage) in the cost of recourse vs. percentage of poisoned instances (5\% to 70\%). We report the median (over all folds) rounded to two decimal places, as well as the statistical significance according to the Mann-Whitney U test (ns $\implies$ p-value $> 0.05$; * $\implies$ p-value $\leq 0.05$; ** $\implies$ p-value $\leq 0.01$; *** $\implies$ p-value $\leq 0.001$.}
    \label{appendix:fig:exp:results:recourse_vs_poisonsamples-1}
\end{figure}

\FloatBarrier

\begin{figure}[h!]
    \centering

    \begin{subfigure}{0.3\textwidth}
         \caption{$\classifier(\cdot)$: SVC -- CF: \emph{NUN}}
         \includegraphics[width=\textwidth]{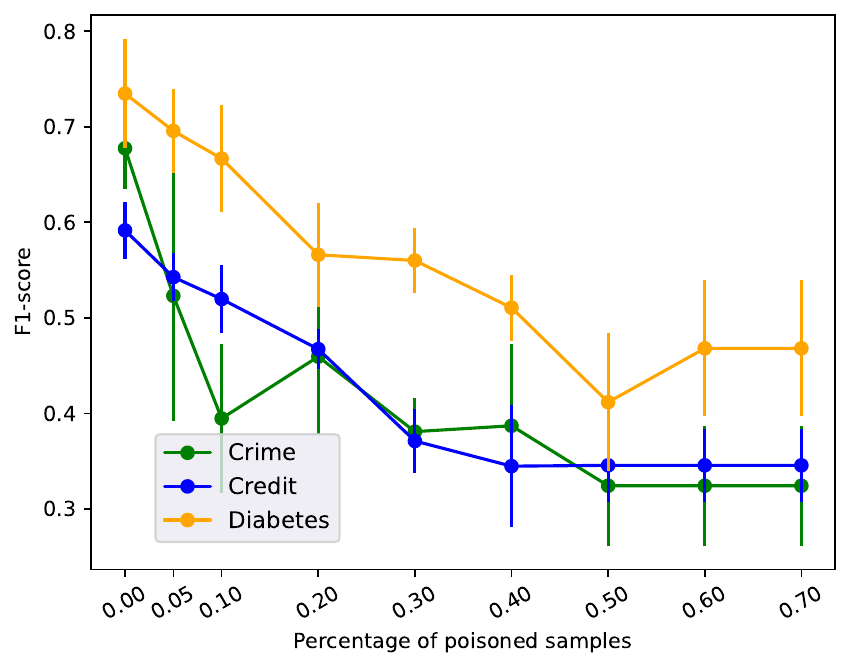}
    \end{subfigure}
    \begin{subfigure}{0.3\textwidth}
         \caption{$\classifier(\cdot)$: SVC -- CF: \emph{DiCE}}
         \includegraphics[width=\textwidth]{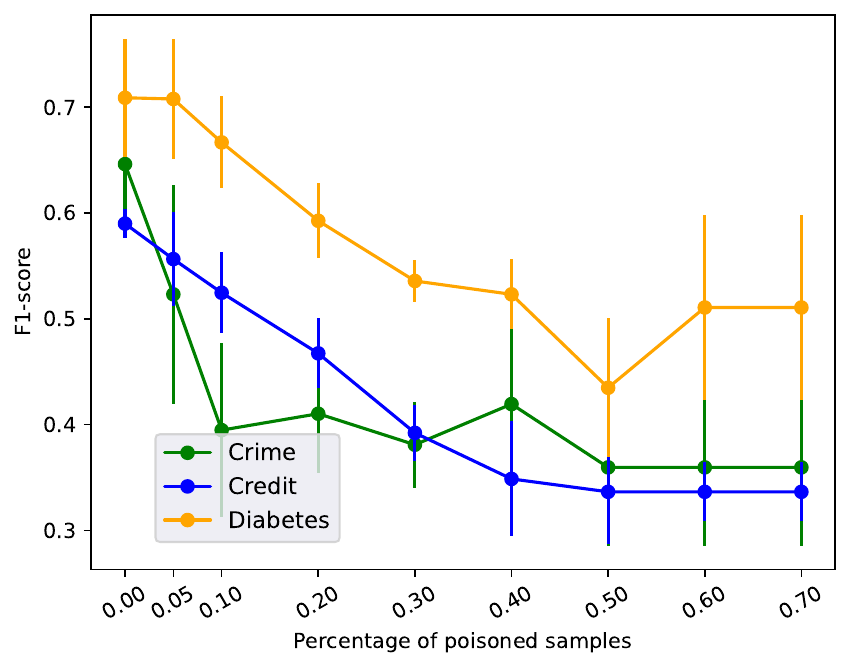}
    \end{subfigure}
    \begin{subfigure}{0.3\textwidth}
         \caption{$\classifier(\cdot)$: SVC -- CF: \emph{Proto}}
         \includegraphics[width=\textwidth]{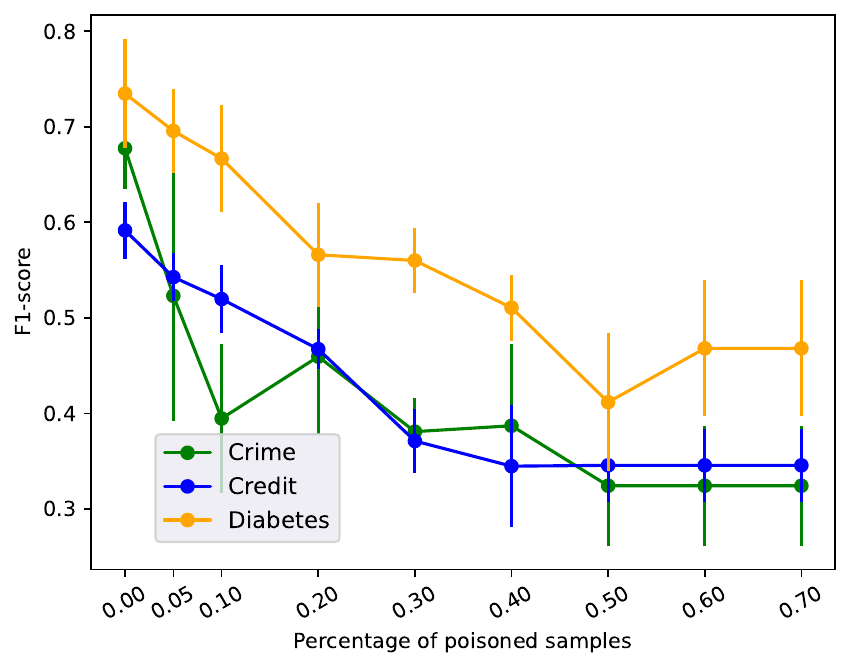}
    \end{subfigure}

    \begin{subfigure}{0.3\textwidth}
         \caption{$\classifier(\cdot)$: RNF -- CF: \emph{Proto}}
         \includegraphics[width=\textwidth]{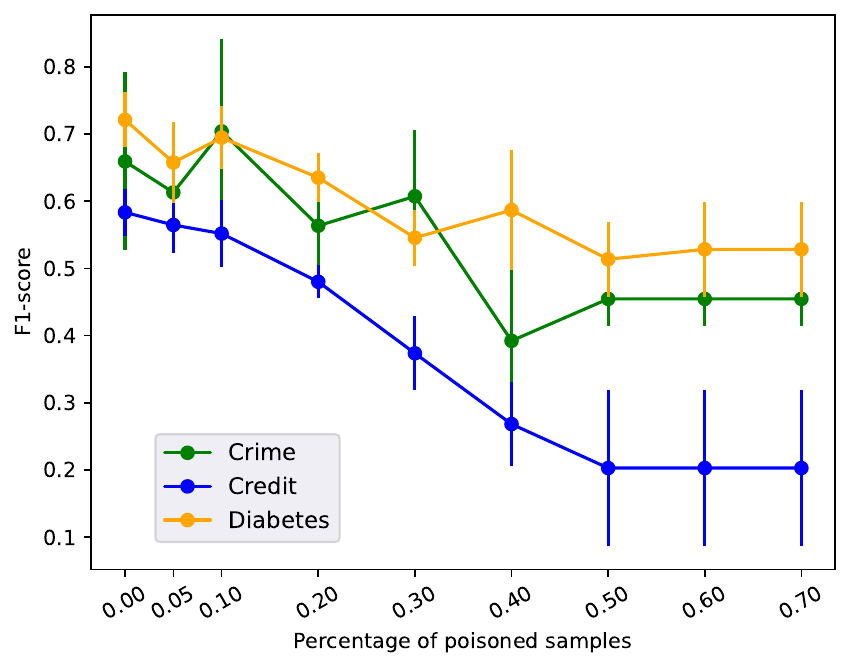}
    \end{subfigure}
    \begin{subfigure}{0.3\textwidth}
         \caption{$\classifier(\cdot)$: RNF -- CF: \emph{NUN}}
         \includegraphics[width=\textwidth]{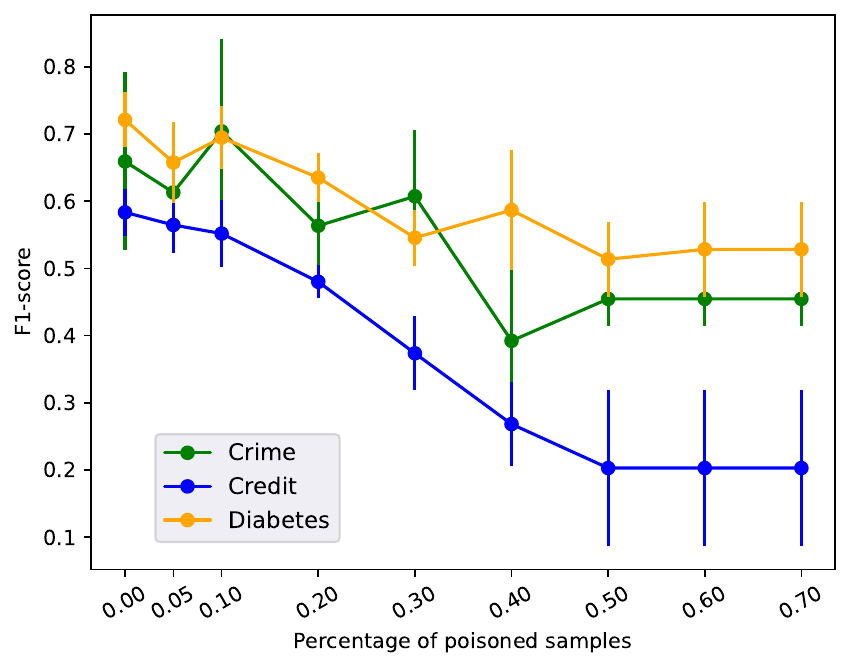}
    \end{subfigure}
    \begin{subfigure}{0.3\textwidth}
         \caption{$\classifier(\cdot)$: RNF -- CF: \emph{DiCE}}
         \includegraphics[width=\textwidth]{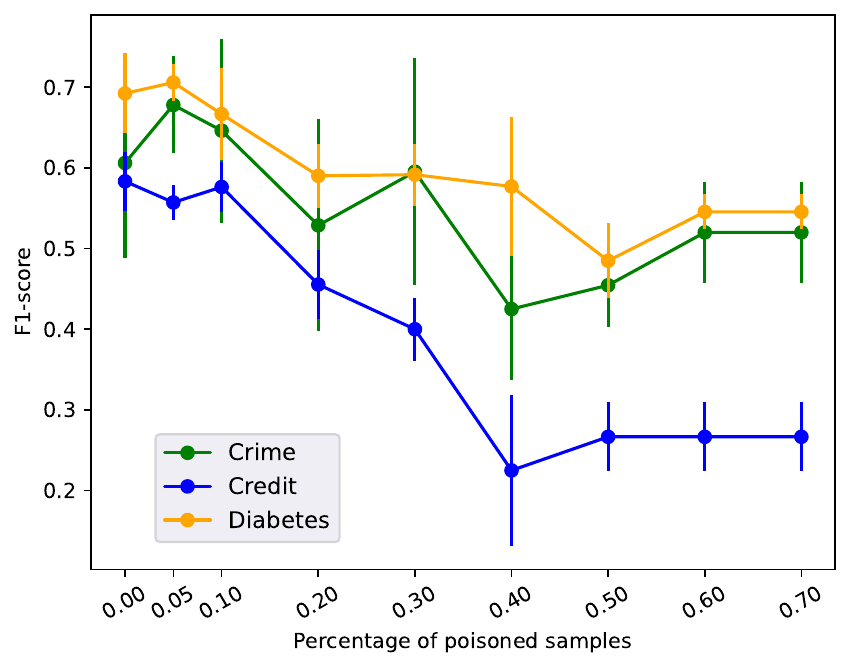}
    \end{subfigure}

    \begin{subfigure}[b]{0.3\textwidth}
         \caption{$\classifier(\cdot)$: DNN -- CF: \emph{NUN}}
         \includegraphics[width=\textwidth]{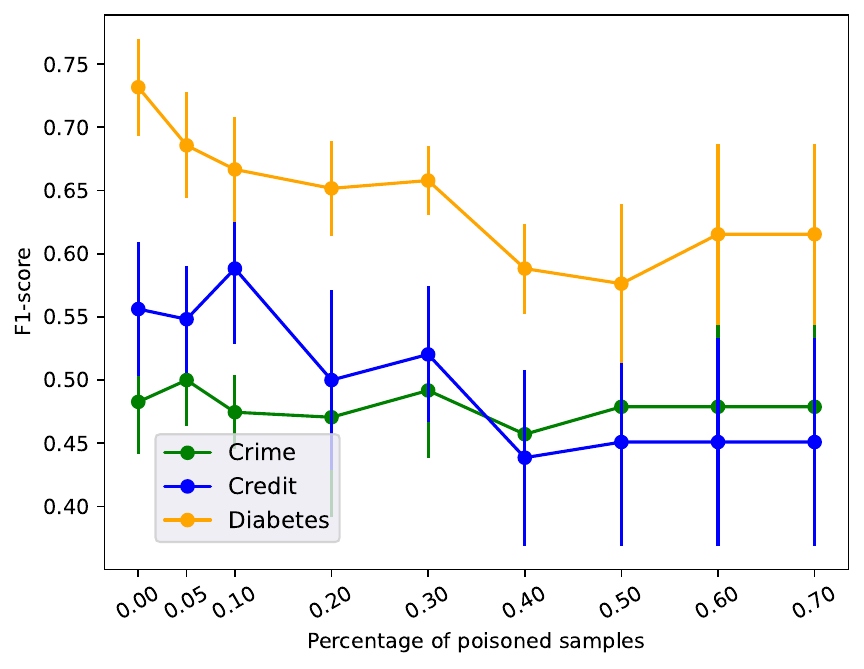}
    \end{subfigure}
    \begin{subfigure}[b]{0.3\textwidth}
         \caption{$\classifier(\cdot)$: DNN -- CF: \emph{DiCE}}
         \includegraphics[width=\textwidth]{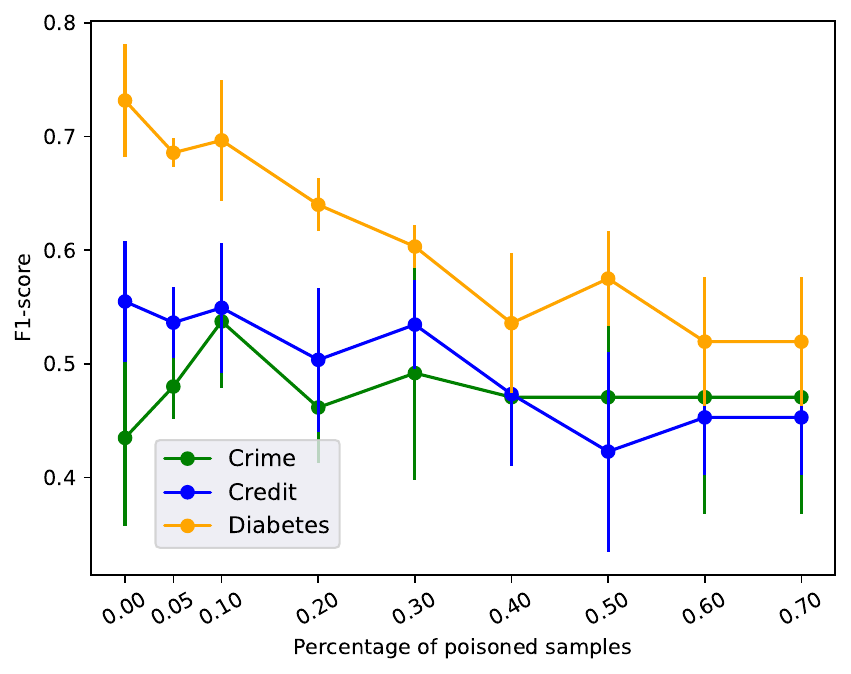}
    \end{subfigure}
       \begin{subfigure}{0.3\textwidth}
         \caption{$\classifier(\cdot)$: DNN -- CF: \emph{Proto}}
         \includegraphics[width=\textwidth]{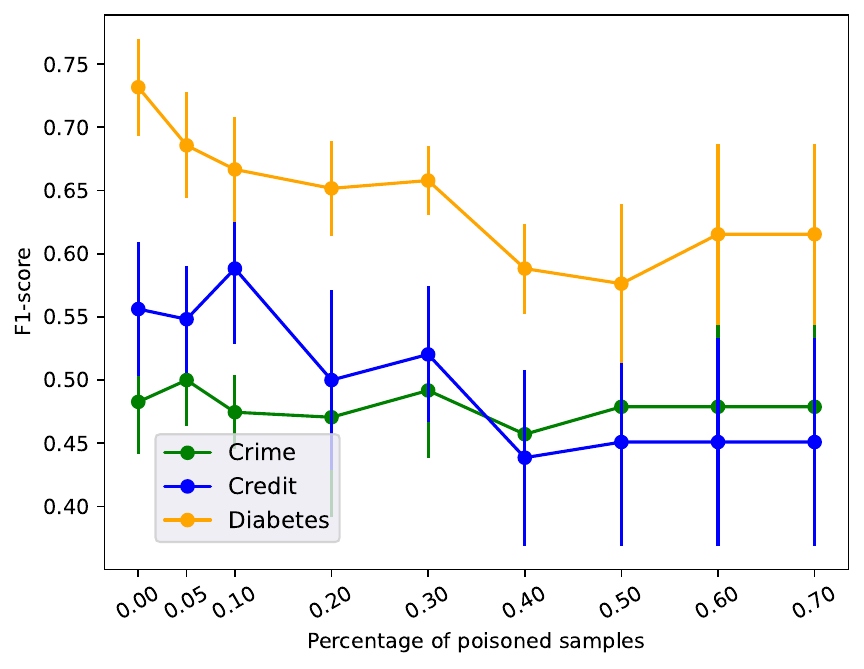}
    \end{subfigure}

    \caption{Global data poisoning attack: Median and standard deviation (over all folds) F1-score of the classifier for different percentages of poisoned samples (0\% to 70\%).}
    \label{appendix:fig:exp:results:accuracy_vs_poisonsamples-1}
\end{figure}

\begin{figure}[h!]
    \centering

    \begin{subfigure}{0.3\textwidth}
         \caption{$\classifier(\cdot)$: SVC -- CF: \emph{NUN}}
         \includegraphics[width=\textwidth]{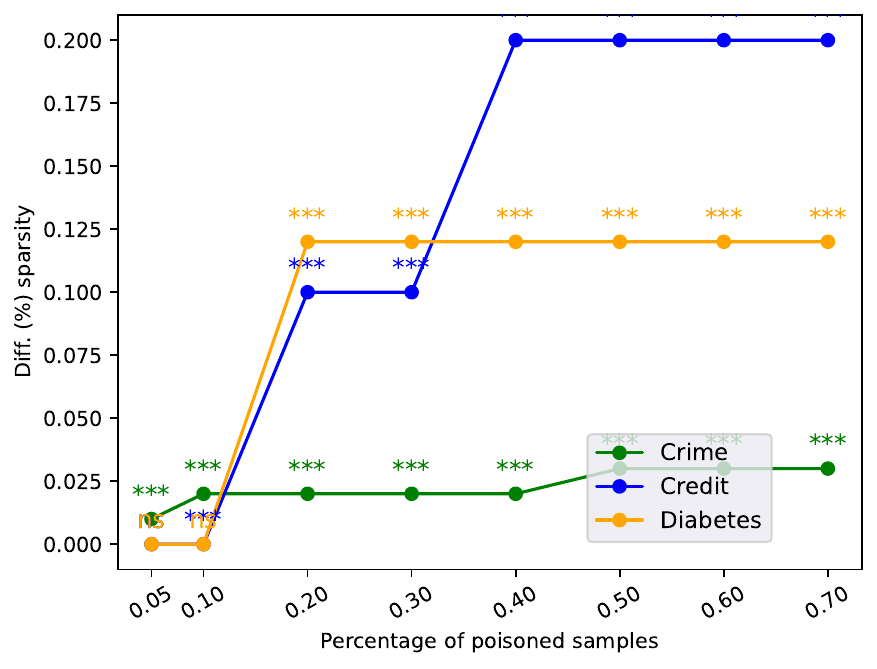}
    \end{subfigure}
    \begin{subfigure}{0.3\textwidth}
         \caption{$\classifier(\cdot)$: SVC -- CF: \emph{DiCE}}
         \includegraphics[width=\textwidth]{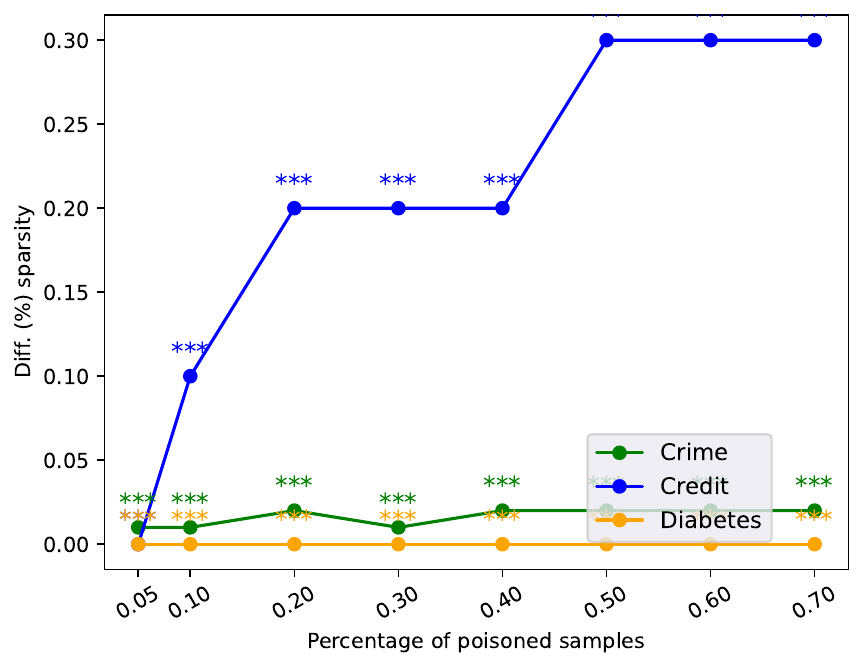}
    \end{subfigure}
    \begin{subfigure}{0.3\textwidth}
         \caption{$\classifier(\cdot)$: SVC -- CF: \emph{Proto}}
         \includegraphics[width=\textwidth]{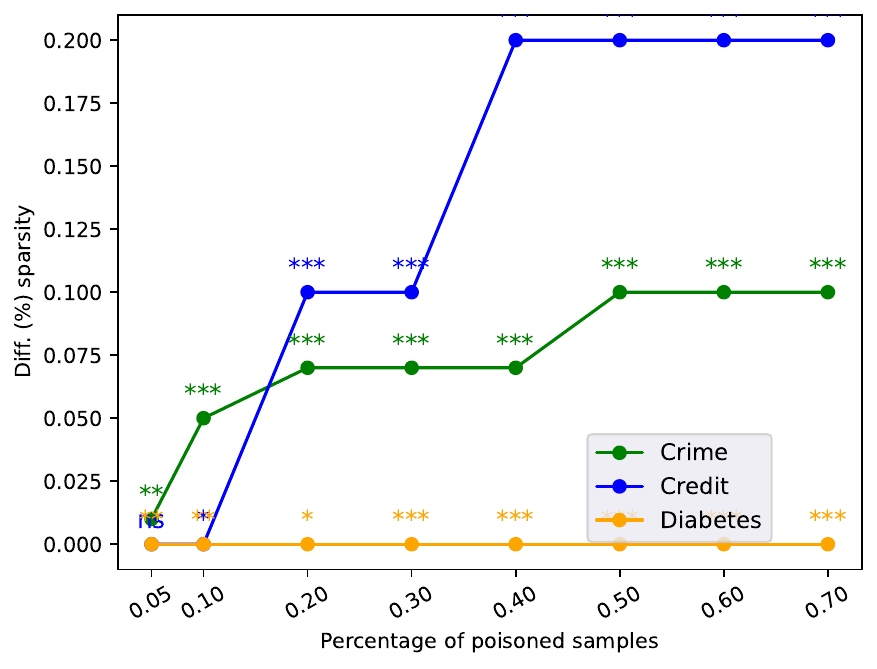}
    \end{subfigure}

    \begin{subfigure}{0.3\textwidth}
         \caption{$\classifier(\cdot)$: RNF -- CF: \emph{Proto}}
         \includegraphics[width=\textwidth]{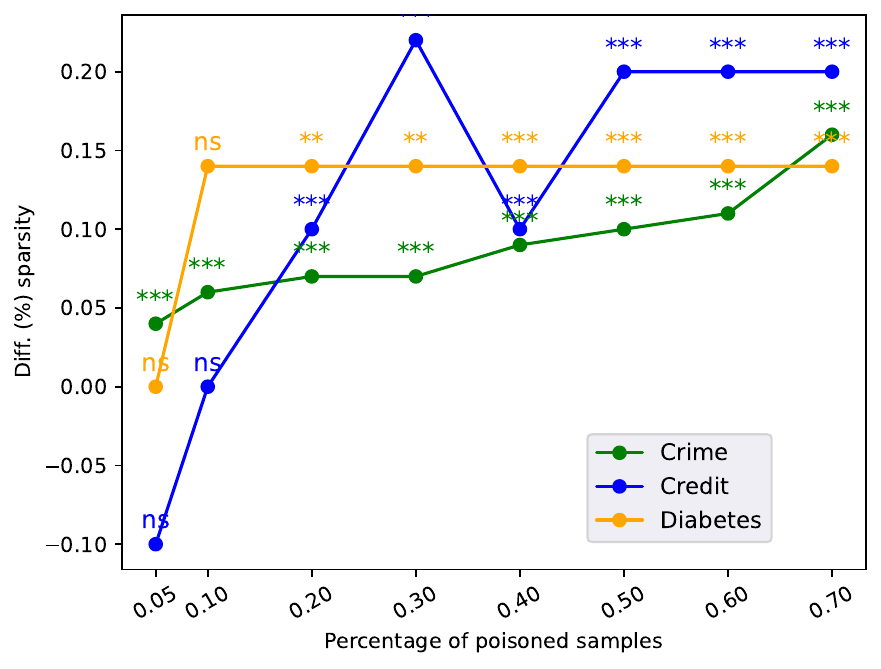}
    \end{subfigure}
    \begin{subfigure}{0.3\textwidth}
         \caption{$\classifier(\cdot)$: RNF -- CF: \emph{NUN}}
         \includegraphics[width=\textwidth]{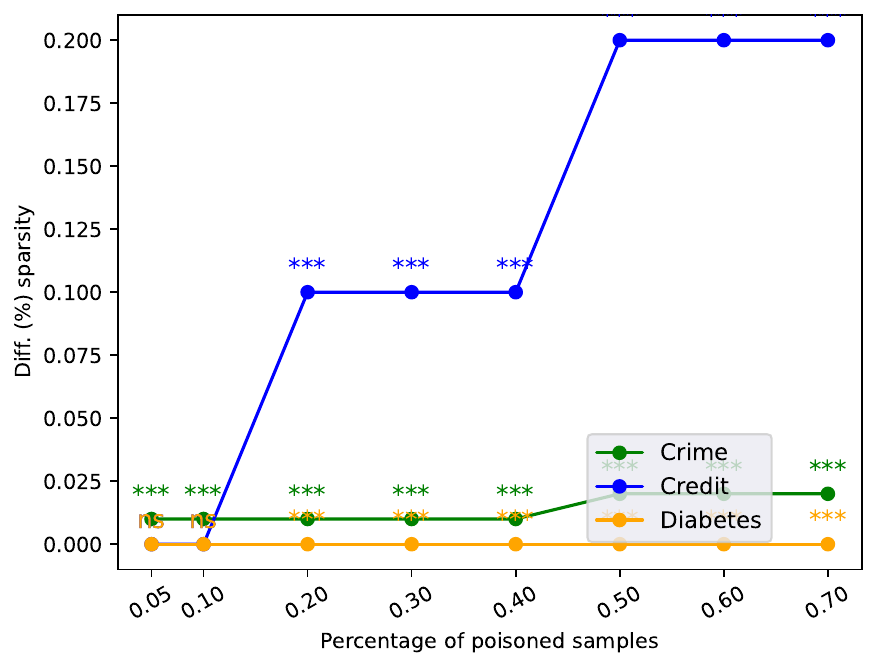}
    \end{subfigure}
    \begin{subfigure}{0.3\textwidth}
         \caption{$\classifier(\cdot)$: RNF -- CF: \emph{DiCE}}
         \includegraphics[width=\textwidth]{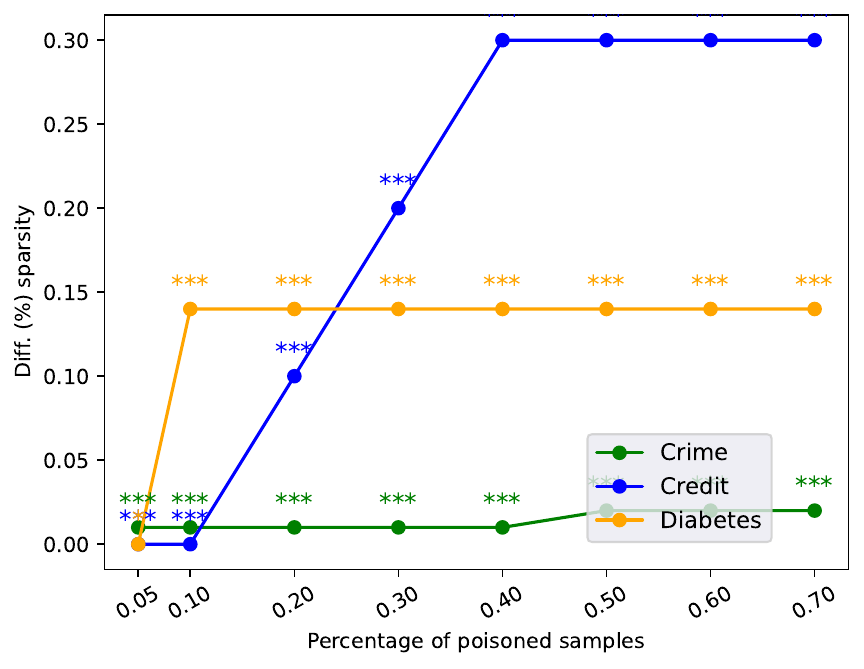}
    \end{subfigure}

    \begin{subfigure}[b]{0.3\textwidth}
         \caption{$\classifier(\cdot)$: DNN -- CF: \emph{NUN}}
         \includegraphics[width=\textwidth]{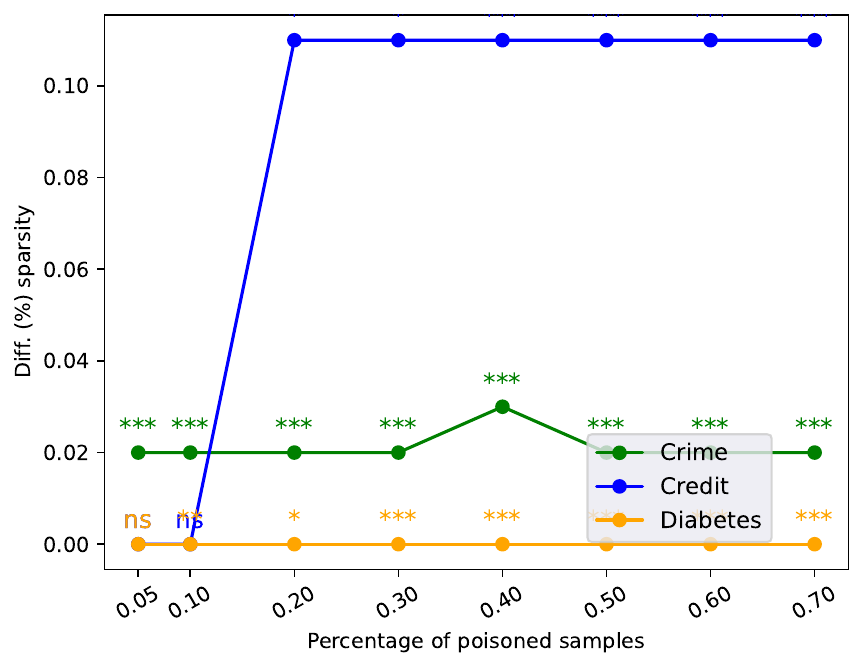}
    \end{subfigure}
    \begin{subfigure}[b]{0.3\textwidth}
         \caption{$\classifier(\cdot)$: DNN -- CF: \emph{DiCE}}
         \includegraphics[width=\textwidth]{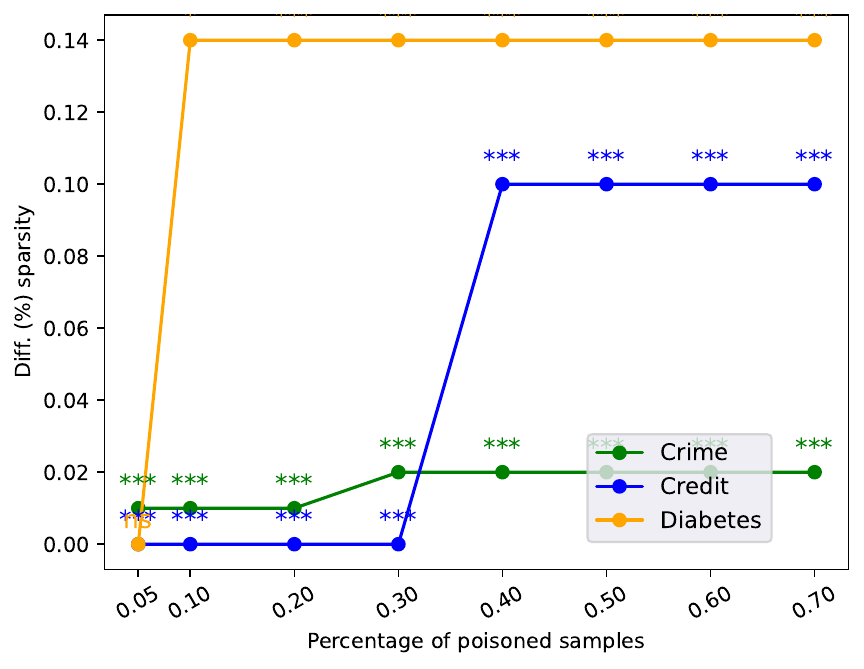}
    \end{subfigure}
        \begin{subfigure}{0.3\textwidth}
         \caption{$\classifier(\cdot)$: DNN -- CF: \emph{Proto}}
         \includegraphics[width=\textwidth]{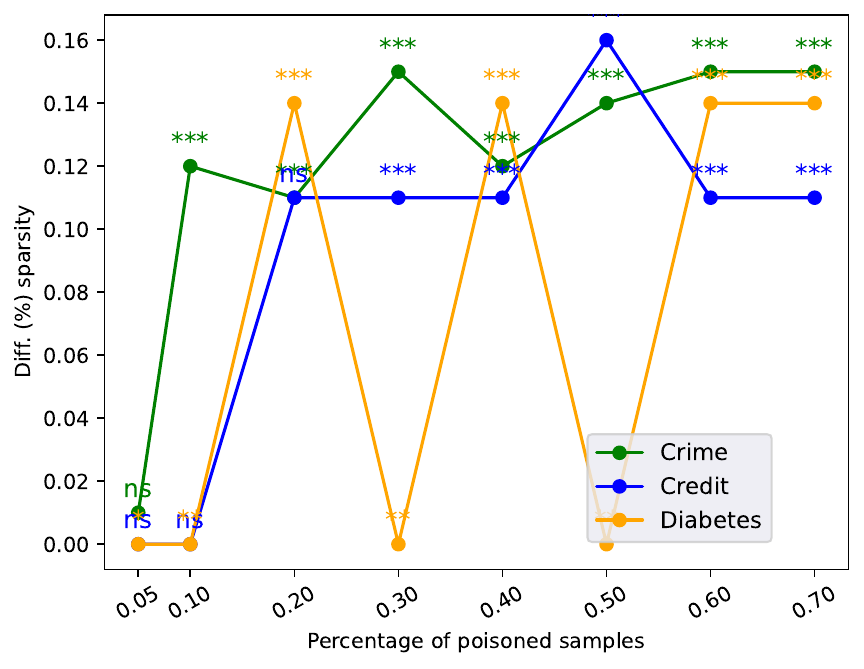}
    \end{subfigure}

    \caption{Global data poisoning attack: Sparsity of the counterfactual explanations for different percentages of poisoned samples (0\% to 70\%). We report the median (over all folds) rounded to two decimal places, as well as the statistical significance according to the Mann-Whitney U test (ns $\implies$ p-value $> 0.05$; * $\implies$ p-value $\leq 0.05$; ** $\implies$ p-value $\leq 0.01$; *** $\implies$ p-value $\leq 0.001$.}
    \label{appendix:fig:exp:results:sparsity_vs_poisonsamples-1}
\end{figure}

\FloatBarrier

\subsection{Sub-group Poisoning Attack}\label{appendix:subgroup}

\begin{table}[h!]
\caption{Nearest Unlike Neighbor (NUN). Difference (percentage) in the cost of recourse between protected groups (see~\refeq{eq:eval:difffairness_percentage}): no vs. poisoning on a \emph{sub-group level}. We report the median (over all folds) rounded to two decimal places, as well as the statistical significance according to the Mann-Whitney U test (ns $\implies$ p-value $> 0.05$; * $\implies$ p-value $\leq 0.05$; ** $\implies$ p-value $\leq 0.01$; *** $\implies$ p-value $\leq 0.001$).
}
\label{table:exp:results:subgroup_mem}
\centering
\begin{adjustbox}{max width=\textwidth}
    \begin{tabular}{cccccccccc}
     \hline
     Classifier & & \multicolumn{8}{c}{Percentage of poisoned samples} \\
     \cline{3-10}
     & Dataset & 0.01 & 0.1 & 0.2 & 0.3 & 0.4 & 0.5 & 0.6 & 0.7 \\
     \hline
    
     \multirow{3}{*}{{SVC}} 
    & Credit & $24\%_{\footnotesize ns}$ & $45\%_{\footnotesize ns}$ & $15\%_{\footnotesize ***}$ & $-15\%_{\footnotesize ***}$ & $52\%_{\footnotesize **}$ & $10\%_{\footnotesize ***}$ & $10\%_{\footnotesize ***}$ & $10\%_{\footnotesize ***}$ \\ 
    & Diabetes & $34\%_{\footnotesize ns}$ & $41\%_{\footnotesize **}$ & $109\%_{\footnotesize ***}$ & $84\%_{\footnotesize ***}$ & $106\%_{\footnotesize ***}$ & $111\%_{\footnotesize ***}$ & $110\%_{\footnotesize ***}$ & $110\%_{\footnotesize ***}$ \\ 
    & Crime & $44\%_{\footnotesize ns}$ & $53\%_{\footnotesize **}$ & $72\%_{\footnotesize ***}$ & $73\%_{\footnotesize ***}$ & $99\%_{\footnotesize ***}$ & $86\%_{\footnotesize ***}$ & $86\%_{\footnotesize ***}$ & $86\%_{\footnotesize ***}$ \\ 
    \hline 
    
     \multirow{3}{*}{{RNF}} 
    & Credit & $131\%_{\footnotesize ns}$ & $144\%_{\footnotesize ns}$ & $69\%_{\footnotesize ns}$ & $325\%_{\footnotesize ns}$ & $519\%_{\footnotesize ns}$ & $88\%_{\footnotesize ns}$ & $381\%_{\footnotesize ns}$ & $381\%_{\footnotesize ns}$ \\ 
    & Diabetes & $275\%_{\footnotesize ns}$ & $225\%_{\footnotesize ns}$ & $319\%_{\footnotesize *}$ & $419\%_{\footnotesize ***}$ & $344\%_{\footnotesize ***}$ & $269\%_{\footnotesize **}$ & $325\%_{\footnotesize ***}$ & $325\%_{\footnotesize ***}$ \\ 
    & Crime & $1\%_{\footnotesize ns}$ & $37\%_{\footnotesize ns}$ & $53\%_{\footnotesize ns}$ & $81\%_{\footnotesize ns}$ & $52\%_{\footnotesize *}$ & $52\%_{\footnotesize ns}$ & $41\%_{\footnotesize ns}$ & $41\%_{\footnotesize ns}$ \\ 
    \hline 
    
     \multirow{3}{*}{{DNN}} 
    & Credit & $373\%_{\footnotesize ns}$ & $227\%_{\footnotesize ns}$ & $82\%_{\footnotesize ns}$ & $364\%_{\footnotesize ns}$ & $-36\%_{\footnotesize ns}$ & $27\%_{\footnotesize ns}$ & $27\%_{\footnotesize ns}$ & $27\%_{\footnotesize ns}$ \\ 
    & Diabetes & $-22\%_{\footnotesize ns}$ & $25\%_{\footnotesize **}$ & $7\%_{\footnotesize ***}$ & $-15\%_{\footnotesize ***}$ & $70\%_{\footnotesize ***}$ & $40\%_{\footnotesize ***}$ & $0\%_{\footnotesize ns}$ & $0\%_{\footnotesize ns}$ \\ 
    & Crime & $86\%_{\footnotesize ns}$ & $64\%_{\footnotesize *}$ & $86\%_{\footnotesize **}$ & $117\%_{\footnotesize ***}$ & $114\%_{\footnotesize ***}$ & $119\%_{\footnotesize ***}$ & $119\%_{\footnotesize ***}$ & $119\%_{\footnotesize ***}$ \\ 
    \hline 
    \end{tabular}
\end{adjustbox}
\end{table}

\begin{table}[h!]
\caption{DiCE. Difference (percentage) in the cost of recourse between protected groups (see~\refeq{eq:eval:difffairness_percentage}): no vs. poisoning on a \emph{sub-group level}. We report the median (over all folds) rounded to two decimal places, as well as the statistical significance according to the Mann-Whitney U test (ns $\implies$ p-value $> 0.05$; * $\implies$ p-value $\leq 0.05$; ** $\implies$ p-value $\leq 0.01$; *** $\implies$ p-value $\leq 0.001$).
}
\label{table:exp:results:subgroup_dice}
\centering
\begin{adjustbox}{max width=\textwidth}
    \begin{tabular}{cccccccccc}
     \hline
     Classifier & & \multicolumn{8}{c}{Percentage of poisoned samples} \\
     \cline{3-10}
     & Dataset & 0.01 & 0.1 & 0.2 & 0.3 & 0.4 & 0.5 & 0.6 & 0.7 \\
     \hline

      \multirow{3}{*}{{SVC}} 
    & Credit & $117\%_{\footnotesize **}$ & $86\%_{\footnotesize ***}$ & $120\%_{\footnotesize ***}$ & $111\%_{\footnotesize ***}$ & $124\%_{\footnotesize ***}$ & $70\%_{\footnotesize ***}$ & $70\%_{\footnotesize ***}$ & $70\%_{\footnotesize ***}$ \\ 
    & Diabetes & $7\%_{\footnotesize ***}$ & $-12\%_{\footnotesize ***}$ & $61\%_{\footnotesize ***}$ & $32\%_{\footnotesize ***}$ & $50\%_{\footnotesize ***}$ & $57\%_{\footnotesize ***}$ & $42\%_{\footnotesize ***}$ & $42\%_{\footnotesize ***}$ \\ 
    & Crime & $6\%_{\footnotesize ns}$ & $34\%_{\footnotesize ***}$ & $69\%_{\footnotesize ***}$ & $59\%_{\footnotesize ***}$ & $59\%_{\footnotesize ***}$ & $56\%_{\footnotesize ***}$ & $56\%_{\footnotesize ***}$ & $56\%_{\footnotesize ***}$ \\ 
    \hline 
    
     \multirow{3}{*}{{RNF}} 
    & Credit & $-28\%_{\footnotesize ***}$ & $44\%_{\footnotesize *}$ & $90\%_{\footnotesize ***}$ & $137\%_{\footnotesize ***}$ & $39\%_{\footnotesize ***}$ & $-39\%_{\footnotesize ***}$ & $-39\%_{\footnotesize ***}$ & $-39\%_{\footnotesize ***}$ \\ 
    & Diabetes & $74\%_{\footnotesize ns}$ & $53\%_{\footnotesize ns}$ & $4\%_{\footnotesize ***}$ & $19\%_{\footnotesize ***}$ & $132\%_{\footnotesize ***}$ & $87\%_{\footnotesize ***}$ & $98\%_{\footnotesize ***}$ & $98\%_{\footnotesize ***}$ \\ 
    & Crime & $-1\%_{\footnotesize ns}$ & $44\%_{\footnotesize ***}$ & $68\%_{\footnotesize **}$ & $67\%_{\footnotesize ***}$ & $62\%_{\footnotesize ***}$ & $59\%_{\footnotesize ***}$ & $59\%_{\footnotesize ***}$ & $59\%_{\footnotesize ***}$ \\ 
    \hline 
    
     \multirow{3}{*}{{DNN}} 
    & Credit & $25\%_{\footnotesize **}$ & $-28\%_{\footnotesize ***}$ & $-73\%_{\footnotesize ***}$ & $-83\%_{\footnotesize ***}$ & $-35\%_{\footnotesize ***}$ & $27\%_{\footnotesize ***}$ & $27\%_{\footnotesize ***}$ & $27\%_{\footnotesize ***}$ \\ 
    & Diabetes & $19\%_{\footnotesize ns}$ & $28\%_{\footnotesize ***}$ & $44\%_{\footnotesize ***}$ & $7\%_{\footnotesize ***}$ & $77\%_{\footnotesize ***}$ & $60\%_{\footnotesize ***}$ & $33\%_{\footnotesize ***}$ & $33\%_{\footnotesize ***}$ \\ 
    & Crime & $59\%_{\footnotesize ***}$ & $66\%_{\footnotesize ***}$ & $68\%_{\footnotesize ***}$ & $70\%_{\footnotesize ***}$ & $73\%_{\footnotesize ***}$ & $84\%_{\footnotesize ***}$ & $84\%_{\footnotesize ***}$ & $84\%_{\footnotesize ***}$ \\ 
    \hline 
    \end{tabular}
\end{adjustbox}
\end{table}

\begin{table}[h!]
\caption{Counterfactuals guided by prototypes (Proto). Difference (percentage) in the cost of recourse between protected groups (see~\refeq{eq:eval:difffairness_percentage}): no vs. poisoning on a \emph{sub-group level}. We report the median (over all folds) rounded to two decimal places, as well as the statistical significance according to the Mann-Whitney U test (ns $\implies$ p-value $> 0.05$; * $\implies$ p-value $\leq 0.05$; ** $\implies$ p-value $\leq 0.01$; *** $\implies$ p-value $\leq 0.001$).
}
\label{table:exp:results:subgroup_proto}
\centering
\begin{adjustbox}{max width=\textwidth}
    \begin{tabular}{cccccccccc}
     \hline
     Classifier & & \multicolumn{8}{c}{Percentage of poisoned samples} \\
     \cline{3-10}
     & Dataset & 0.01 & 0.1 & 0.2 & 0.3 & 0.4 & 0.5 & 0.6 & 0.7 \\
     \hline

     \multirow{3}{*}{{SVC}} 
    & Credit & $957\%_{\footnotesize ns}$ & $118\%_{\footnotesize *}$ & $162\%_{\footnotesize ***}$ & $2032\%_{\footnotesize ***}$ & $2450\%_{\footnotesize **}$ & $150\%_{\footnotesize ***}$ & $-57\%_{\footnotesize ***}$ & $0\%_{\footnotesize ns}$ \\ 
    & Diabetes & $12\%_{\footnotesize ns}$ & $-12\%_{\footnotesize ***}$ & $41\%_{\footnotesize ***}$ & $8\%_{\footnotesize ***}$ & $10\%_{\footnotesize ***}$ & $62\%_{\footnotesize ***}$ & $38\%_{\footnotesize ***}$ & $66\%_{\footnotesize ***}$ \\ 
    & Crime & $17\%_{\footnotesize ns}$ & $25\%_{\footnotesize *}$ & $35\%_{\footnotesize ***}$ & $40\%_{\footnotesize ***}$ & $33\%_{\footnotesize ***}$ & $18\%_{\footnotesize ***}$ & $17\%_{\footnotesize ***}$ & $14\%_{\footnotesize ***}$ \\ 
    \hline 
    
     \multirow{3}{*}{{RNF}} 
    & Credit & $341\%_{\footnotesize ns}$ & $-68\%_{\footnotesize *}$ & $-71\%_{\footnotesize **}$ & $-1\%_{\footnotesize *}$ & $-10\%_{\footnotesize **}$ & $-80\%_{\footnotesize ns}$ & $-68\%_{\footnotesize **}$ & $375\%_{\footnotesize *}$ \\ 
    & Diabetes & $57\%_{\footnotesize ns}$ & $33\%_{\footnotesize ns}$ & $103\%_{\footnotesize **}$ & $53\%_{\footnotesize ***}$ & $137\%_{\footnotesize **}$ & $113\%_{\footnotesize ***}$ & $95\%_{\footnotesize ***}$ & $111\%_{\footnotesize ***}$ \\ 
    & Crime & $7\%_{\footnotesize ns}$ & $-3\%_{\footnotesize ns}$ & $19\%_{\footnotesize *}$ & $12\%_{\footnotesize *}$ & $15\%_{\footnotesize ns}$ & $4\%_{\footnotesize **}$ & $11\%_{\footnotesize ns}$ & $10\%_{\footnotesize *}$ \\ 
    \hline 
    
     \multirow{3}{*}{{DNN}} 
    & Credit & $0\%_{\footnotesize ns}$ & $65\%_{\footnotesize ns}$ & $-28\%_{\footnotesize ns}$ & $37\%_{\footnotesize ns}$ & $-90\%_{\footnotesize ns}$ & $-61\%_{\footnotesize ns}$ & $3\%_{\footnotesize ns}$ & $191\%_{\footnotesize ns}$ \\ 
    & Diabetes & $-21\%_{\footnotesize ns}$ & $-21\%_{\footnotesize *}$ & $0\%_{\footnotesize ns}$ & $41\%_{\footnotesize ***}$ & $37\%_{\footnotesize ***}$ & $35\%_{\footnotesize ***}$ & $42\%_{\footnotesize **}$ & $25\%_{\footnotesize ***}$ \\ 
    & Crime & $68\%_{\footnotesize ns}$ & $45\%_{\footnotesize **}$ & $63\%_{\footnotesize *}$ & $91\%_{\footnotesize ***}$ & $108\%_{\footnotesize ***}$ & $94\%_{\footnotesize *}$ & $104\%_{\footnotesize *}$ & $90\%_{\footnotesize **}$ \\ 
    \hline 
    \end{tabular}
\end{adjustbox}
\end{table}

\begin{figure}[h!]
    \centering

    \begin{subfigure}[b]{0.3\textwidth}
         \caption{$\classifier(\cdot)$: SVC -- CF: \emph{NUN}}
         \includegraphics[width=\textwidth]{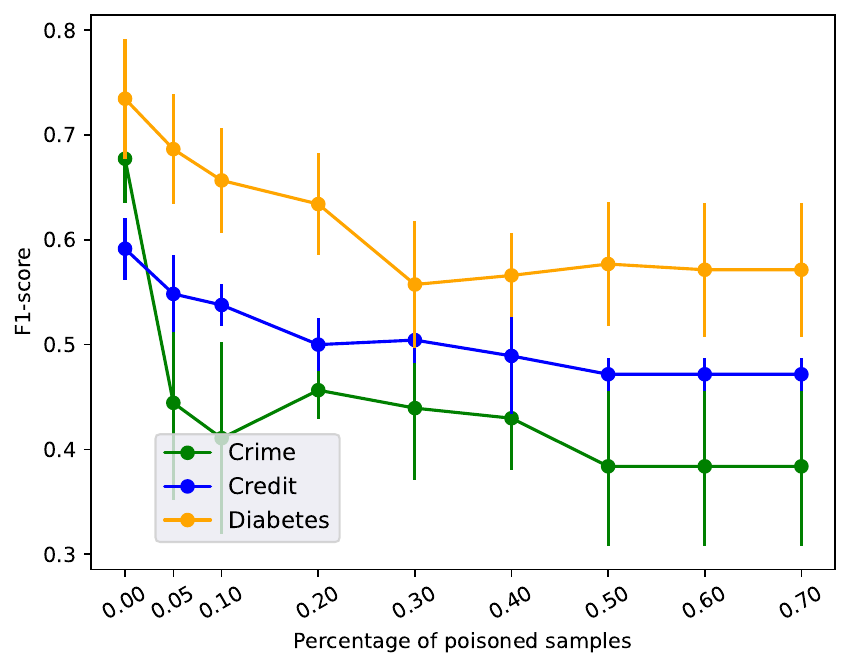}
    \end{subfigure}
    \begin{subfigure}[b]{0.3\textwidth}
         \caption{$\classifier(\cdot)$: SVC -- CF: \emph{DiCE}}
         \includegraphics[width=\textwidth]{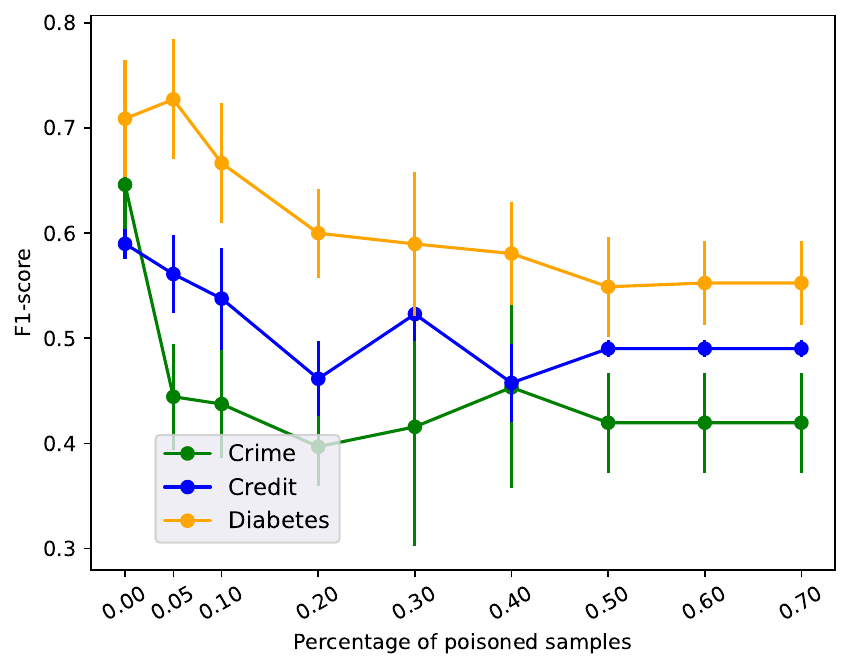}
    \end{subfigure}
    \begin{subfigure}[b]{0.3\textwidth}
         \caption{$\classifier(\cdot)$: SVC -- CF: \emph{Proto}}
         \includegraphics[width=\textwidth]{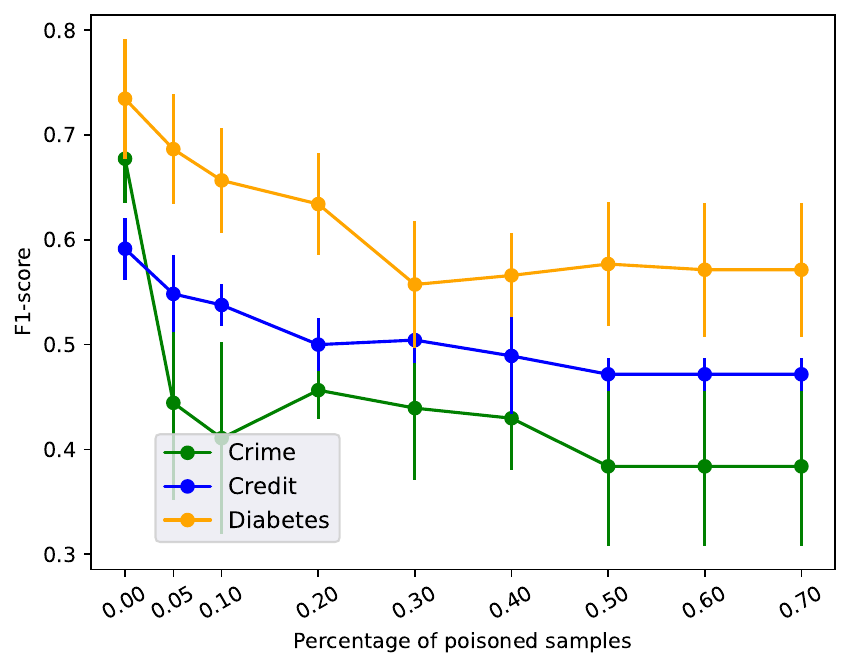}
    \end{subfigure}
    
    \begin{subfigure}[b]{0.3\textwidth}
         \caption{$\classifier(\cdot)$: RNF -- CF: \emph{NUN}}
         \includegraphics[width=\textwidth]{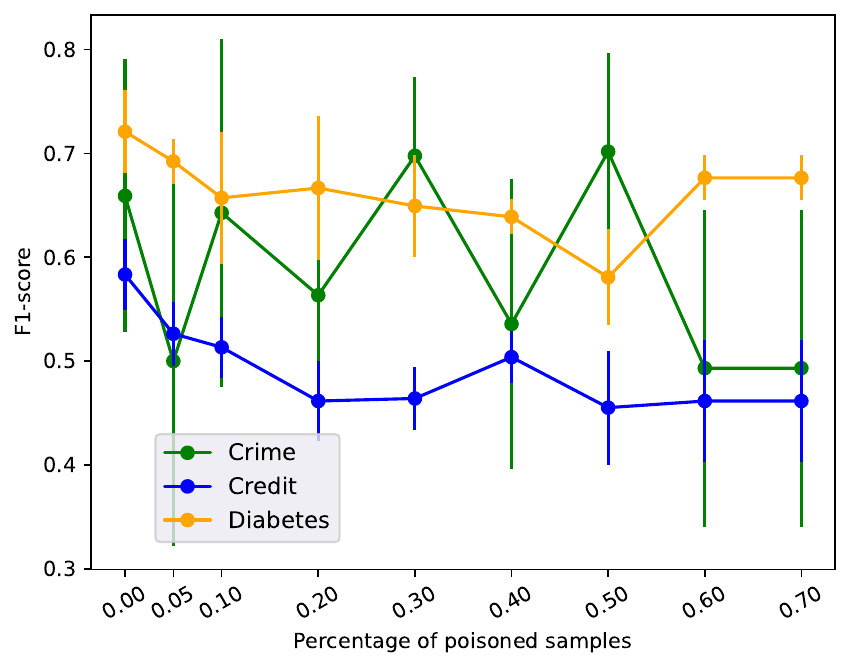}
    \end{subfigure}
    \begin{subfigure}[b]{0.3\textwidth}
         \caption{$\classifier(\cdot)$: RNF -- CF: \emph{DiCE}}
         \includegraphics[width=\textwidth]{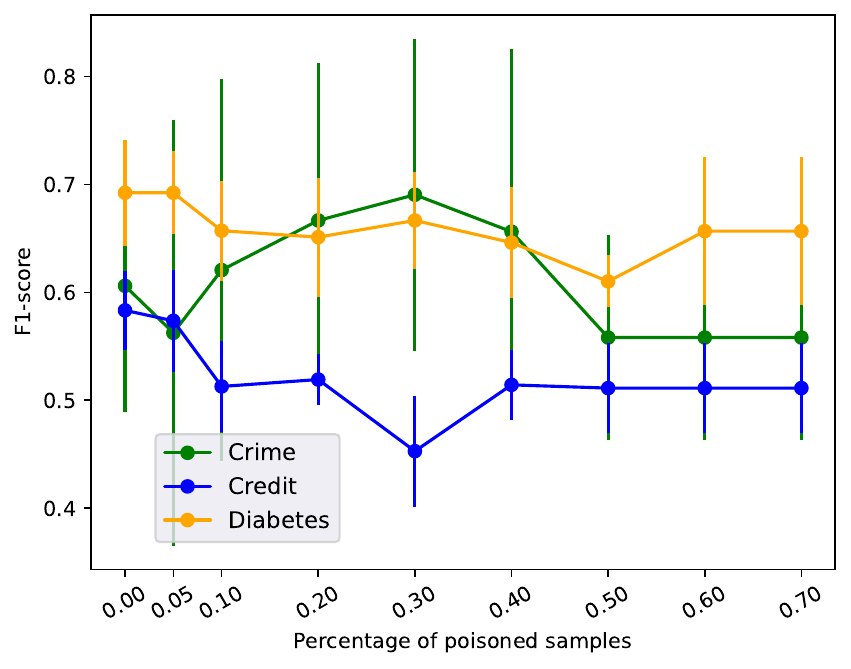}
    \end{subfigure}
    \begin{subfigure}[b]{0.3\textwidth}
         \caption{$\classifier(\cdot)$: RNF -- CF: \emph{Proto}}
         \includegraphics[width=\textwidth]{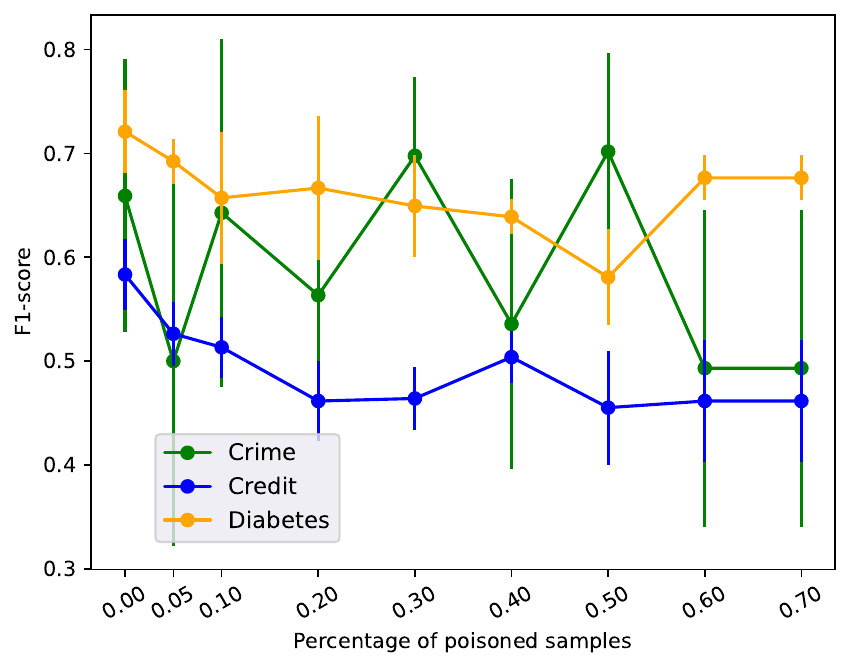}
    \end{subfigure}

    \begin{subfigure}[b]{0.3\textwidth}
         \caption{$\classifier(\cdot)$: DNN -- CF: \emph{DiCE}}
         \includegraphics[width=\textwidth]{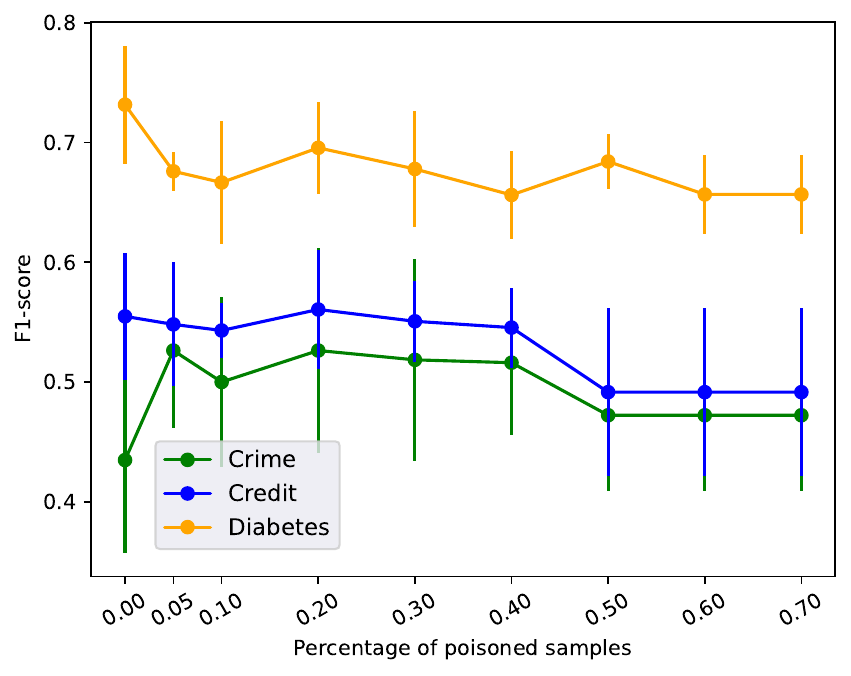}
    \end{subfigure}
    \begin{subfigure}[b]{0.3\textwidth}
         \caption{$\classifier(\cdot)$: DNN -- CF: \emph{Proto}}
         \includegraphics[width=\textwidth]{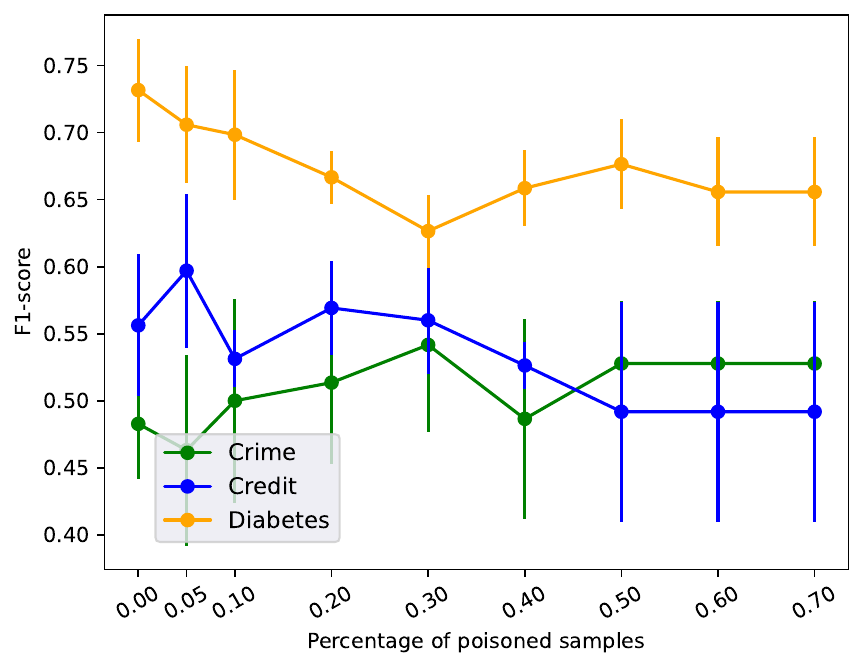}
    \end{subfigure}
       \begin{subfigure}[b]{0.3\textwidth}
         \caption{$\classifier(\cdot)$: DNN -- CF: \emph{NUN}}
         \includegraphics[width=\textwidth]{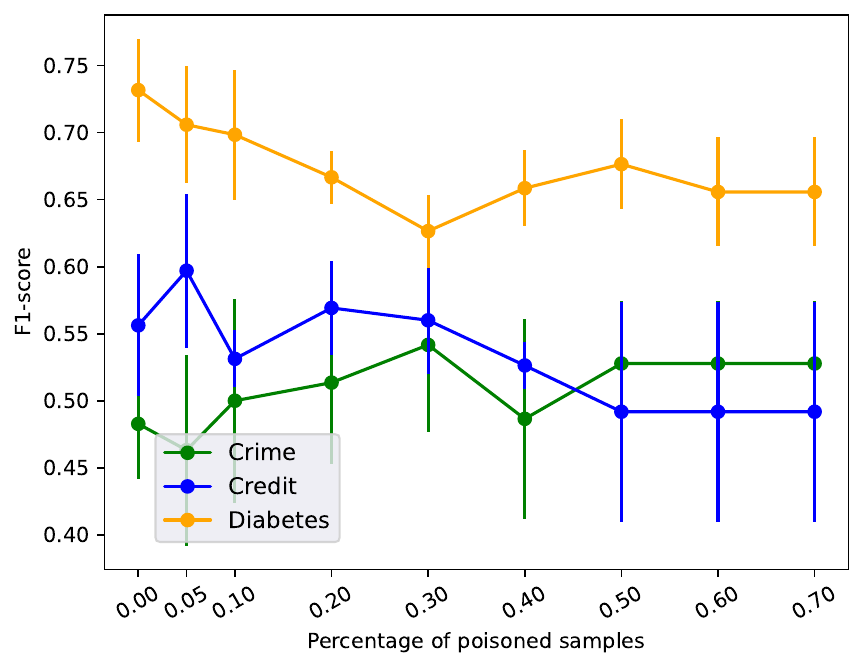}
    \end{subfigure}
    
    \caption{Sub-group data poisoning attack: Median and standard deviation (over all folds) F1-score of the classifier for different percentages of poisoned samples (0\% to 70\%).}
    \label{appendix:fig:exp:results_fairness:accuracy_vs_poisonsamples-1}
\end{figure}

\begin{figure}[h!]
    \centering

    \begin{subfigure}[b]{0.3\textwidth}
         \caption{$\classifier(\cdot)$: SVC -- CF: \emph{NUN}}
         \includegraphics[width=\textwidth]{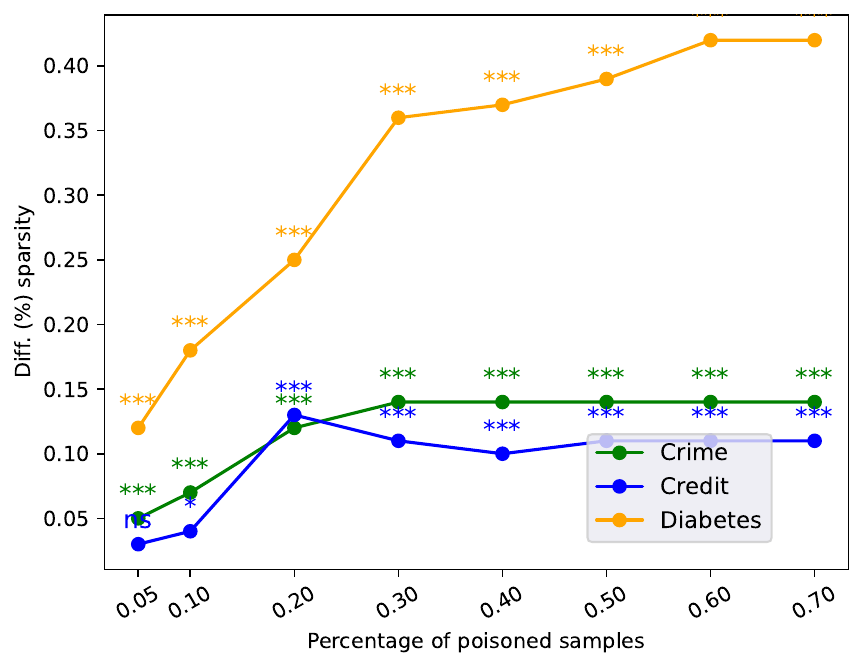}
    \end{subfigure}
    \begin{subfigure}[b]{0.3\textwidth}
         \caption{$\classifier(\cdot)$: SVC -- CF: \emph{DiCE}}
         \includegraphics[width=\textwidth]{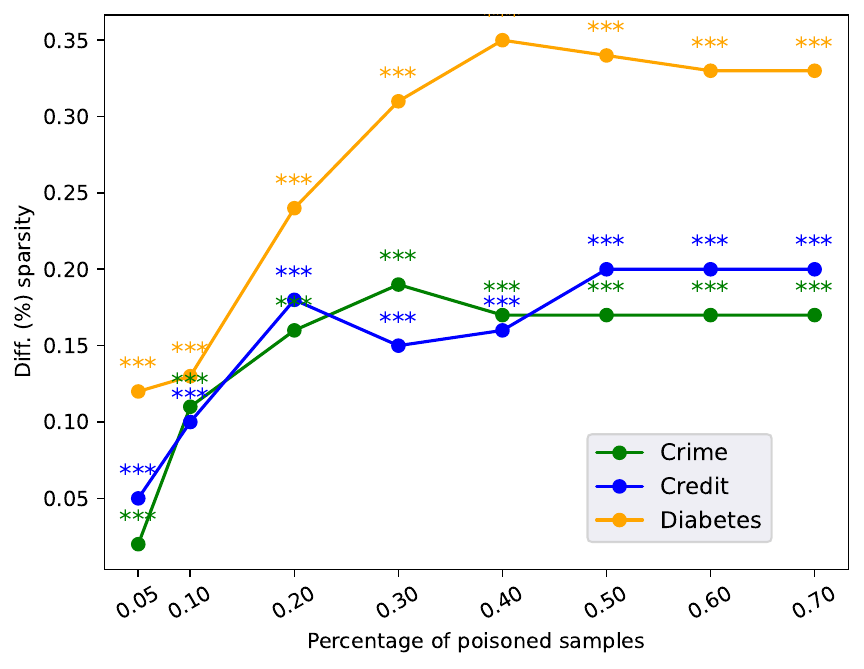}
    \end{subfigure}
    \begin{subfigure}[b]{0.3\textwidth}
         \caption{$\classifier(\cdot)$: SVC -- CF: \emph{Proto}}
         \includegraphics[width=\textwidth]{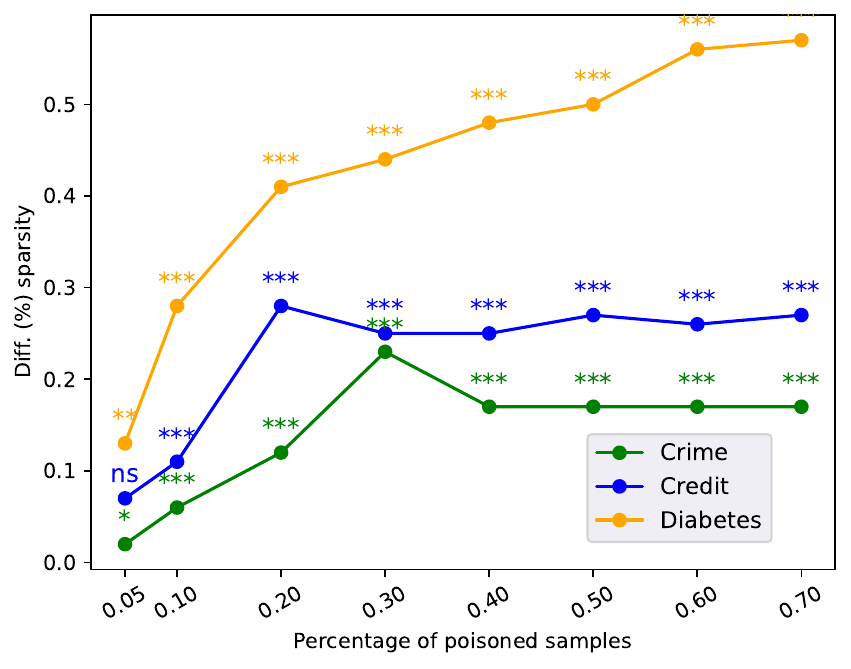}
    \end{subfigure}
    
    \begin{subfigure}[b]{0.3\textwidth}
         \caption{$\classifier(\cdot)$: RNF -- CF: \emph{NUN}}
         \includegraphics[width=\textwidth]{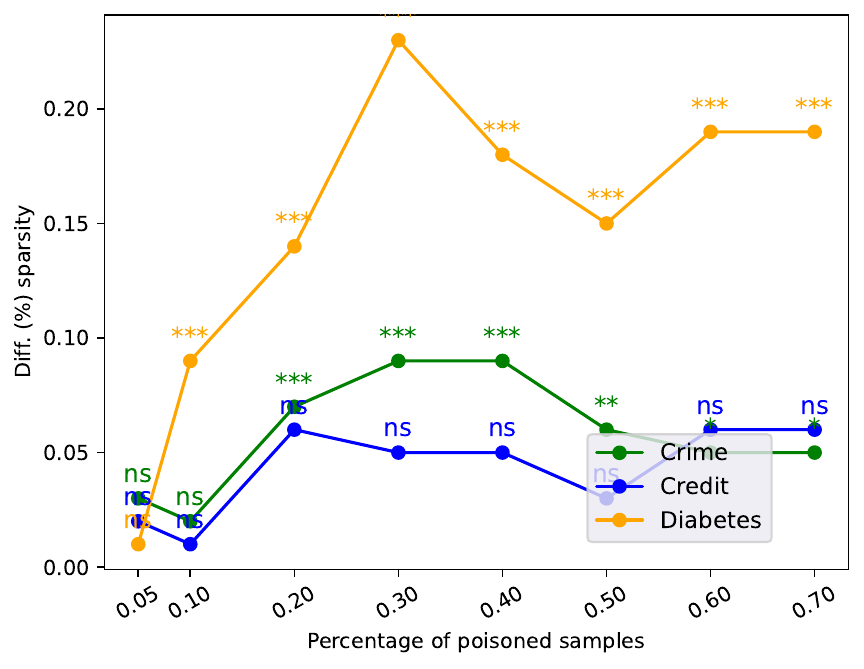}
    \end{subfigure}
    \begin{subfigure}[b]{0.3\textwidth}
         \caption{$\classifier(\cdot)$: RNF -- CF: \emph{DiCE}}
         \includegraphics[width=\textwidth]{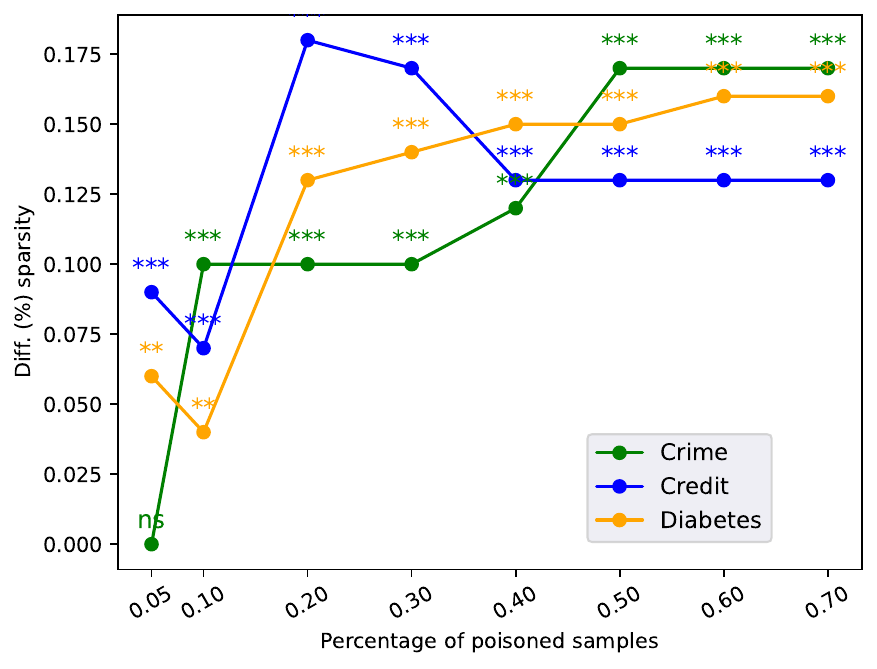}
    \end{subfigure}
    \begin{subfigure}[b]{0.3\textwidth}
         \caption{$\classifier(\cdot)$: RNF -- CF: \emph{Proto}}
         \includegraphics[width=\textwidth]{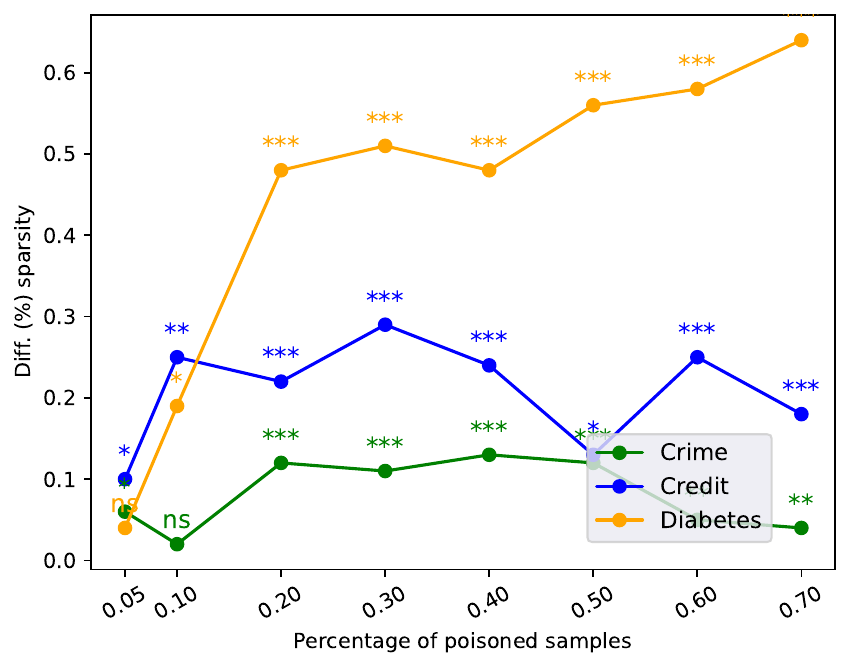}
    \end{subfigure}

    \begin{subfigure}[b]{0.3\textwidth}
         \caption{$\classifier(\cdot)$: DNN -- CF: \emph{DiCE}}
         \includegraphics[width=\textwidth]{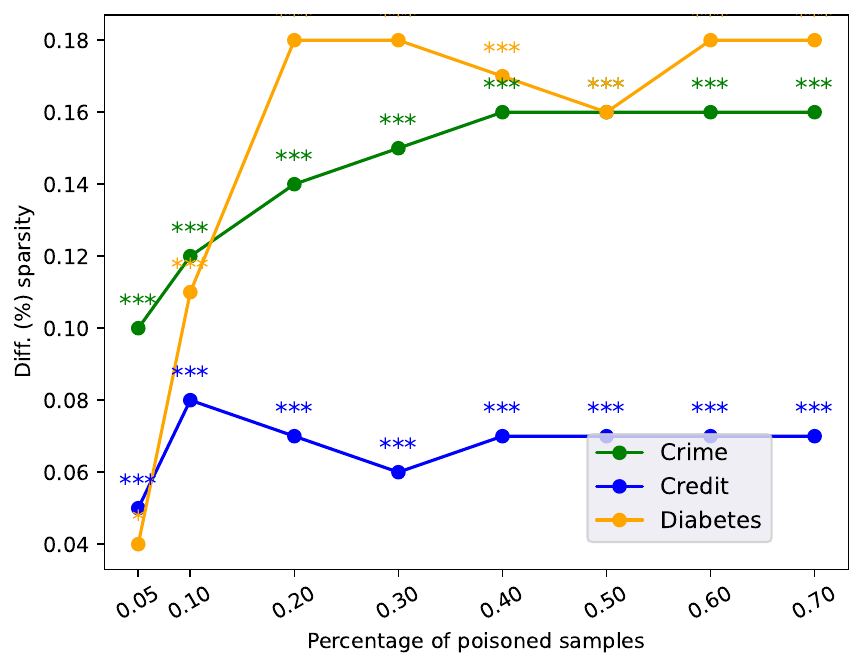}
    \end{subfigure}
    \begin{subfigure}[b]{0.3\textwidth}
         \caption{$\classifier(\cdot)$: DNN -- CF: \emph{Proto}}
         \includegraphics[width=\textwidth]{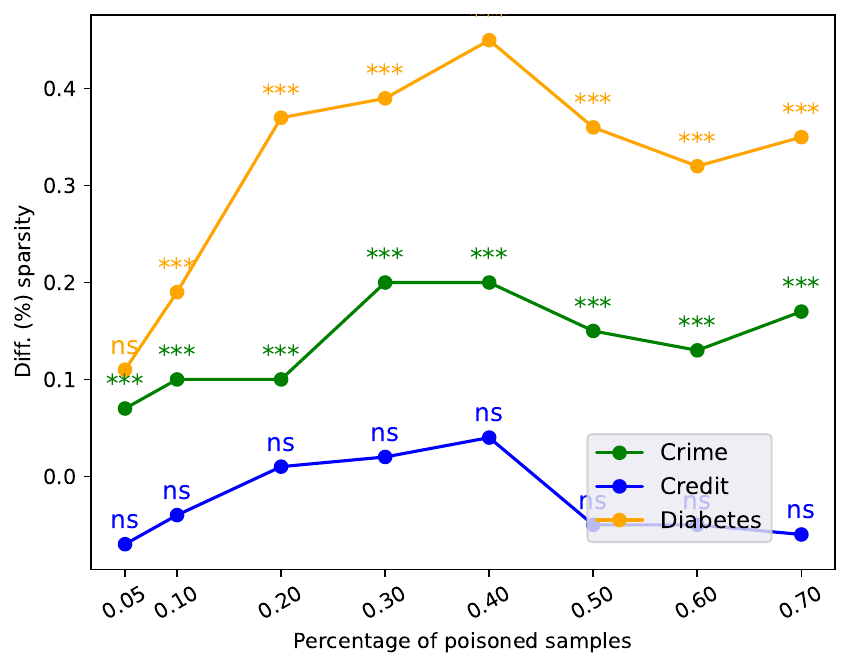}
    \end{subfigure}
    \begin{subfigure}[b]{0.3\textwidth}
         \caption{$\classifier(\cdot)$: DNN -- CF: \emph{NUN}}
         \includegraphics[width=\textwidth]{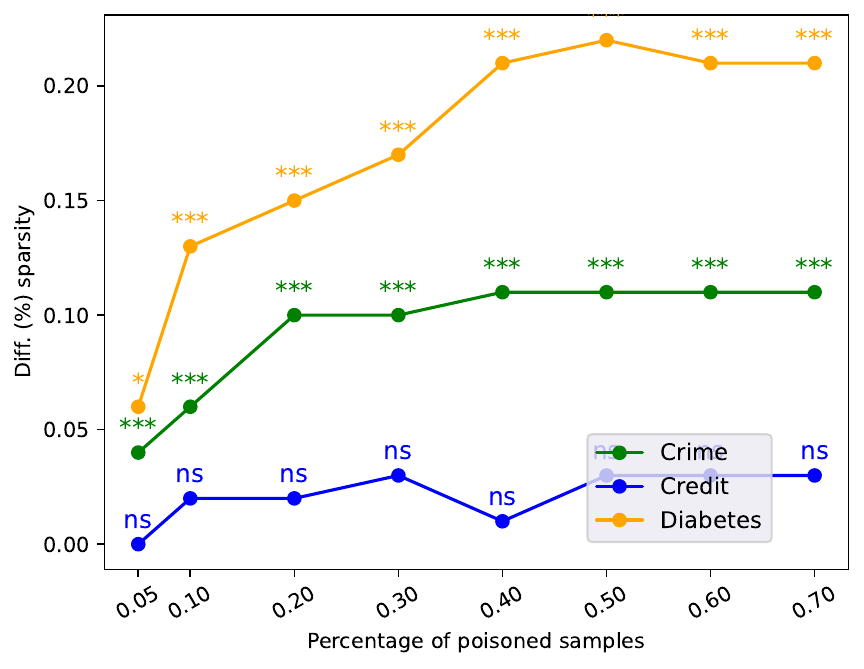}
    \end{subfigure}
    
    \caption{Sub-group data poisoning attack: Difference in the sparsity of the counterfactual explanations for different percentages of poisoned samples (0\% to 70\%). We report the median (over all folds) rounded to two decimal places, as well as the statistical significance according to the Mann-Whitney U test (ns $\implies$ p-value $> 0.05$; * $\implies$ p-value $\leq 0.05$; ** $\implies$ p-value $\leq 0.01$; *** $\implies$ p-value $\leq 0.001$}
    \label{appendix:fig:exp:results_fairness:sparsity_vs_poisonsamples-1}
\end{figure}

\FloatBarrier

\subsection{Local Poisoning Attack}\label{appendix:local}
\begin{figure}[h!]
    \centering

    \begin{subfigure}[b]{0.49\textwidth}
         \includegraphics[width=\textwidth]{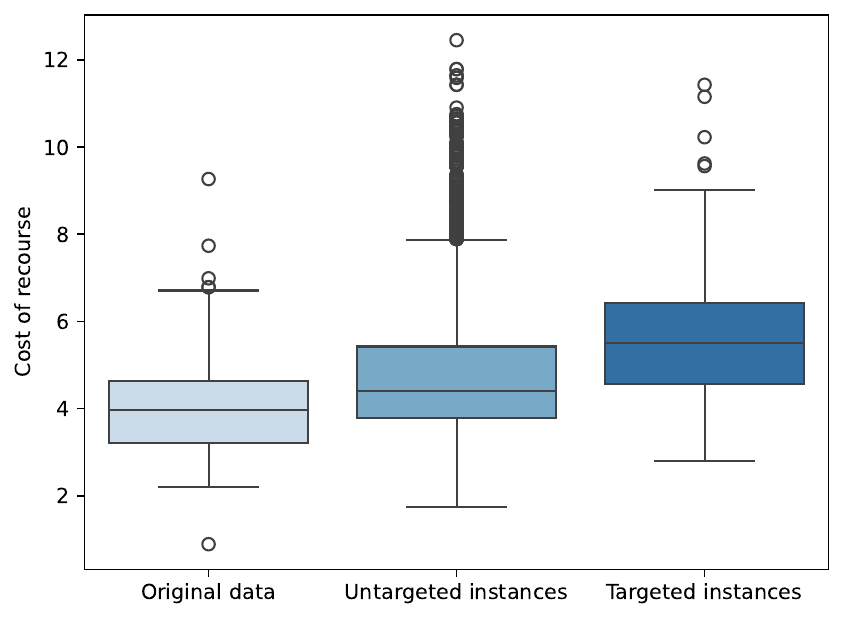}
         \caption{CF: \emph{NUN}}
    \end{subfigure}
    \hfill
    \begin{subfigure}[b]{0.49\textwidth}
         \includegraphics[width=\textwidth]{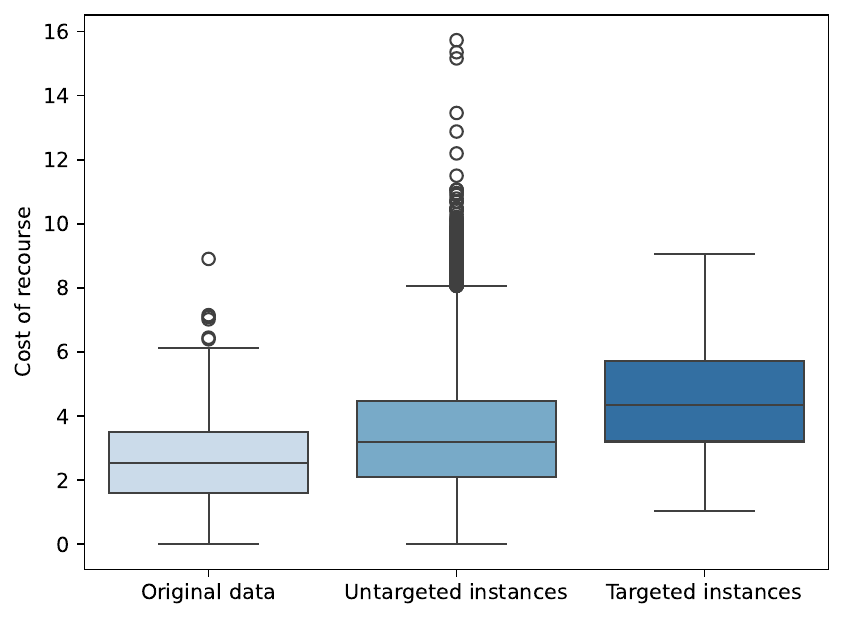}
        \caption{CF: \emph{Proto}}
    \end{subfigure}
    \vfill

    \begin{subfigure}[b]{0.49\textwidth}
         \includegraphics[width=\textwidth]{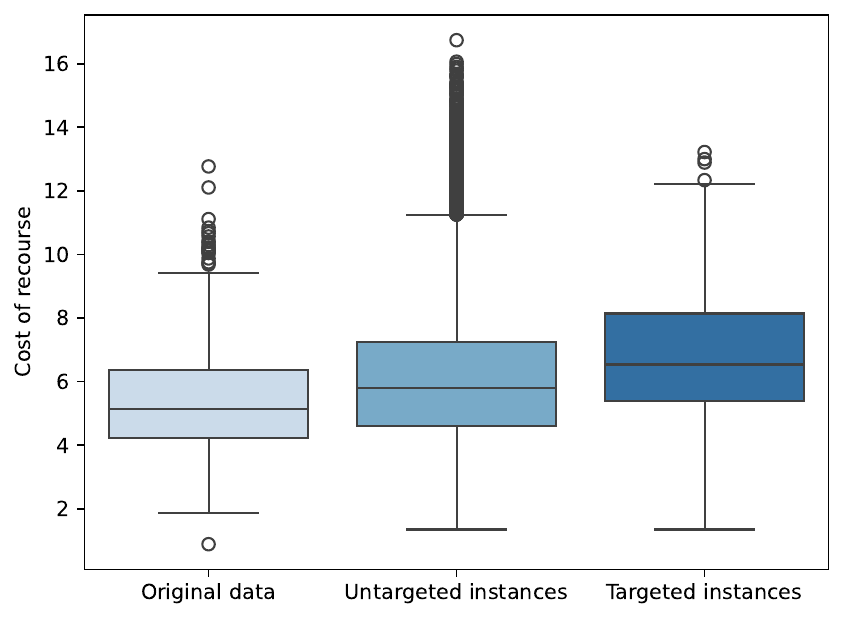}
         \caption{CF: \emph{DiCE}}
    \end{subfigure}
    \hfill
    
    \caption{\emph{Local} data poisoning: Cost of recourse (over all test samples) in the case of the diabetes data set and a DNN classifier. Cost of recourse without any data poisoning, of untargeted instances and targeted instances in a local data poisoning.}
    \label{appendix:fig:expresults:plots_local}
\end{figure}
\FloatBarrier

\subsection{Detection of Poisonous Instances}\label{appendix:detection}
\begin{figure}[h!]
    \centering

    \begin{subfigure}[b]{0.3\textwidth}
         \caption{$\classifier(\cdot)$: SVC -- \emph{Isloation Forest}}
         \includegraphics[width=\textwidth]{exp-results-plots-defense-svc_iforest_recall.pdf}
    \end{subfigure}
    \hfill
    \begin{subfigure}[b]{0.3\textwidth}
         \caption{$\classifier(\cdot)$: SVC -- \emph{k-NN-defense}}
         \includegraphics[width=\textwidth]{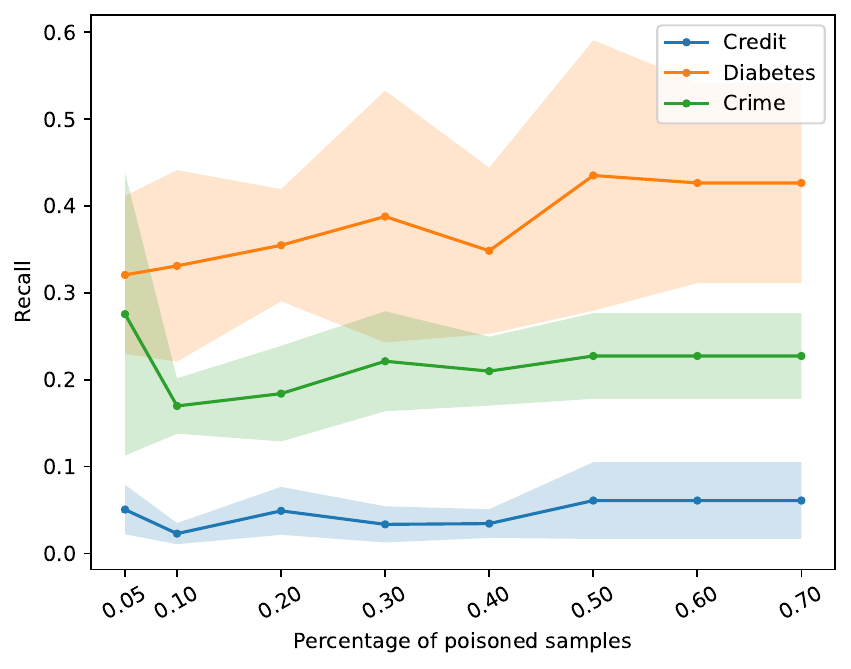}
    \end{subfigure}
    \hfill
    \begin{subfigure}[b]{0.3\textwidth}
         \caption{$\classifier(\cdot)$: SVC -- \emph{$\ell_2$-defense}}
         \includegraphics[width=\textwidth]{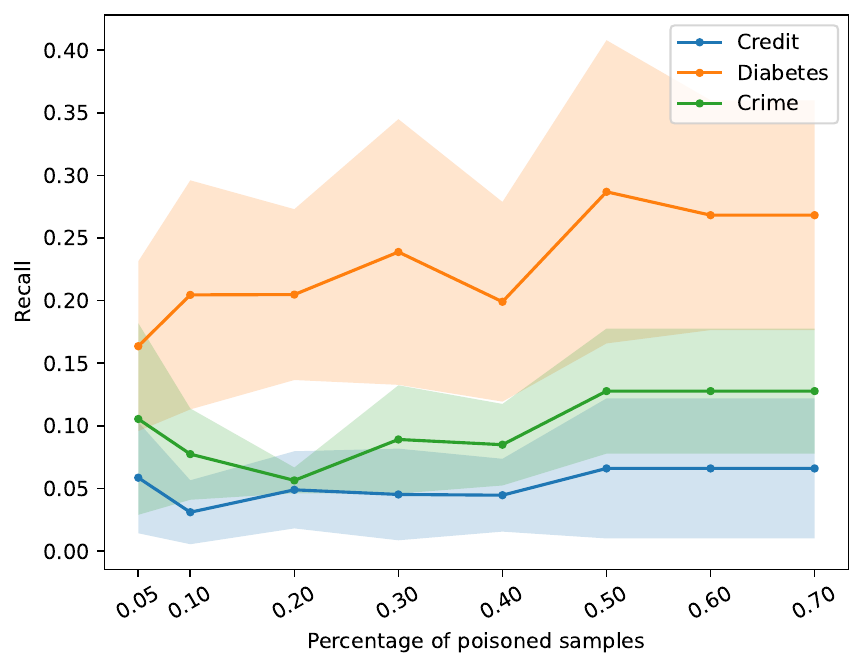}
    \end{subfigure}
    
    \begin{subfigure}[b]{0.3\textwidth}
         \caption{$\classifier(\cdot)$: SVC -- \emph{slabdefense}}
         \includegraphics[width=\textwidth]{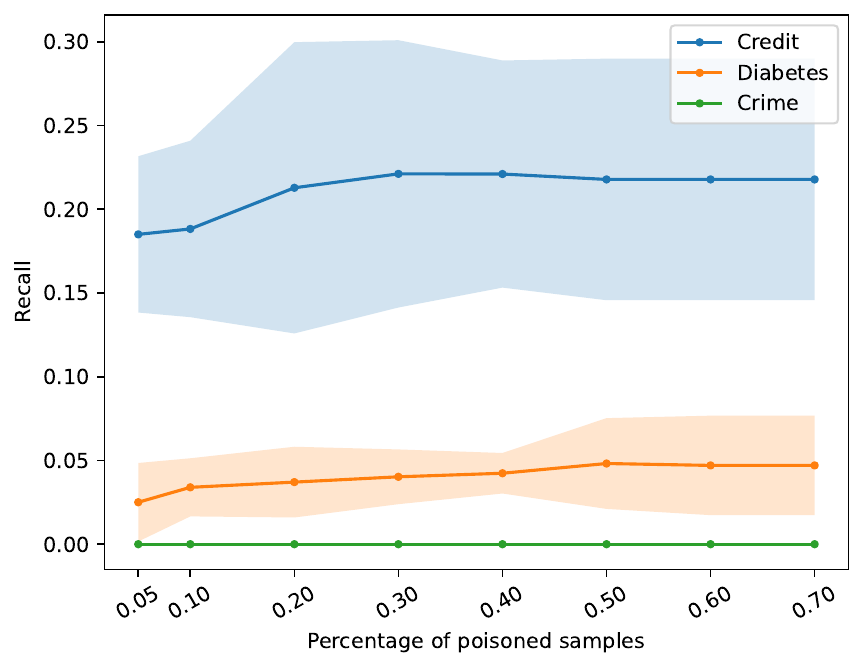}
    \end{subfigure}
    \hfill
    \begin{subfigure}[b]{0.3\textwidth}
         \caption{$\classifier(\cdot)$: SVC -- \emph{Local Outlier Factory}}
         \includegraphics[width=\textwidth]{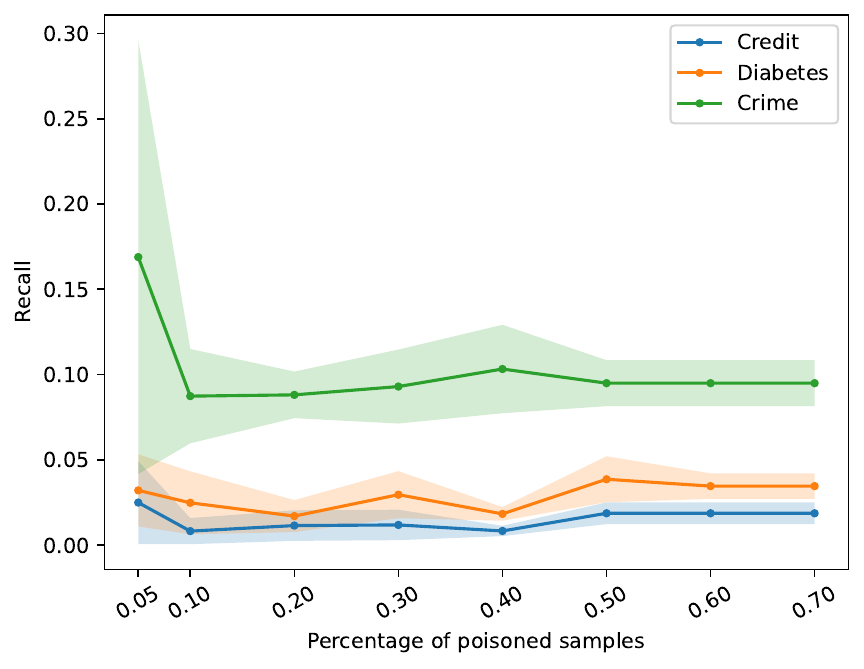}
    \end{subfigure}
    \hfill
    \begin{subfigure}[b]{0.3\textwidth}
         \caption{$\classifier(\cdot)$: RNF -- \emph{Isloation Forest}}
         \includegraphics[width=\textwidth]{exp-results-plots-defense-randforest_iforest_recall.pdf}
    \end{subfigure}
    
    \begin{subfigure}[b]{0.3\textwidth}
         \caption{$\classifier(\cdot)$: RNF -- \emph{k-NN-defense}}
         \includegraphics[width=\textwidth]{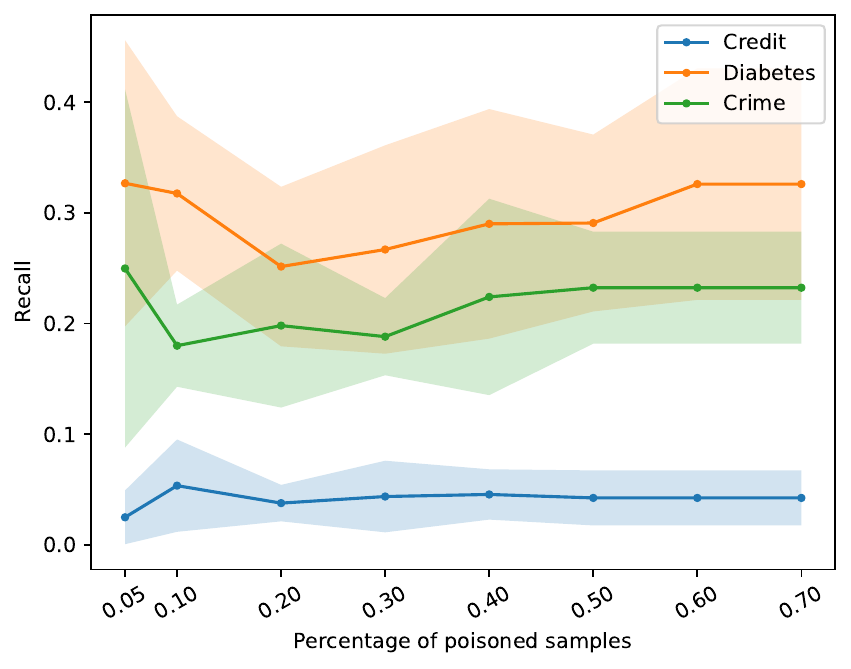}
    \end{subfigure}
    \hfill
    \begin{subfigure}[b]{0.3\textwidth}
         \caption{$\classifier(\cdot)$: RNF -- \emph{$\ell_2$-defense}}
         \includegraphics[width=\textwidth]{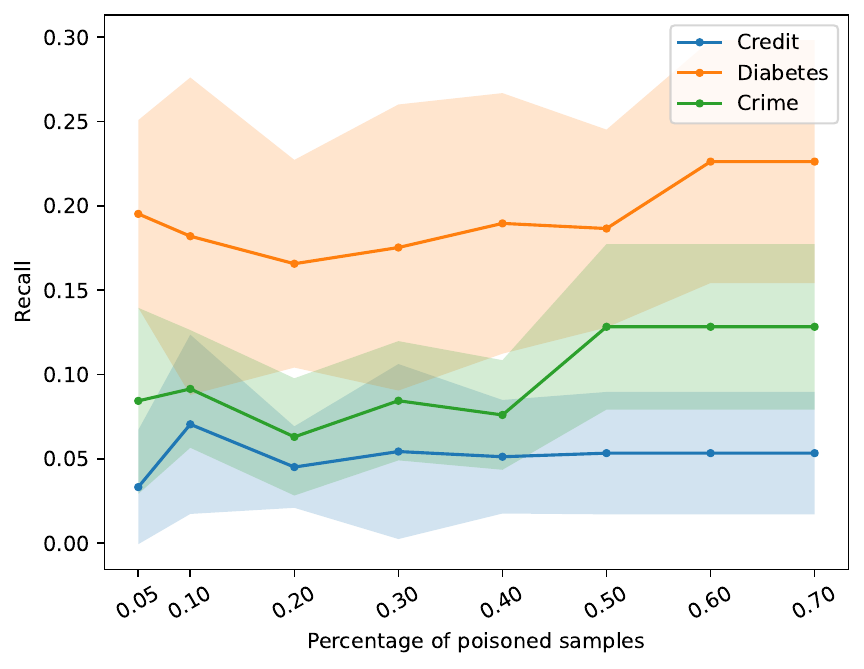}
    \end{subfigure}
    \hfill
    \begin{subfigure}[b]{0.3\textwidth}
         \caption{$\classifier(\cdot)$: RNF -- \emph{slab-defense}}
         \includegraphics[width=\textwidth]{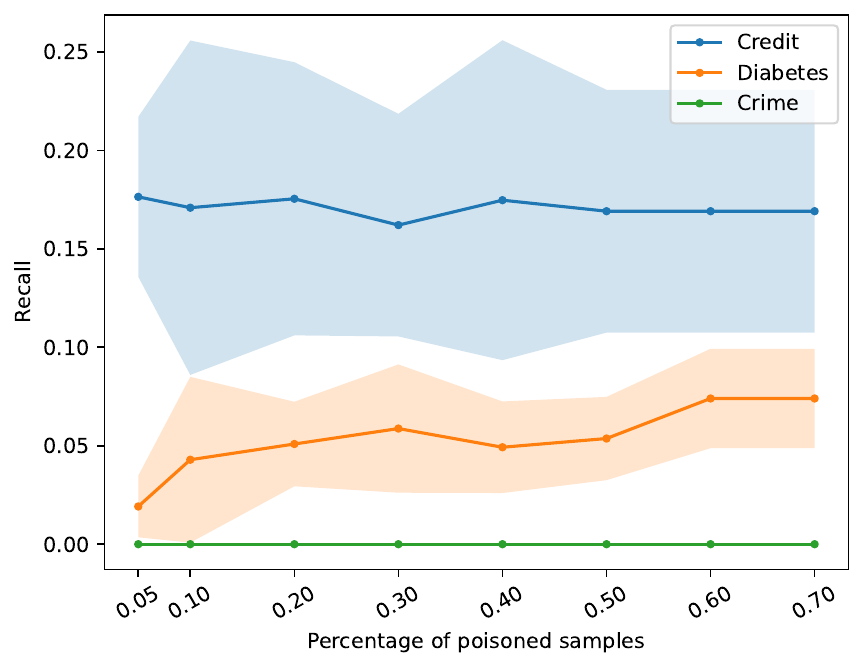}
    \end{subfigure}
    
    \begin{subfigure}[b]{0.3\textwidth}
         \caption{$\classifier(\cdot)$: RNF -- \emph{Local Outlier Factory}}
         \includegraphics[width=\textwidth]{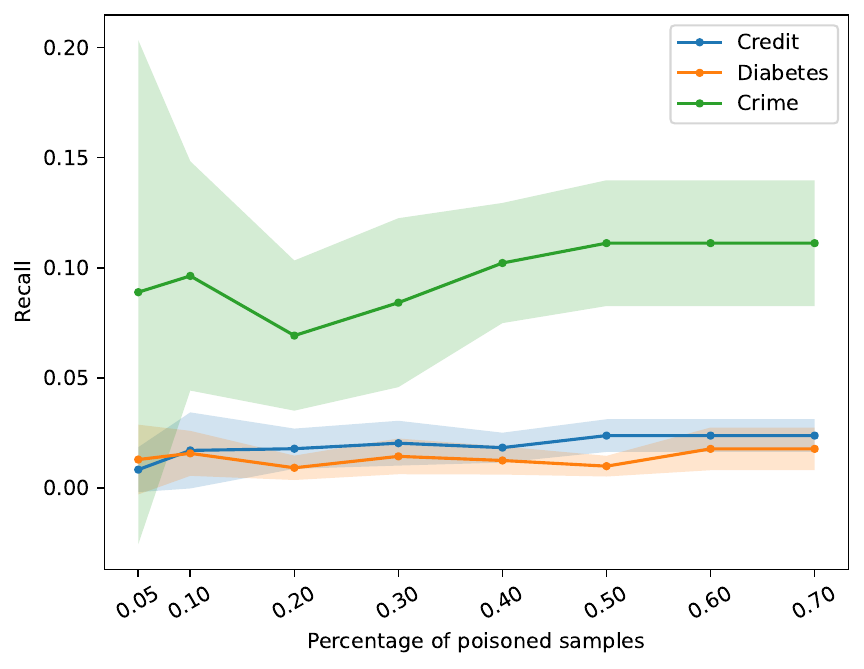}
    \end{subfigure}
    \hfill
    \begin{subfigure}[b]{0.3\textwidth}
         \caption{$\classifier(\cdot)$: DNN -- \emph{Isloation Forest}}
         \includegraphics[width=\textwidth]{exp-results-plots-defense-dnn_iforest_recall.pdf}
    \end{subfigure}
    \hfill
    \begin{subfigure}[b]{0.3\textwidth}
         \caption{$\classifier(\cdot)$: DNN -- \emph{k-NN-defense}}
         \includegraphics[width=\textwidth]{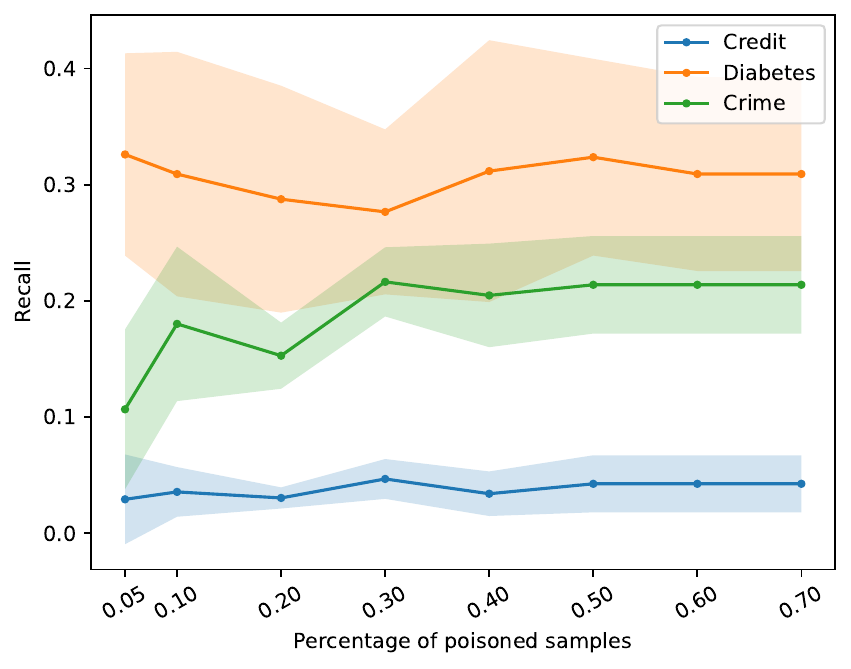}
    \end{subfigure}
    
    \begin{subfigure}[b]{0.3\textwidth}
         \caption{$\classifier(\cdot)$: DNN -- \emph{$\ell_2$-defense}}
         \includegraphics[width=\textwidth]{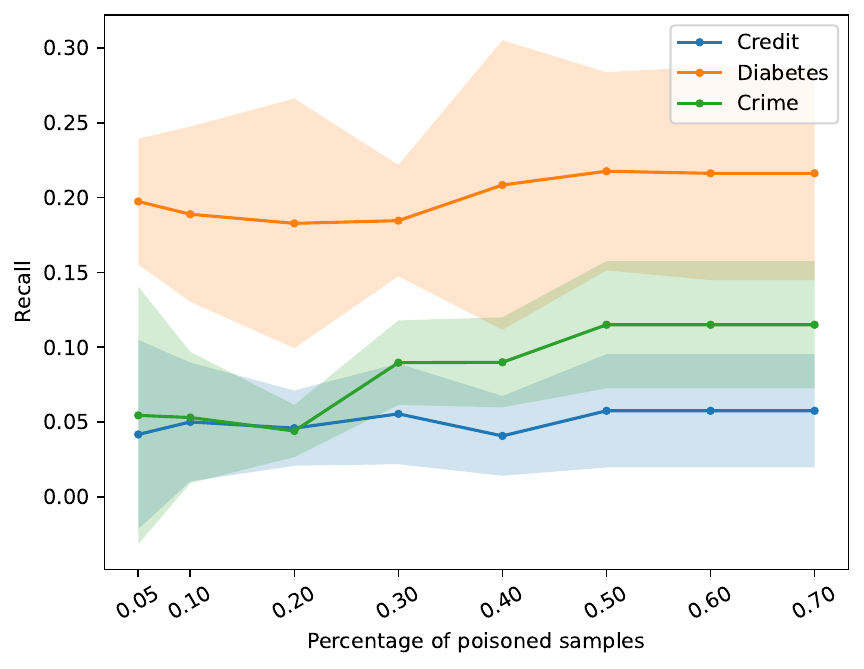}
    \end{subfigure}
    \hfill
    \begin{subfigure}[b]{0.3\textwidth}
         \caption{$\classifier(\cdot)$: DNN -- \emph{slab-defense}}
         \includegraphics[width=\textwidth]{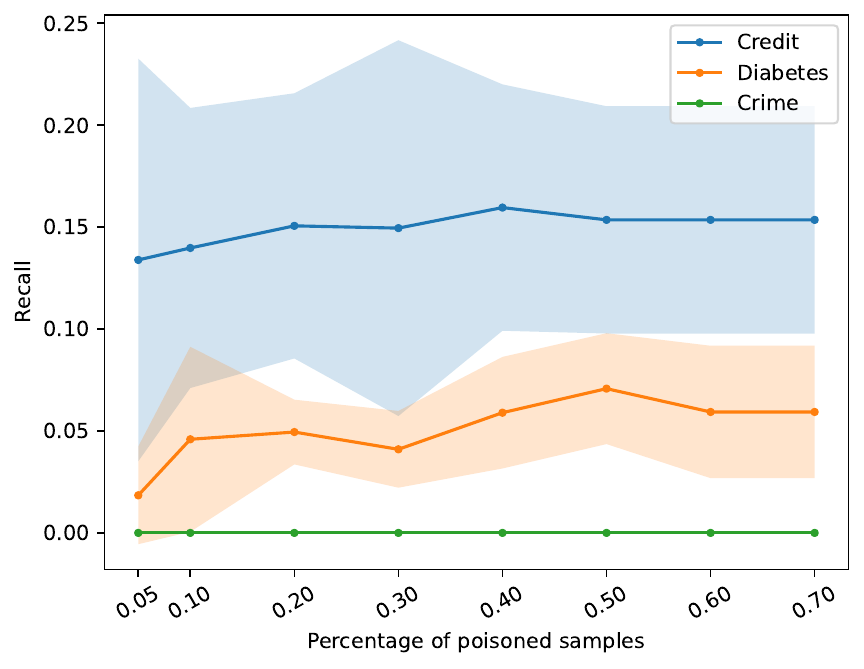}
    \end{subfigure}
    \hfill
    \begin{subfigure}[b]{0.3\textwidth}
         \caption{$\classifier(\cdot)$: DNN -- \emph{Local Outlier Factory}}
         \includegraphics[width=\textwidth]{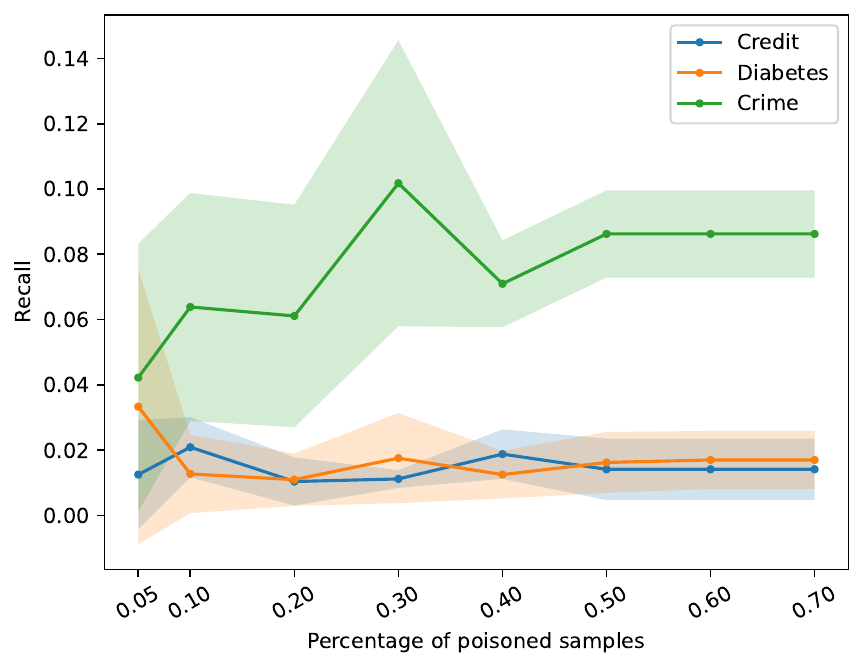}
    \end{subfigure}

    \caption{Global data poisoning attack: Recall of different data sanitization methods evaluated on different percentages of poisoned samples (0\% to 70\%). We report the mean and standard deviation over all folds (larger numbers are better).}

    \label{fig:defenses:results_all}
\end{figure}
\FloatBarrier

\subsection{Ablation study}
\begin{figure}[h!]
    \centering

    \begin{subfigure}{0.3\textwidth}
         \caption{$\classifier(\cdot)$: SVC -- CF: \emph{NUN}}
         \includegraphics[width=\textwidth]{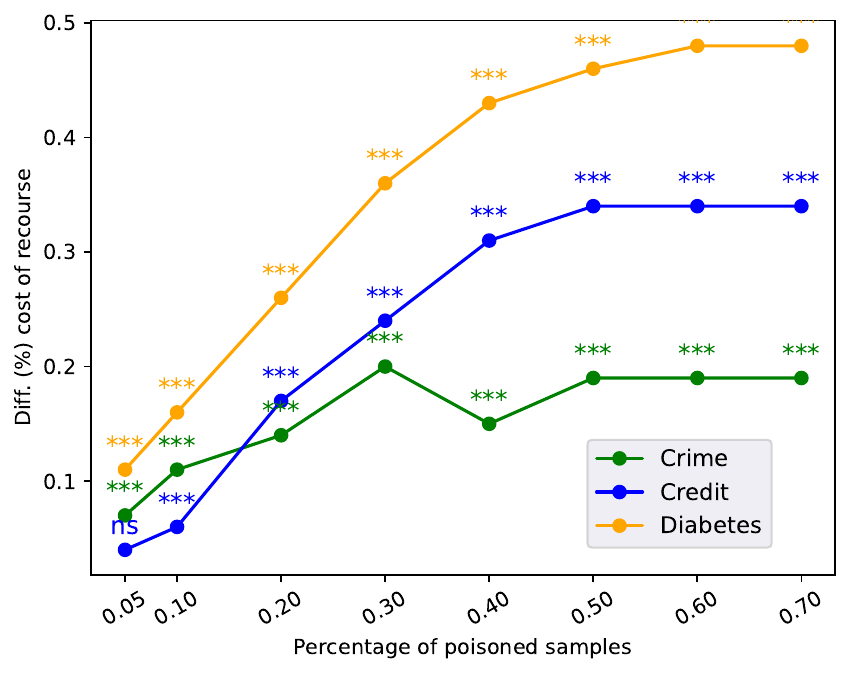}
    \end{subfigure}
    \begin{subfigure}{0.3\textwidth}
         \caption{$\classifier(\cdot)$: SVC -- CF: \emph{DiCE}}
         \includegraphics[width=\textwidth]{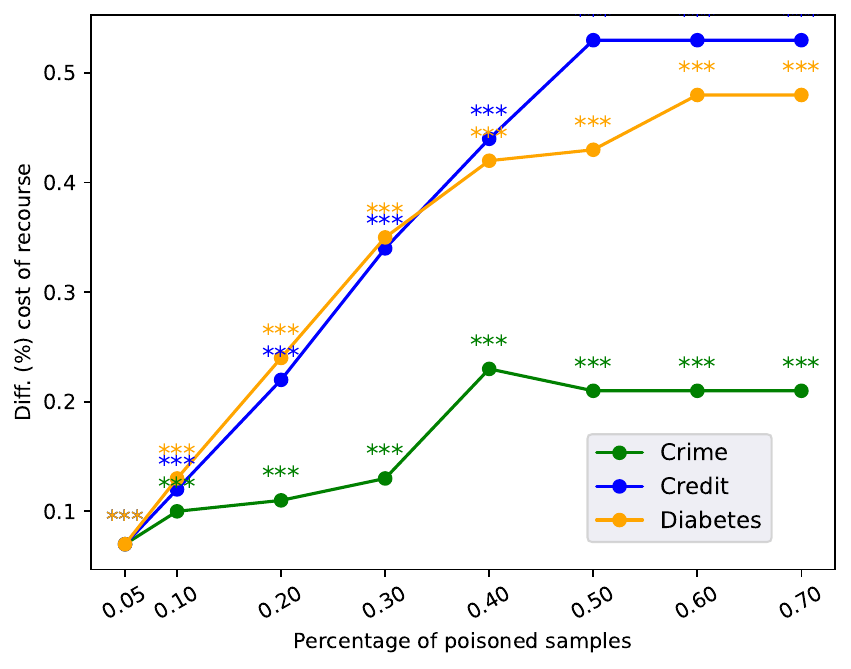}
    \end{subfigure}
    \begin{subfigure}{0.3\textwidth}
         \caption{$\classifier(\cdot)$: SVC -- CF: \emph{Proto}}
         \includegraphics[width=\textwidth]{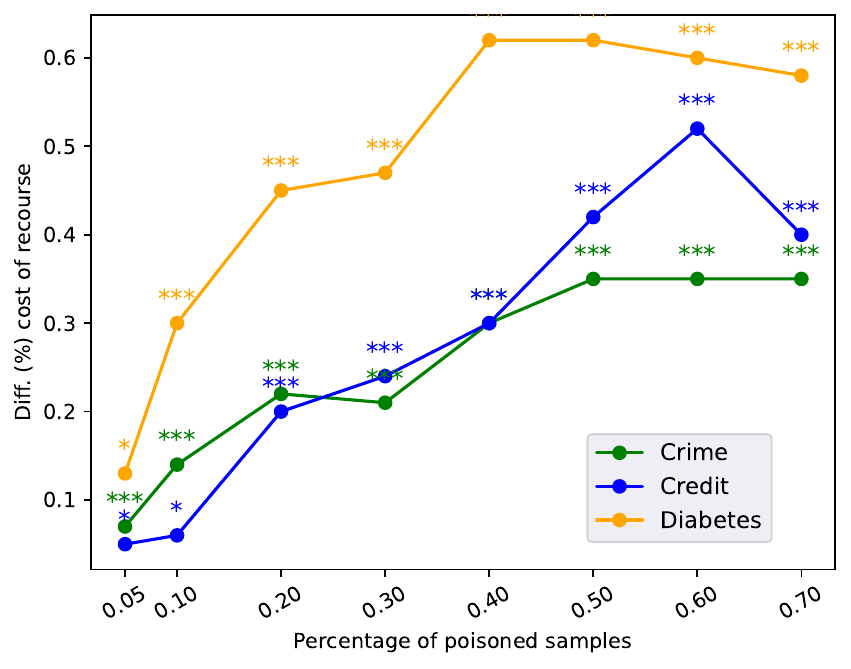}
    \end{subfigure}
    
    \begin{subfigure}{0.3\textwidth}
         \caption{$\classifier(\cdot)$: RNF -- CF: \emph{Proto}}
         \includegraphics[width=\textwidth]{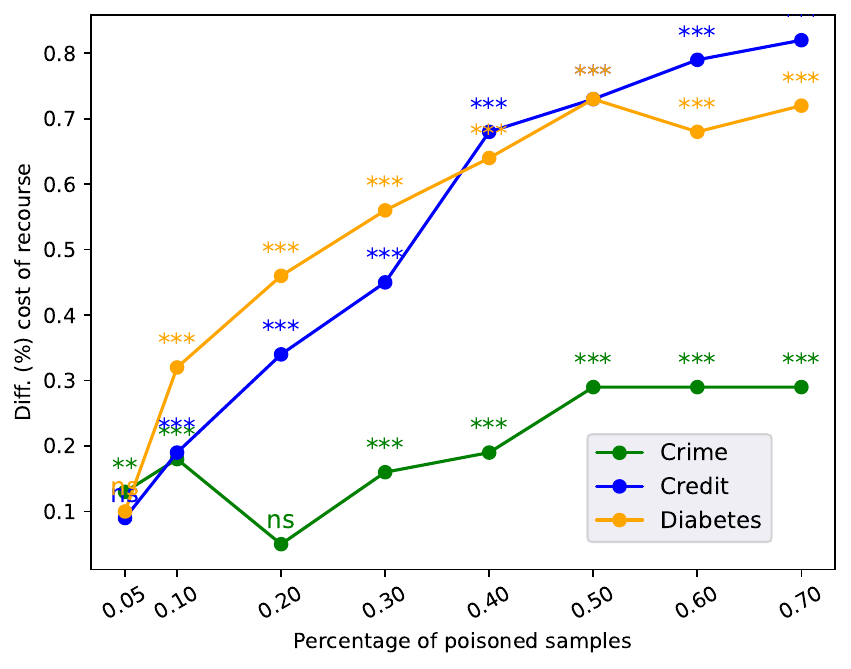}
    \end{subfigure}
        \begin{subfigure}[b]{0.3\textwidth}
         \caption{$\classifier(\cdot)$: RNF -- CF: \emph{NUN}}
         \includegraphics[width=\textwidth]{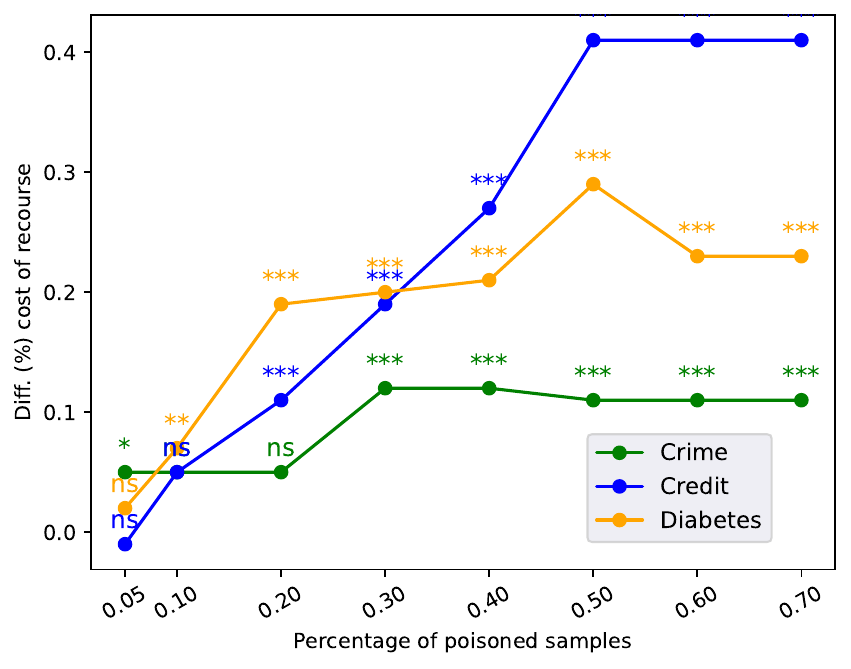}
    \end{subfigure}
    \begin{subfigure}[b]{0.3\textwidth}
         \caption{$\classifier(\cdot)$: RNF -- CF: \emph{DiCE}}
         \includegraphics[width=\textwidth]{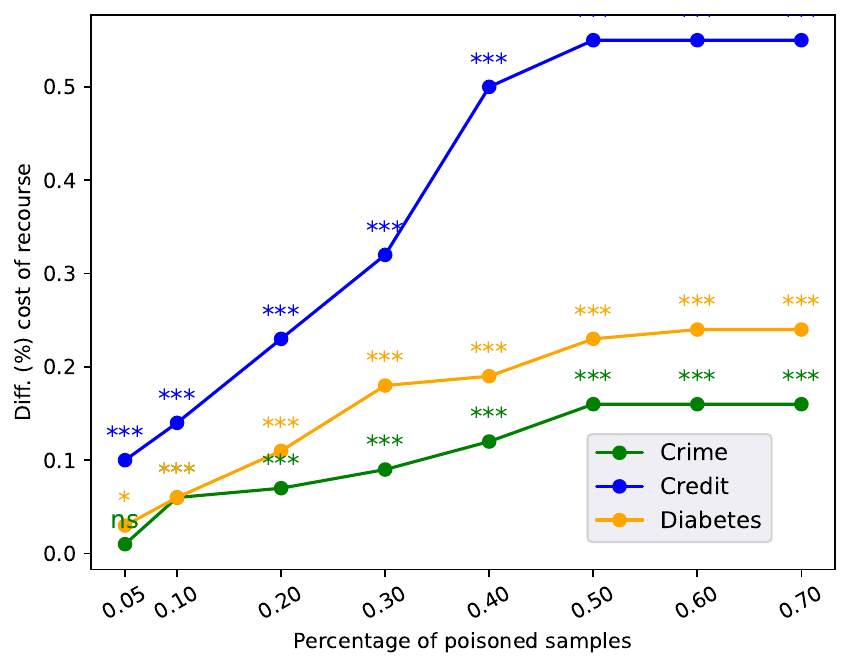}
    \end{subfigure}

    \begin{subfigure}[b]{0.3\textwidth}
         \caption{$\classifier(\cdot)$: DNN -- CF: \emph{NUN}}
         \includegraphics[width=\textwidth]{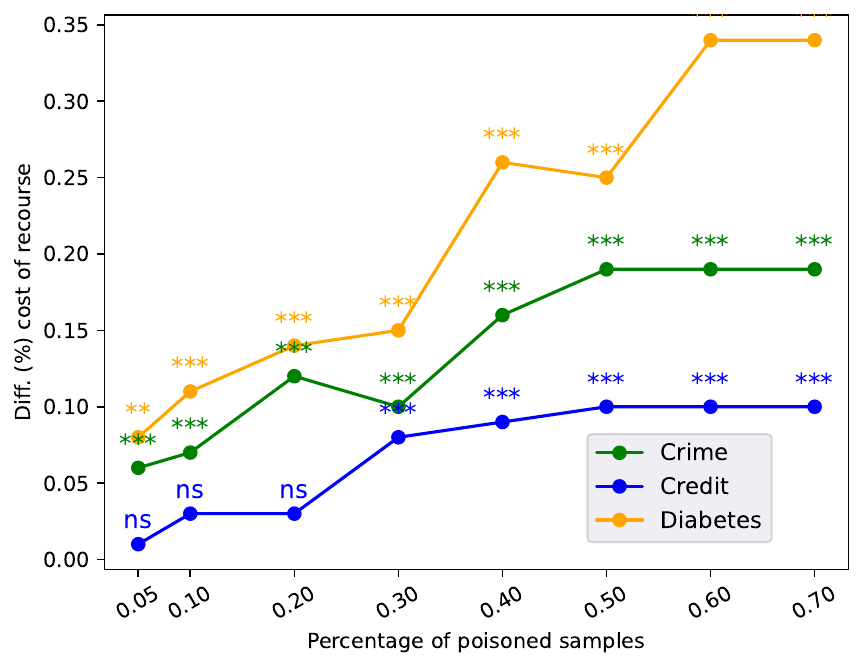}
    \end{subfigure}
    \begin{subfigure}[b]{0.3\textwidth}
         \caption{$\classifier(\cdot)$: DNN -- CF: \emph{DiCE}}
         \includegraphics[width=\textwidth]{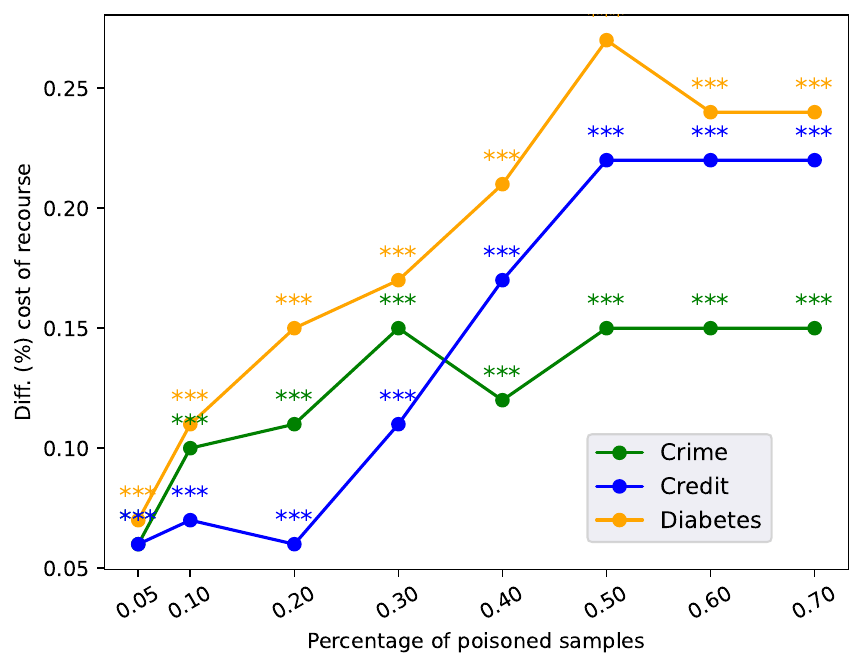}
    \end{subfigure}
    \begin{subfigure}{0.3\textwidth}
         \caption{$\classifier(\cdot)$: DNN -- CF: \emph{Proto}}
         \includegraphics[width=\textwidth]{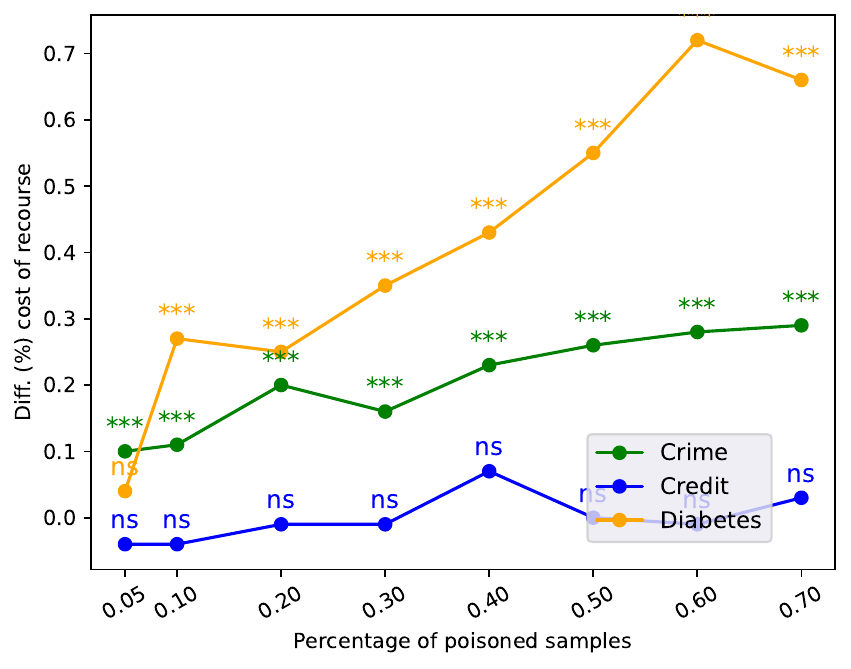}
    \end{subfigure}
    
    \caption{\emph{Ablation study (uniform sampling)} - Global data poisoning attack: Difference (percentage) in the cost of recourse vs. percentage of poisoned instances (5\% to 70\%). We report the median (over all folds) rounded to two decimal places, as well as the statistical significance according to the Mann-Whitney U test (ns $\implies$ p-value $> 0.05$; * $\implies$ p-value $\leq 0.05$; ** $\implies$ p-value $\leq 0.01$; *** $\implies$ p-value $\leq 0.001$.}
    \label{appendix:fig:exp:results:ablation:recourse_vs_poisonsamples-1}
\end{figure}

\FloatBarrier



\end{document}